\documentclass{article}

\PassOptionsToPackage{numbers,sort}{natbib}

 \usepackage[main, final]{neurips_2026}

\usepackage[utf8]{inputenc} 
\usepackage[T1]{fontenc}    
\usepackage{hyperref}       
\usepackage{url}            
\usepackage{booktabs}       
\usepackage{amsfonts}       
\usepackage{nicefrac}       
\usepackage{microtype}      

\usepackage{adjustbox}
\usepackage{tabularx}
\usepackage{array}
\usepackage{multirow}
\usepackage{colortbl}
\usepackage{subfigure}
\usepackage{enumitem}
\usepackage{wrapfig}
\usepackage{colortbl}
\usepackage{caption}
\usepackage[dvipsnames]{xcolor}

\usepackage{comment}
\usepackage{algorithm}
\usepackage{algorithmic}
\usepackage{amsmath}
\usepackage{amssymb}
\usepackage{mathtools}
\usepackage{amsthm}
\usepackage{bm,bbm}
\usepackage{multirow}

\newlength{\imgheight}

\DeclareMathOperator*{\expectation}{\mathbb{E}}
\newcommand{\ours}{\text{SaND}}

\title{Skill-Adaptive Noise Scheduling for Diffusion Policies}

\author{%
  Woo Kyung Kim\textsuperscript{1}, Gwangpyo Yoo\textsuperscript{1}, Eunyoung Park\textsuperscript{2}, Honguk Woo\textsuperscript{1}\thanks{Corresponding author.} \\
  \textsuperscript{1} Department of Computer Science and Engineering, Sungkyunkwan University\\
  \textsuperscript{2} Department of Artificial Intelligence, Sungkyunkwan University\\
  \texttt{\{kwk2696,necrocathy,eunyoung927,hwoo\}@skku.edu}
}

\begin{document}

\maketitle

\begin{abstract}
  Recent advances in diffusion policies have demonstrated strong effectiveness when integrated into hierarchical skill-based learning frameworks, leveraging offline data to capture multimodal and temporally abstracted behaviors. However, prior approaches have rarely considered the diffusion process as a learnable component, adopting fixed noise schedules applied uniformly across skills, regardless of their distinct behavioral contexts. Such a design fundamentally limits the achievable performance, particularly for contact-rich and dexterous manipulation tasks where different skills exhibit distinct couplings between action dimensions.
  In this paper, we present a skill-adaptive noise diffusion framework ($\ours$) which introduces a learnable, skill-conditioned noise schedule that modulates the diffusion process per action dimension, adapting the noising dynamics to each skill. 
  To prevent the learned noise schedule from collapsing into trivial solutions, we employ an auxiliary inverse objective that reconstructs the skill embedding from the Legendre coefficients of the derivative of the noise schedule, encouraging the schedules to remain skill-discriminative.
  This formulation naturally enables adaptive control over the denoising steps for each skill, where noise schedules are clustered in the Legendre coefficient space, and the appropriate number of denoising steps is determined at the point where reconstruction error abruptly increases. 
  At inference, we identify each skill embedding's cluster from its noise schedule and apply the corresponding number of denoising steps, dedicating more compute to skills requiring fine-grained control and less to simpler ones.
  %
  Through experiments in simulation and on real robot hardware, we demonstrate that $\ours$ outperforms existing skill learning and planning baselines on contact-rich and dexterous manipulation tasks, achieving higher success rates while requiring fewer denoising steps.
\end{abstract}

\section{Introduction} \label{sec:intro}
Hierarchical skill-based learning has emerged as a promising paradigm for robotic manipulation by abstracting temporal behaviors into latent skill representations~\cite{sprl:spirl,sprl:fist}. Meanwhile, diffusion policies have demonstrated strong capability in modeling sequential decision-making tasks, owing to their ability to capture multimodal action distributions~\cite{DP:DiffBC,DP:DD,DP:DiffQL} and generate coherent trajectory-level plans~\cite{DP:Diffuser,DP:DV,DP:HD}. Building on these successes, recent works have augmented skill-based frameworks with diffusion models to further enhance the expressiveness of skill decoders~\cite{DP:playfusion,DP:DuSkill,DP:ICPAD,DP:SkillDiffuser,DP:LCD}. 

Despite their effectiveness, prior methods share key limitations due to their reliance on a fixed noise schedule across all skills and action dimensions. Such uniform denoising prevents the model from leveraging mutual information among actions, i.e., using easily inferred dimensions to guide the harder ones. This is particularly important in contact-rich and dexterous manipulation tasks, where kinematic contact constraints and dynamic equilibrium tightly couple action dimensions. Moreover, these coupling patterns vary across skills, as different behaviors yield distinct contact and force structures.
Theoretically, a fixed noise schedule yields a loose bound on the diffusion ELBO~\cite{DM:MuLAN}. Under the deterministic denoising process of diffusion policies~\cite{DP:DD,DP:BESO,DP:2DP}, the resulting inaccuracy propagates irreversibly and cannot be mitigated by additional denoising steps (see Appendix~\ref{sec:appendix:c:step}).

\begin{figure*}[t] 
    \centering
    \includegraphics[width=.96\textwidth]{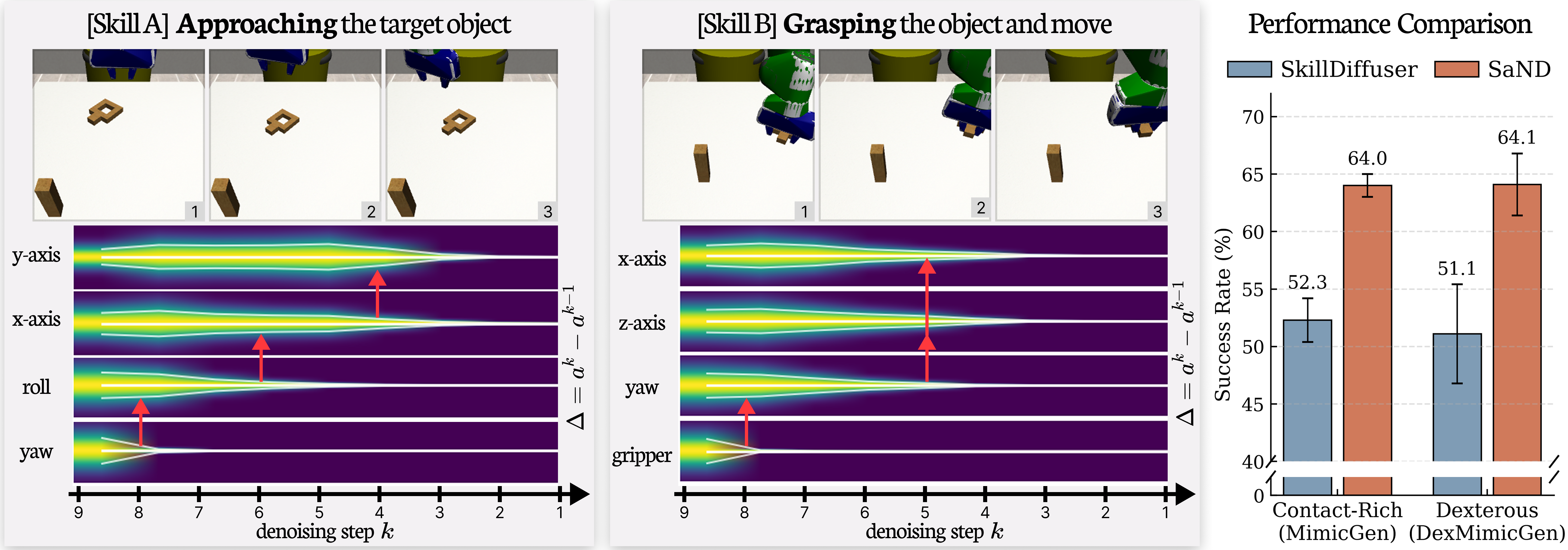}
    \caption{Concept of $\ours$: The first two panels visualize per-step denoising residuals along each action dimension for two different skills (approaching and grasping), illustrating that $\ours$ learns distinct per-dimension denoising priorities for each skill. Red arrows mark the convergence point of each action dimension, showing that dimensions are resolved at distinct stages of the denoising process. The rightmost panel compares success rates of $\ours$ with those of SkillDiffuser, which uses a fixed noise schedule, on contact-rich and dexterous manipulation tasks, showing consistent gains.} 
    \label{fig:intro}
\end{figure*}
To address this, inspired by adaptive noise schedules in image diffusion models~\cite{DM:VDM,DM:MuLAN}, we propose to learn a skill-conditioned noise schedule that adapts the denoising rate per action dimension for each skill. 
Such adaptive scheduling enables prioritized denoising across action dimensions, first resolving dimensions that are easier to predict and subsequently refining more contact-sensitive ones, yielding a tighter ELBO on the action distributions.
This is analogous to the Knothe–Rosenblatt map, which guarantees exact likelihood evaluation by decomposing a joint distribution into a chain of conditionals~\cite{backhoff2017causal}. 
%
In Figure~\ref{fig:intro}, we visualize the per-step denoising residuals along each action dimension for two different skills learned with a skill-conditioned noise schedule.
When approaching the target object, rotational dimensions (yaw, roll) resolve first to align the end-effector orientation, while positional dimensions ($x$, $y$) converge later. In contrast, during grasping, the gripper actuation dimension commits early to secure a firm grasp on the object, while positional dimensions are refined later.
This skill-aware prioritization 
yields average success rate gains of $11.7$ percentage points (pp) on contact-rich tasks and $13.0$ pp on dexterous manipulation tasks over the most competitive baseline, SkillDiffuser~\cite{DP:SkillDiffuser}, as shown on the rightmost side of the figure.
Beyond improving generation quality, the geometry of the learned schedule naturally reveals the denoising budget required for each skill, thereby enabling adaptive allocation of denoising steps at inference time.

To this end, we propose the \textbf{Skill-Adaptive Noise Diffusion ($\ours$)} framework, designed to learn robust skills for contact-rich and dexterous manipulation tasks that demand fine-grained coordination across tightly coupled action dimensions.
Building on skill diffusion, a hierarchical skill-learning paradigm with a skill encoder and a diffusion-based decoder, we introduce a skill-conditioned noise schedule that modulates the noise injection across action dimensions based on the skill embedding. To prevent the schedule from collapsing into a skill-agnostic (fixed) one, we employ an auxiliary inverse schedule that reconstructs the skill embedding from the Legendre coefficients of the noise schedule's derivative. 
Furthermore, the learned noise schedule naturally encodes the denoising demand of each skill, which allows us to adaptively determine the number of denoising steps without additional training. We introduce a schedule-driven adaptive denoising mechanism, which allocates denoising steps per skill based on the geometry of the learned schedule. 
Specifically, we cluster skills by the Legendre coefficients of their corresponding noise schedules and identify, for each cluster, the minimal denoising budget that maintains reconstruction quality. At inference, we select the total number of denoising steps based on the inferred skill, yielding improved inference efficiency with negligible impact on task performance. 
The contributions of our work are summarized as follows: 
\begin{itemize}
    \item We propose the $\ours$ framework, skill diffusion augmented with a skill-conditioned noise schedule that modulates noise injection across action dimensions based on the skill embedding, paired with an auxiliary inverse schedule to prevent the schedule from collapsing.
    \item We introduce a schedule-driven adaptive denoising mechanism that assigns a distinct number of denoising steps for each skill based on the geometry of the learned noise schedule, enabling efficient inference without sacrificing task performance.
    \item Through extensive experiments on contact-rich and bimanual dexterous tasks, we show that $\ours$ consistently outperforms existing diffusion-based skill learning and planning baselines while accelerating inference.
\end{itemize}

\section{Related Work}
\noindent\textbf{Diffusion Policies.} Recently, diffusion models have been widely adopted for sequential decision-making problems. A straightforward approach is to directly model the policy as a diffusion process over actions~\cite{DP:DiffBC,DP:DP,DP:DiffQL}, with several extensions to offline reinforcement learning settings~\cite{DP:DiffQL,DP:CEP,DP:EDP}. Another line of work formulates diffusion as a trajectory optimization problem, performing planning by iteratively denoising state-action sequences~\cite{DP:Diffuser,DP:HD,DP:DV}. 
SkillDiffuser~\cite{DP:SkillDiffuser} and LDuS~\cite{DP:LDuS} incorporate diffusion into hierarchical skill-based frameworks to model long-horizon manipulation via latent skill representations.
Despite these advances, none of these approaches considers learning a skill-conditioned noise schedule. Inspired by adaptive noise schedules in image diffusion models~\cite{DM:VDM,DM:MuLAN}, we extend this idea to skill generation, enabling the diffusion process to adaptively modulate noise injection across action dimensions based on the skill embedding.

Meanwhile, several studies have explored accelerating the inference of diffusion policies. CP~\cite{DP:CP} and OneDP~\cite{DP:OneDP} distill diffusion policies into fewer-step models, while D3P~\cite{DP:D3P} learns to dynamically adjust the number of denoising steps to the current context via online adaptation. 
Our work achieves a similar effect  by leveraging the geometry of our skill-conditioned noise schedule to adaptively determine per-skill denoising steps, thereby addressing the varying denoising complexity across skills. This emerges as a favorable side-effect rather than an explicit design objective.

\noindent\textbf{Robotic Manipulation.} Robot learning has achieved remarkable progress on classical locomotion~\cite{dataset:d4rl,classic:dt} and manipulation benchmarks~\cite{dataset:metaworld,classic:l2m}. Building on this success, the community has increasingly turned toward more complex scenarios that demand richer physical interaction with the environment. This includes contact-rich manipulation~\cite{dataset:mimicgen,env:factory}, where the space of possible contacts varies with object geometry and pose, and dexterous manipulation~\cite{dataset:dexmimicgen,dataset:BiDexHands}, where multi-fingered hands must coordinate fine-grained motions to perform intricate manipulations. 
To address these challenges, FORGE~\cite{forge} introduces a force-aware manipulation policy for contact-rich tasks, while DexHandDiff~\cite{DP:DexHandDiff} proposes a diffusion-based planning framework for adaptive dexterous manipulation tasks via guidance from large language models.
Yet, these approaches address such tasks by introducing external signals, while adapting the diffusion process itself to individual skills remains underexplored.
In this work, we develop a skill-based framework broadly applicable to various robotic manipulation settings, proposing skill-adaptive noise diffusion for robust skill learning and schedule-driven adaptive denoising for efficient inference.


\section{Preliminaries}
\noindent\textbf{Notations.} Subscript $t$ denotes environment timesteps, and superscript $k$ denotes diffusion steps. To avoid ambiguity between the superscript and an exponent, we write the $n$-th power of $\sigma^{(k)}$ as $[\sigma^{(k)}]^n$.

\noindent\textbf{Diffusion Models.} 
 %
%
To capture the dependency across action dimensions, which a squashed diagonal Gaussian cannot express, recent works adopt diffusion models to model expressive action distributions~\cite{DP:DiffBC,DP:DD}.
The \textit{forward process} in diffusion maps an input $x$ into a noise-corrupted variable $x^k$ for $k \in [0,1]$ by gradually adding Gaussian noise, where $k=0$ corresponds to the clean input and $k=1$ corresponds to the fully noised version. By discretizing the diffusion step into $K$ steps of width $1/K$, the forward process factorizes into a Markov chain, 
\begin{equation}
    q(x^{0:1}) = 
    q(x^0) \prod_{i=1}^K q(x^{k(i)}|x^{s(i)}), 
    \quad 
    \text{where} \ k(i) = i / K \ \text{and} \ s(i) = (i - 1) / K.
\end{equation} 

The marginal distribution of $x^k$ is given by $q(x^k|x) = \mathcal{N}(x^k;\alpha^kx,[\sigma^{(k)}]^2 \bm{I})$ where $\alpha^k,\sigma^k \in \mathbb{R}_+$ are the noise schedule functions of $k$. Then, the diffusion model $\epsilon_\theta$ is trained to predict the noise from a noise-corrupted input via the denoising score matching (DSM) loss~\cite{DM:DDPM}:
\begin{equation} \label{eq:dm_loss}
    \mathcal{L}(\theta) = \expectation_{k \sim U[0,1],\epsilon \sim \mathcal{N}(0,\bm{I})} 
    \left[ \lambda^k \|\epsilon_\theta(\alpha^k x + \sigma^{k}\epsilon,k) - \epsilon\|^2_2\right],
\end{equation}
where $\lambda^k$ is a reweighting term. 
In practice, we adopt $v$-parameterization, where the model predicts $v^{k} = \alpha^{k} \epsilon - \sigma^{k} \tau$. For readability, we write the remainder of this section in terms of noise $\epsilon$ prediction.

\noindent\textbf{Skill Diffusion.} 
Given a dataset $\mathcal{D} = \{ d_i \}_{i \leq N}$, where each demonstration $d = \{(s_t, a_t)\}_{t \leq H}$ is a sequence of state-action pairs, hierarchical skill-based policy learning aims to extract latent skill representations $z \in \mathbb{R}^{d_z}$ that capture temporal behaviors (i.e., skills) in an unsupervised manner~\cite{sprl:spirl,sprl:fist}. Here, a skill is defined as a sequence of $h$ consecutive actions $\bm{a}=\{a_t,...,a_{t+h-1}\}$, and $\tau=(s_{t:t+h-1},a_{t:t+h-1})$ denotes the corresponding skill trajectory.
Recent works incorporate diffusion models for skill-based policy learning~\cite{DP:SkillDiffuser,DP:LDuS}, i.e., \textit{skill diffusion},  comprising a skill encoder $q_{\theta_\text{e}}(z|\tau)$ and a diffusion-based skill decoder $\epsilon_{\theta_\text{d}}(\bm{a}^k,k,z)$, optimized via
\begin{equation}
    \mathcal{L}(\theta_\text{e}, \theta_\text{d}) = \expectation_{\tau \sim \mathcal{D}, z \sim q_{\theta_\text{e}}(z|\tau)} 
    [ \underbrace{\| \epsilon_{\theta_\text{d}}(\bm{a}^k,k,z) - \epsilon \|_2^2}_{\text{DSM loss}}
    + \beta \underbrace{D_{\text{KL}}(q_{\theta_\text{e}}(z|\tau),p(z)) }_{\text{regularization loss}} ]
\end{equation}
where $\bm{a}^k$ denotes the noise-corrupted action sequence at diffusion step $k$, $D_{\text{KL}}$ is the Kullback--Leibler (KL) divergence, $p(z) = \mathcal{N}(0,\bm{I})$ is a standard Gaussian prior, and $\beta$ is a regularization coefficient. 
Depending on the implementation, the skill decoder reconstructs either the skill or the full skill trajectory, or alternatively predicts a state sequence with actions inferred via inverse dynamics.

\begin{figure*}[t] 
    \centering
    \includegraphics[width=\textwidth]{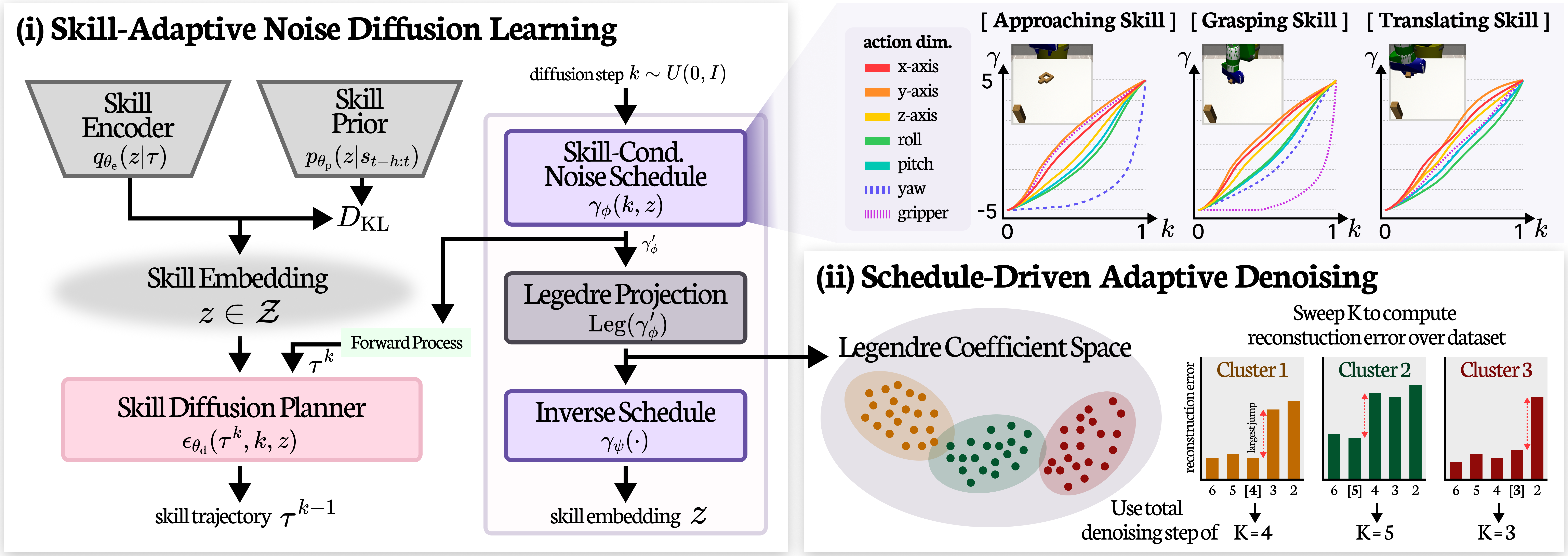}
    \caption{$\ours$ framework: In (\romannumeral 1), skills are learned via a diffusion-based hierarchical structure with a skill-conditioned noise schedule and an inverse schedule. The skill-conditioned noise schedule assigns action dimension-wise noise rates for each skill, capturing mutual dependencies among action dimensions. 
    In (\romannumeral 2), skills are clustered according to the Legendre coefficients of their noise schedule derivatives, and denoising budgets are assigned per cluster based on the smallest $K$  that maintains reconstruction quality. This enables skill-adaptive inference with reduced computational cost.}
    \label{fig:approach}
\end{figure*}
\section{$\ours$ Framework}

\subsection{Overall framework}
To reduce the variational gap by implicitly embedding mutual dependencies among action dimensions into the noise schedule, 
we present $\ours$, a framework that consists of two phases: (\romannumeral 1) skill-adaptive noise diffusion learning and (\romannumeral 2) schedule-driven adaptive denoising at inference, as illustrated in Figure~\ref{fig:approach}. 
In (\romannumeral 1), we extend the skill diffusion framework with a skill-conditioned noise schedule that adaptively adjusts the noise injection rates across action dimensions based on the skill embedding. To ensure that the learned noise schedule captures skill-specific structure, we introduce an inverse schedule as an auxiliary module, which provides a supervisory signal by recovering the skill embedding from the Legendre coefficients of the learned schedule's derivative.
In (\romannumeral 2), prior to deployment, we cluster skills by their learned noise schedules and sweep over the number of denoising steps to identify the minimal value that maintains reconstruction quality for each cluster. At deployment, the inferred skill embedding is mapped to its nearest cluster, from which the corresponding number of denoising steps is retrieved. This enables skill-aware compute allocation, where contact-rich skills that demand precision receive more denoising steps, while routine skills use fewer.

\subsection{Skill-Adaptive Noise Diffusion}
We employ skill diffusion, which encodes an $h$-length skill trajectory $\tau=(s_{t:t+h-1},a_{t:t+h-1})$ into a skill embedding $z$. The skill diffusion framework consists of three components: a skill encoder $q_{\theta_\text{e}}$ that maps the skill trajectory to the skill embedding, a skill prior $p_{\theta_\text{p}}$ that predicts the skill embedding from the past states, and a skill diffusion planner $\epsilon_{\theta_\text{d}}$ that reconstructs the skill trajectory, i.e.,
%
%
\begin{equation}
    q_{\theta_\text{e}}: \tau \mapsto z \in \mathbb{R}^{d_z}, 
    \quad 
    p_{\theta_\text{p}}: s_{t-h:t} \mapsto z, 
    \quad
    \epsilon_{\theta_\text{d}}: (\tau^k,k,z) \mapsto \epsilon.
\end{equation}

%
\begin{algorithm}[t]
    \begin{algorithmic}[1]
    \STATE{\textbf{Inputs}: dataset $\mathcal{D}$, skill encoder $q_{\theta_\text{e}}$, skill prior $p_{\theta_\text{p}}$, skill diffusion planner $\epsilon_{\theta_\text{d}}$, skill-conditioned noise schedule $\gamma_\phi$, inverse schedule $\gamma_\psi$}
    \WHILE{not converged}
        \STATE{$(\tau,s_{t-h:t}) \sim \mathcal{D}$; $k \sim \mathcal{U}[0, 1]$; $\epsilon \sim \mathcal{N}(0, \bm{I})$}
        \STATE{$z \sim q_{\theta_\text{e}}(\cdot|\tau)$; $\gamma^k \leftarrow \gamma_\phi(k, z)$}
        \STATE{$\alpha^k \leftarrow \sqrt{\text{sigmoid}(-\gamma^k)}$; $\sigma^k \leftarrow \sqrt{\text{sigmoid}(\gamma^k)}$; $\tau^k \leftarrow \alpha^k \tau + \sigma^k \epsilon$} \hfill $\triangleright$ Forward diffusion
        \STATE{Update $q_{\theta_\text{e}}, \epsilon_{\theta_\text{d}}, \gamma_\phi, \gamma_\psi$ using $\mathcal{L}_{\text{du}} + \eta \mathcal{L}_{\text{inv}}$ in~\eqref{loss:diffusion} and~\eqref{loss:inverse}}  \hfill $\triangleright$ $\eta >0$ is a weight for $\mathcal{L}_{\text{inv}}$ 
        \STATE{Update $p_{\theta_\text{p}}$ using $\mathcal{L}_{\text{pri}}$ in~\eqref{loss:prior}}
    \ENDWHILE
    \end{algorithmic}
    \caption{Skill-Adaptive Noise Diffusion} 
    \label{algo:SaND}
\end{algorithm}
%
\noindent\textbf{Skill-Conditioned Noise Schedule.} Due to the limited capability of a fixed noise schedule to capture skill-specific dependencies among action dimensions, as discussed in Section~\ref{sec:intro},
we extend the skill diffusion framework into a skill-adaptive noise diffusion framework that adaptively adjusts the noise injection rates across action dimensions based on the skill embedding $z$. To this end, we introduce a trainable noise schedule $\gamma_\phi$ that maps a diffusion step $k$ and the skill embedding $z$ to an $N$-dimensional vector, where $N$ is the action dimensionality. 
We implement the noise schedule as a monotonic fifth-degree polynomial following MuLAN~\cite{DM:MuLAN}, with fixed boundary values $\gamma_\phi(0,z)=\gamma_\text{min}$ and $\gamma_\phi(1,z)=\gamma_\text{max}$. Concretely, it is factorized as $\gamma_\phi(k, z) = f(k; g_\phi(z))$, where $g_\phi$ predicts polynomial coefficients and $f$ evaluates the corresponding polynomial at $k$. 
%
Then, $\gamma_\phi(k,z)$ is passed through a sigmoid function to yield the variance schedule parameters $\alpha^k$ and $\sigma^k$ as follows:
\begin{equation}
\begin{gathered}
    g_\phi: \mathbb{R}^{d_z} \rightarrow \mathbb{R}^{6 \times N}, \quad 
    f(k,C) = \sum_{n=0}^{5} C_n k^n, \quad
    \gamma_\phi(k,z) = f(k; g_\phi(z)) \in \mathbb{R}^{N}, 
    \\
    \alpha^k = \sqrt{\text{sigmoid}(-\gamma_\phi(k,z))}, \quad 
    \sigma^k = \sqrt{\text{sigmoid}(\gamma_\phi(k,z))},
\end{gathered}
\end{equation}
where $C_n \in \mathbb{R}^{N}$ is an $n$-degree coefficient vector across all action dimensions.
%
Note that we learn a noise schedule only over the action dimensions, rather than the full state-action space. State dimensions use a fixed noise schedule. 
%
Then, the training objective of the skill-adaptive noise diffusion is defined as
\begin{equation} \label{loss:diffusion}
    \mathcal{L}_{\text{du}}({\theta_\text{e}},{\theta_\text{d}},{\phi}) 
    = \underset{k \sim U[0,1], \\ \epsilon \sim \mathcal{N}(0,\bm{I})}
    {\expectation} 
    \left[\gamma'_\phi(k,z)\|\epsilon_{\theta_\text{d}}(\tau^k,k,\text{sg}(z))-\epsilon\|^2_2\right] 
    \! + \!
    \beta D_{\mathrm{KL}}\left(q_{\theta_\text{e}}(z|\tau),p(z)\right),
\end{equation}
where $\text{sg}(\cdot)$ is the stop-gradient operator and $\gamma'_{\phi} = \partial_{k}\gamma_{\phi}$ serves as the reweighting term. 
%
We apply a stop-gradient to $z$ in the DSM loss to block the direct gradient path to the skill encoder, ensuring that gradients propagate through $\gamma_\phi$ to learn a skill-adaptive noise schedule. 
%

At inference time, the skill encoder $q_{\theta_\text{e}}(z|\tau)$ is unavailable since the skill trajectory $\tau$ is not observed. We therefore learn a skill prior $p_{\theta_\text{p}}$ that predicts the skill embedding from the past state sequence $s_{t-h:t}$, trained to match the skill encoder's posterior as
\begin{equation} \label{loss:prior}
    \mathcal{L}_{\text{pri}}(\theta_\text{p}) 
    = D_{\text{KL}}\left(q_{\theta_\text{e}}(z|\tau),p_{\theta_\text{p}}(z|s_{t-h:t})\right).
\end{equation}
 
\noindent\textbf{Inverse Schedule.} 
When transitioning between skills (e.g., from moving to grasping), the state observations barely change, causing representation collapse during learning. This in turn causes the noise schedule to degenerate into a fixed one. 
%
To prevent this, we introduce an inverse schedule $\gamma_\psi$ as an auxiliary module that reconstructs the skill embedding from the learned noise schedule.
Since the noise schedule cannot be fed into a model directly, we instead exploit the structural properties of its polynomial form. As a monotone function with fixed boundary values, the noise schedule is fully characterized by its derivative. We therefore project $\gamma'_\phi$ onto a Legendre polynomial basis, whose orthogonality over the derivative space provides a natural discriminative representation of schedule shapes.
Then, the loss for the inverse schedule is defined as
\begin{equation} \label{loss:inverse}
    \mathcal{L}_{\text{inv}}(\phi, \psi) 
    = \underset{}{\expectation} \left[ \|\gamma_\psi(\text{Leg}(\gamma'_\phi(k,z))) - z\|^2_2 \right], 
\end{equation}
where $\text{Leg}(\cdot)$ denotes the projection coefficients onto the Legendre basis (Legendre coefficients). 
In practice, rather than regressing $z$ directly, we regress to a prototype skill embedding from a codebook, similar to VQ-VAE~\cite{vqvae}. Details and pseudocode are provided in Appendix~\ref{appendix:B:ours} and Algorithm~\ref{algo:SaND}.

\subsection{Schedule-Driven Adaptive Denoising} \label{sec:method:adaptive_denoising}
In sequential decision-making tasks, not all contexts require the same number of denoising steps at inference~\cite{DP:D3P}. Some skills demand precise control, while others can be resolved with fewer denoising steps. As illustrated in Figure~\ref{fig:intro}, our skill-conditioned noise schedule naturally captures this variation: different skills exhibit distinct convergence behavior across action dimensions, suggesting that the learned schedule encodes the denoising demand of each skill. We therefore use the geometry of the learned noise schedule as a proxy for determining the per-skill denoising budget.


Specifically, we iterate over the dataset to obtain the Legendre coefficients $\text{Leg}(\gamma'_\phi(k,z))$ of the learned noise schedule's derivative across action dimensions, and apply K-means clustering to group skills by their noise schedule geometry. %
%
%
%
We sweep the total number of denoising steps $K$ in decreasing order, computing the reconstruction MSE over the skills within each cluster at each $K$. The optimal $K$ is determined as the smallest value before a sharp increase in MSE, identified via the Kneedle algorithm~\cite{kneedle_algo}. 
At inference, given a skill embedding $z$ inferred from the skill prior $p_{\theta_\text{p}}$, we compute $\text{Leg}(\gamma'_\phi(k,z))$, identify its nearest cluster, and retrieve the corresponding number of denoising steps.
Since $\gamma_\phi$ is implemented as a 4-layer MLP, this lookup introduces negligible overhead relative to the cost of additional denoising steps.
See Algorithm~\ref{alg:adaptive_denoising} in Appendix~\ref{appendix:B:ours} for the full procedure.

\section{Experiments}
\subsection{Experiment Settings}
\noindent\textbf{Benchmarks.} We evaluate our method on two robot simulation platforms: Robosuite~\cite{sim:robosuite} and Isaac Gym~\cite{sim:isaacgym}. For Robosuite, we consider two benchmark datasets targeting \textit{contact-rich} and \textit{bimanual dexterous} manipulation tasks. Specifically, \textbf{MimicGen}~\cite{dataset:mimicgen} focuses on single-arm manipulation with a Panda robot, while \textbf{DexMimicGen}~\cite{dataset:dexmimicgen} targets dual-arm manipulation using either Panda robots equipped with dexterous hands or a GR-1 humanoid. We train on the publicly available 500 demonstrations per task from each benchmark, and report the task success rate (\%). 
For Isaac Gym, we use benchmark environments from \textbf{Bi-DexHands}~\cite{dataset:BiDexHands}, which features \textit{bimanual dexterous} manipulation tasks. We train on 100 demonstrations per task, and report the accumulated reward.
%

\noindent\textbf{Baselines.} For comparison, we consider several diffusion-based planning and hierarchical skill learning baselines.
(\romannumeral 1) \textbf{BC-RNN}~\cite{dataset:robomimic} serves as a basic baseline, employing a recurrent neural network. 
(\romannumeral 2) \textbf{DiffBC}~\cite{DP:DiffBC} models the policy directly as a diffusion process over actions.
(\romannumeral 3) \textbf{DP}~\cite{DP:DP} predicts a chunk of future actions as a single diffusion process.
(\romannumeral 4) \textbf{OneDP}~\cite{DP:OneDP} distills a pretrained diffusion policy into a one-step action generator via score distillation.
(\romannumeral 5) \textbf{Diffuser}~\cite{DP:Diffuser} performs planning by iteratively denoising full state-action trajectories via a diffusion model.
(\romannumeral 6) \textbf{DV}~\cite{DP:DV} is a diffusion-based planning framework that consolidates key design principles from modern diffusion planners. 
(\romannumeral 7) \textbf{SkillDiffuser}~\cite{DP:SkillDiffuser} is a hierarchical skill framework with a diffusion decoder that generates state trajectories, with actions inferred via an inverse dynamics model. 
(\romannumeral 8) \textbf{LDuS}~\cite{DP:LDuS} is a diffusion-based hierarchical skill framework that generates full state-action trajectories. 
%
All diffusion-based baselines use a fixed cosine schedule, whereas $\ours$ employs a learnable skill-conditioned noise schedule.
%
%
For the diffusion-based baselines, we report the number of denoising steps $K$ required for each skill or action generation at inference.


\subsection{Main Results}
\begin{table}[t]
\caption{Performance on Robosuite benchmarks: All baselines and $\ours$ are evaluated on 5 different random seeds, and success rate (\%) is reported in 95\% confidence interval.}
\label{tab:main:1}
\centering
\begin{adjustbox}{width=.98\textwidth}
\begin{tabular}{l|c|
>{\centering\arraybackslash}p{0.1\linewidth}
>{\centering\arraybackslash}p{0.1\linewidth}
>{\centering\arraybackslash}p{0.1\linewidth}
>{\centering\arraybackslash}p{0.1\linewidth}
>{\centering\arraybackslash}p{0.1\linewidth}
>{\centering\arraybackslash}p{0.1\linewidth}
|
>{\centering\arraybackslash}p{0.1\linewidth}}
\toprule
\multicolumn{9}{l}{\textbf{Benchmark: Contact-Rich Manipulation (MimicGen~\cite{dataset:mimicgen}) -- Low Dimensional Observation}}  \\
\midrule
\textbf{Methods} & \textbf{$K$} & \textbf{Square} & \textbf{Threading} & \textbf{Mug} & \textbf{Hammer} & \textbf{Coffee} & \textbf{Kitchen} & \textbf{Average} \\
\midrule
BC-RNN & -
& 53.5{\color{gray}\scriptsize$\pm$5.5} & 42.0{\color{gray}\scriptsize$\pm$11.4} 
& 55.2{\color{gray}\scriptsize$\pm$5.6} & 48.2{\color{gray}\scriptsize$\pm$3.6}
& 52.2{\color{gray}\scriptsize$\pm$7.5} & 52.2{\color{gray}\scriptsize$\pm$12.0} 
& 50.5{\color{gray}\scriptsize$\pm$2.8}  \\
DiffBC & 10
& 44.3{\color{gray}\scriptsize$\pm$1.6}     & 57.6{\color{gray}\scriptsize$\pm$1.4} 
& 43.1{\color{gray}\scriptsize$\pm$5.9}     & 37.5{\color{gray}\scriptsize$\pm$2.8}
& 68.2{\color{gray}\scriptsize$\pm$10.9}    & 28.6{\color{gray}\scriptsize$\pm$4.1} 
& 46.5{\color{gray}\scriptsize$\pm$1.1}      \\
DP & 10
& 50.6{\color{gray}\scriptsize$\pm$2.6}     & 48.8{\color{gray}\scriptsize$\pm$3.5} 
& \textbf{62.1{\color{gray}\scriptsize$\pm$3.6}}     & 50.1{\color{gray}\scriptsize$\pm$1.1}
& 47.8{\color{gray}\scriptsize$\pm$9.2}    & 49.2{\color{gray}\scriptsize$\pm$2.9} 
& 51.4{\color{gray}\scriptsize$\pm$1.4}      \\
OneDP & 1
& 47.1{\color{gray}\scriptsize$\pm$2.8}     & 45.6{\color{gray}\scriptsize$\pm$3.0} 
& 59.8{\color{gray}\scriptsize$\pm$6.6}     & 47.1{\color{gray}\scriptsize$\pm$2.3}
& 44.4{\color{gray}\scriptsize$\pm$10.6}    & 41.4{\color{gray}\scriptsize$\pm$4.0} 
& 47.6{\color{gray}\scriptsize$\pm$2.1}      \\
Diffuser & 10
& 45.6{\color{gray}\scriptsize$\pm$1.5}     & 31.5{\color{gray}\scriptsize$\pm$6.9} 
& 54.2{\color{gray}\scriptsize$\pm$6.2}     & 47.0{\color{gray}\scriptsize$\pm$0.7}
& 45.4{\color{gray}\scriptsize$\pm$7.7}     & 47.0{\color{gray}\scriptsize$\pm$4.0} 
& 45.1{\color{gray}\scriptsize$\pm$3.0}      \\
DV & 10
& 58.8{\color{gray}\scriptsize$\pm$2.3} & 44.4{\color{gray}\scriptsize$\pm$3.7}
& 57.3{\color{gray}\scriptsize$\pm$4.1} & 50.0{\color{gray}\scriptsize$\pm$4.1}
& 63.2{\color{gray}\scriptsize$\pm$13.8} & 41.0{\color{gray}\scriptsize$\pm$3.1}
& 52.5{\color{gray}\scriptsize$\pm$3.6} \\
SkillDiffuser & 10
& 53.6{\color{gray}\scriptsize$\pm$1.6} & 40.9{\color{gray}\scriptsize$\pm$6.5} 
& 58.3{\color{gray}\scriptsize$\pm$3.4} & 47.5{\color{gray}\scriptsize$\pm$2.7}
& 69.0{\color{gray}\scriptsize$\pm$9.5} & 44.2{\color{gray}\scriptsize$\pm$3.4} 
& 52.3{\color{gray}\scriptsize$\pm$1.9} \\
LDuS & 10
& 44.7{\color{gray}\scriptsize$\pm$2.5} & 42.7{\color{gray}\scriptsize$\pm$0.7} 
& 57.6{\color{gray}\scriptsize$\pm$3.6} & 47.4{\color{gray}\scriptsize$\pm$2.8}
& 58.7{\color{gray}\scriptsize$\pm$3.6} & 46.4{\color{gray}\scriptsize$\pm$2.7} 
& 49.6{\color{gray}\scriptsize$\pm$0.5}  \\
\midrule
\rowcolor{gray!20} \textbf{$\ours$} & 10
& {60.9{\color{gray}\scriptsize$\pm$1.9}} 
& {66.9{\color{gray}\scriptsize$\pm$5.5}} 
& {60.0{\color{gray}\scriptsize$\pm$1.6}} 
& {58.4{\color{gray}\scriptsize$\pm$1.9}}
& \textbf{82.4{\color{gray}\scriptsize$\pm$2.2}} 
& \textbf{55.3{\color{gray}\scriptsize$\pm$4.4}} 
& \textbf{64.0{\color{gray}\scriptsize$\pm$1.0}}  \\
\rowcolor{gray!20} \textbf{$\ours$+AD} & 3.7
& \textbf{61.7{\color{gray}\scriptsize$\pm$3.2}} 
& \textbf{66.9{\color{gray}\scriptsize$\pm$3.9}} 
& 55.9{\color{gray}\scriptsize$\pm$5.0} 
& \textbf{59.6{\color{gray}\scriptsize$\pm$2.2}}
& 81.0{\color{gray}\scriptsize$\pm$2.8} 
& 54.6{\color{gray}\scriptsize$\pm$3.6} 
& 63.3{\color{gray}\scriptsize$\pm$0.8}  \\
\midrule \midrule
\multicolumn{9}{l}{\textbf{Benchmark: Bimanual Dexterous Manipulation (DexMimicGen~\cite{dataset:dexmimicgen}) -- Multimodal Observation}}  \\
\midrule
\textbf{Methods} & \textbf{$K$} & \textbf{Box} & \textbf{Drawer} & \textbf{Tray} & \textbf{Pouring} & \textbf{Coffee} & \textbf{Can} & \textbf{Average} \\
\midrule
BC-RNN & -
& 42.6{\color{gray}\scriptsize$\pm$7.3} & 61.8{\color{gray}\scriptsize$\pm$7.4} 
& 14.2{\color{gray}\scriptsize$\pm$4.4} & 65.2{\color{gray}\scriptsize$\pm$8.6}
& 27.7{\color{gray}\scriptsize$\pm$2.7} & 62.4{\color{gray}\scriptsize$\pm$5.0} 
& 45.6{\color{gray}\scriptsize$\pm$3.4}  \\
DiffBC & 10
& 47.6{\color{gray}\scriptsize$\pm$10.7} & 79.8{\color{gray}\scriptsize$\pm$4.2}
& 12.0{\color{gray}\scriptsize$\pm$7.9} & 27.0{\color{gray}\scriptsize$\pm$10.0}
& 10.0{\color{gray}\scriptsize$\pm$6.0} & 9.2{\color{gray}\scriptsize$\pm$7.8} 
& 30.9{\color{gray}\scriptsize$\pm$3.4}  \\
DP & 10
& 35.2{\color{gray}\scriptsize$\pm$8.2} & 49.6{\color{gray}\scriptsize$\pm$5.1} 
& 25.4{\color{gray}\scriptsize$\pm$4.6} & 12.0{\color{gray}\scriptsize$\pm$7.8}
& 19.4{\color{gray}\scriptsize$\pm$9.4} & 40.0{\color{gray}\scriptsize$\pm$17.5} 
& 30.3{\color{gray}\scriptsize$\pm$3.9} \\
OneDP & 1
& 23.8{\color{gray}\scriptsize$\pm$6.3} & 39.6{\color{gray}\scriptsize$\pm$3.4} 
& 20.6{\color{gray}\scriptsize$\pm$8.6} & 0.0{\color{gray}\scriptsize$\pm$0.0}
& 19.0{\color{gray}\scriptsize$\pm$7.2} & 22.2{\color{gray}\scriptsize$\pm$9.4} 
& 20.9{\color{gray}\scriptsize$\pm$1.0} \\
Diffuser & 10
& 59.6{\color{gray}\scriptsize$\pm$7.3} & 70.4{\color{gray}\scriptsize$\pm$6.9} 
& 26.4{\color{gray}\scriptsize$\pm$9.2} & 50.4{\color{gray}\scriptsize$\pm$4.4}
& 14.6{\color{gray}\scriptsize$\pm$4.1} & 47.4{\color{gray}\scriptsize$\pm$6.2} 
& 44.8{\color{gray}\scriptsize$\pm$2.2}  \\
DV & 10
& 71.6{\color{gray}\scriptsize$\pm$9.6} & 56.4{\color{gray}\scriptsize$\pm$9.1}
& 14.8{\color{gray}\scriptsize$\pm$4.7} & 68.8{\color{gray}\scriptsize$\pm$10.2}
& 26.6{\color{gray}\scriptsize$\pm$5.7} & 38.0{\color{gray}\scriptsize$\pm$16.5}
& 46.0{\color{gray}\scriptsize$\pm$5.6} \\
SkillDiffuser & 10
& 68.6{\color{gray}\scriptsize$\pm$8.0} & 66.2{\color{gray}\scriptsize$\pm$5.7} 
& 38.6{\color{gray}\scriptsize$\pm$7.8} & 57.0{\color{gray}\scriptsize$\pm$11.9}
& 24.0{\color{gray}\scriptsize$\pm$4.2} & 52.4{\color{gray}\scriptsize$\pm$14.9} 
& 51.1{\color{gray}\scriptsize$\pm$4.3}  \\
LDuS & 10
& 57.6{\color{gray}\scriptsize$\pm$9.4} & 76.6{\color{gray}\scriptsize$\pm$8.6} 
& 35.2{\color{gray}\scriptsize$\pm$4.7} & 35.4{\color{gray}\scriptsize$\pm$3.7}
& 25.8{\color{gray}\scriptsize$\pm$4.3} & 52.0{\color{gray}\scriptsize$\pm$10.4} 
& 47.1{\color{gray}\scriptsize$\pm$3.4} \\
\midrule
\rowcolor{gray!20} \textbf{$\ours$} & 10
& {79.4{\color{gray}\scriptsize$\pm$5.5}} 
& \textbf{81.0{\color{gray}\scriptsize$\pm$1.4}}
& {40.6{\color{gray}\scriptsize$\pm$6.0}} 
& \textbf{71.0{\color{gray}\scriptsize$\pm$13.3}}
& \textbf{32.0{\color{gray}\scriptsize$\pm$4.6}} 
& {80.8{\color{gray}\scriptsize$\pm$8.7}}
& \textbf{64.1{\color{gray}\scriptsize$\pm$2.7}} \\

\rowcolor{gray!20} \textbf{$\ours$+AD} & 5.4
& \textbf{83.6{\color{gray}\scriptsize$\pm$4.9}} 
& {80.6{\color{gray}\scriptsize$\pm$4.8}}
& \textbf{42.0{\color{gray}\scriptsize$\pm$7.8}} 
& 63.6{\color{gray}\scriptsize$\pm$10.8}
& 30.8{\color{gray}\scriptsize$\pm$5.3} 
& \textbf{82.2{\color{gray}\scriptsize$\pm$8.9}} 
& {63.8{\color{gray}\scriptsize$\pm$1.9}} \\
\bottomrule
\end{tabular}
\end{adjustbox}
\end{table}
%

\noindent\textbf{Performance on contact-rich manipulation tasks.} In the upper part of Table~\ref{tab:main:1}, we evaluate performance on six \textit{contact-rich} manipulation tasks in MimicGen using a single Panda robot equipped with a parallel-jaw gripper~\cite{dataset:mimicgen}. We consider a \textit{low-dimensional} observation space, which consists of the object state and the proprioceptive state of the robot. As shown, $\ours$ yields the best performance on all tasks except Mug, outperforming the most competitive baseline SkillDiffuser by $11.7$ pp in average success rate. 
Furthermore, $\ours$ with schedule-driven adaptive denoising, denoted as $\ours$+AD, reduces the total number of denoising steps $K$ by $63.2\%$ while exhibiting only a negligible degradation of $0.7$ pp in average performance. In some cases, $\ours$+AD even achieves better performance than $\ours$, as it automatically skips under-optimized denoising regions and finds solutions with more plausible actions. 
%
%
Note that $K$ varies across tasks and skills for $\ours$+AD; we report the average here, with per-task $K$ provided in Table~\ref{tab:ablation:2} in Appendix~\ref{sec:appendix:c:ablation_ad}. 

\noindent\textbf{Performance on bimanual dexterous manipulation tasks.} In the lower part of Table~\ref{tab:main:1}, we evaluate performance on six \textit{bimanual dexterous} manipulation tasks in DexMimicGen using dual Panda robots equipped with dexterous hands (first three tasks) and the GR-1 humanoid (last three tasks). Here, we consider a \textit{multimodal} observation space, which consists of three RGB images from an agent-view camera and wrist cameras on both hands, along with the proprioceptive state of the robot. 
As shown, $\ours$ yields the best performance, outperforming the strongest baseline, SkillDiffuser, by $13.0$ pp in average success rate. Furthermore, $\ours$+AD reduces $K$ by $45.6\%$ while maintaining comparable performance to $\ours$.
Additionally, Table~\ref{tab:main:2} reports performance on the Bi-DexHands benchmark in Isaac Gym using two Shadow dexterous hands with floating bases, under a \textit{low-dimensional} observation space. Similarly, $\ours$ consistently achieves the best performance across all tasks.  
%

In these experiments, DiffBC exhibits the lowest performance, as it lacks temporal modeling. Planning-based (Diffuser, DV) and skill-based (SkillDiffuser, LDuS) baselines improve performance by incorporating temporal dependencies or skill abstractions, respectively. 
Meanwhile, OneDP reduces the denoising steps of DP from $10$ to $1$ through distillation, yet loses considerable performance relative to DP, particularly on more complex dexterous manipulation tasks.
Yet, all baselines plateau beyond a certain performance level, suggesting that their fixed noise schedule fundamentally limits their capability to handle the complexity of contact-rich and dexterous manipulation tasks. 
$\ours$ outperforms the baselines by employing a skill-conditioned noise schedule along with an inverse schedule for auxiliary supervision. Such a design ensures that different skills are diffused at distinct noise levels across action dimensions, enabling more robust skill trajectory generation. Furthermore, our schedule-driven adaptive denoising reduces the number of denoising steps $K$ by assigning skill-specific denoising budgets according to the geometry of the noise schedule, with minimal performance degradation or even marginal gains in some cases.

\begin{table}[t]
\caption{Performance on Isaac Gym benchmarks: All baselines and $\ours$ are evaluated on 3 different random seeds, and accumulated reward is reported in 95\% confidence interval.}
\label{tab:main:2}
\centering
\begin{adjustbox}{width=.98\textwidth}
\begin{tabular}{l|c|
>{\centering\arraybackslash}p{0.1\linewidth}
>{\centering\arraybackslash}p{0.1\linewidth}
>{\centering\arraybackslash}p{0.1\linewidth}
>{\centering\arraybackslash}p{0.1\linewidth}
>{\centering\arraybackslash}p{0.1\linewidth}
>{\centering\arraybackslash}p{0.1\linewidth}
|
>{\centering\arraybackslash}p{0.1\linewidth}}
\toprule
\multicolumn{9}{l}{\textbf{Benchmark: Bimanual Dexterous Manipulation (Bi-DexHands~\cite{dataset:BiDexHands}) -- Low Dimensional Observation}}  \\
\midrule
\textbf{Methods} & \textbf{$K$} & \textbf{Catch} & \textbf{Door} & \textbf{Lift} & \textbf{Scissors} & \textbf{Pen} & \textbf{Cup} & \textbf{Average} \\
\midrule
BC-RNN & -
& 21.9{\color{gray}\scriptsize$\pm$0.6} 
& 329.8{\color{gray}\scriptsize$\pm$49.1}
& 142.3{\color{gray}\scriptsize$\pm$97.3} 
& 293.6{\color{gray}\scriptsize$\pm$91.6}
& 79.3{\color{gray}\scriptsize$\pm$25.4} 
& 400.5{\color{gray}\scriptsize$\pm$130.3}
& 211.2{\color{gray}\scriptsize$\pm$17.0} \\
DiffBC & 5
& 18.8{\color{gray}\scriptsize$\pm$2.2} 
& 362.0{\color{gray}\scriptsize$\pm$23.4}
& 235.8{\color{gray}\scriptsize$\pm$26.9} 
& 535.6{\color{gray}\scriptsize$\pm$40.3}
& 171.5{\color{gray}\scriptsize$\pm$3.7} 
& 514.7{\color{gray}\scriptsize$\pm$19.5}
& 306.4{\color{gray}\scriptsize$\pm$6.8} \\
SkillDiffuser & 5
& 14.7{\color{gray}\scriptsize$\pm$0.6} 
& 394.1{\color{gray}\scriptsize$\pm$18.4}
& 95.1{\color{gray}\scriptsize$\pm$30.2} 
& 570.1{\color{gray}\scriptsize$\pm$11.5}
& 147.4{\color{gray}\scriptsize$\pm$24.3} 
& 564.0{\color{gray}\scriptsize$\pm$26.0}
& 297.6{\color{gray}\scriptsize$\pm$8.7} \\
DV & 5
& 19.9{\color{gray}\scriptsize$\pm$0.7} 
& 349.9{\color{gray}\scriptsize$\pm$17.2}
& 241.8{\color{gray}\scriptsize$\pm$10.1} 
& 565.3{\color{gray}\scriptsize$\pm$25.0}
& 172.9{\color{gray}\scriptsize$\pm$1.0} 
& 529.9{\color{gray}\scriptsize$\pm$17.4}
& 313.3{\color{gray}\scriptsize$\pm$3.3} \\
\midrule
\rowcolor{gray!20} \textbf{$\ours$} & 5
& \textbf{24.0{\color{gray}\scriptsize$\pm$0.7}}
& \textbf{437.9{\color{gray}\scriptsize$\pm$21.1}}
& \textbf{242.8{\color{gray}\scriptsize$\pm$8.8}}
& \textbf{620.7{\color{gray}\scriptsize$\pm$10.7}}
& \textbf{175.3{\color{gray}\scriptsize$\pm$3.5}}
& \textbf{595.0{\color{gray}\scriptsize$\pm$16.6}}
& \textbf{349.3{\color{gray}\scriptsize$\pm$5.5}} \\
\bottomrule
\end{tabular}
\end{adjustbox}
\end{table}

\subsection{Ablation Study} \label{sec:exp:ablation}
\noindent\textbf{Ablation on skill-adaptive noise diffusion.}
Table~\ref{tab:ablation:1} shows an ablation study on the key components of our skill-adaptive noise diffusion. \textbf{$\ours$-Legendre} removes the Legendre projection $\text{Leg}(\cdot)$ applied to $\gamma'_\phi$ prior to the inverse schedule; \textbf{$\ours$-Inverse} further removes the entire inverse schedule $\gamma_\psi$; and \textbf{$\ours$-Schedule} removes the skill-conditioned noise schedule $\gamma_\phi$ entirely, falling back to a fixed noise schedule.
As shown, removing each component leads to a consistent drop in average performance, validating the design choices behind our skill-adaptive noise diffusion. Specifically, on MimicGen, $\ours$-Legendre, $\ours$-Inverse, and $\ours$-Schedule incur average success rate degradations of $2.7$ pp, $3.6$ pp, and $6.4$ pp, respectively. A similar trend is observed on DexMimicGen, with performance drops of $3.2$ pp, $3.4$ pp, and $7.1$ pp, confirming the necessity of all three components.
We further ablate the skill conditioning and the per-dimension design in Appendix~\ref{appendix:C:ablation_sand}.



\begin{table}[h]
\caption{Ablation on the key components of skill-adaptive noise diffusion}
\label{tab:ablation:1}
\centering
\begin{adjustbox}{width=.98\textwidth}
\begin{tabular}{l|
>{\centering\arraybackslash}p{0.1\linewidth}
>{\centering\arraybackslash}p{0.1\linewidth}
>{\centering\arraybackslash}p{0.1\linewidth}
>{\centering\arraybackslash}p{0.1\linewidth}
>{\centering\arraybackslash}p{0.1\linewidth}
>{\centering\arraybackslash}p{0.1\linewidth}
|
>{\centering\arraybackslash}p{0.1\linewidth}
}
\toprule
\multicolumn{8}{l}{\textbf{Benchmark: Contact-Rich Manipulation (MimicGen~\cite{dataset:mimicgen}) -- Low Dimensional Observation}}  \\
\midrule
\textbf{Methods} & \textbf{Square} & \textbf{Threading} & \textbf{Mug} & \textbf{Hammer} & \textbf{Coffee} & \textbf{Kitchen} & \textbf{Average} \\
\midrule
\textbf{$\ours$}
& \textbf{60.9{\color{gray}\scriptsize$\pm$1.9}} 
& \textbf{66.9{\color{gray}\scriptsize$\pm$5.5}} 
& \textbf{60.0{\color{gray}\scriptsize$\pm$1.6}} & \textbf{58.4{\color{gray}\scriptsize$\pm$1.9}}
& \textbf{82.4{\color{gray}\scriptsize$\pm$2.2}} & \textbf{55.3{\color{gray}\scriptsize$\pm$4.4}} 
& \textbf{64.0{\color{gray}\scriptsize$\pm$1.0}}  \\
$\ours$-Legendre
& 60.5{\color{gray}\scriptsize$\pm$2.3} & 61.3{\color{gray}\scriptsize$\pm$9.3} 
& 56.3{\color{gray}\scriptsize$\pm$6.5} & 56.2{\color{gray}\scriptsize$\pm$2.4}
& 81.6{\color{gray}\scriptsize$\pm$7.0} & 52.1{\color{gray}\scriptsize$\pm$8.9} 
& 61.3{\color{gray}\scriptsize$\pm$1.5}  \\
$\ours$-Inverse
& 56.9{\color{gray}\scriptsize$\pm$1.5} & 61.6{\color{gray}\scriptsize$\pm$7.7} 
& 58.4{\color{gray}\scriptsize$\pm$5.5} & 55.3{\color{gray}\scriptsize$\pm$2.3}
& 77.2{\color{gray}\scriptsize$\pm$8.5} & 53.2{\color{gray}\scriptsize$\pm$4.7} 
& 60.4{\color{gray}\scriptsize$\pm$1.7}  \\
$\ours$-Schedule
& 49.7{\color{gray}\scriptsize$\pm$7.0} & 61.5{\color{gray}\scriptsize$\pm$10.5} 
& 55.8{\color{gray}\scriptsize$\pm$5.2} & 55.4{\color{gray}\scriptsize$\pm$2.7}
& 73.6{\color{gray}\scriptsize$\pm$9.5} & 49.6{\color{gray}\scriptsize$\pm$5.9} 
& 57.6{\color{gray}\scriptsize$\pm$3.5}  \\
\midrule\midrule
\multicolumn{8}{l}{\textbf{Benchmark: Bimanual Dexterous Manipulation (DexMimicGen~\cite{dataset:dexmimicgen}) -- Multimodal Observation}}  \\
\midrule
\textbf{Methods} & \textbf{Box} & \textbf{Drawer} & \textbf{Tray} & \textbf{Pouring} & \textbf{Coffee} & \textbf{Can} & \textbf{Average} \\
\midrule
\textbf{$\ours$} 
& \textbf{79.4{\color{gray}\scriptsize$\pm$5.5}} & \textbf{81.0{\color{gray}\scriptsize$\pm$1.4}}
& \textbf{40.6{\color{gray}\scriptsize$\pm$6.0}} & \textbf{71.0{\color{gray}\scriptsize$\pm$13.3}}
& \textbf{32.0{\color{gray}\scriptsize$\pm$4.6}} & \textbf{80.8{\color{gray}\scriptsize$\pm$8.7}}
& \textbf{64.1{\color{gray}\scriptsize$\pm$2.7}} \\
$\ours$-Legendre 
& 74.0{\color{gray}\scriptsize$\pm$6.2} & 79.6{\color{gray}\scriptsize$\pm$4.5}
& 36.4{\color{gray}\scriptsize$\pm$14.6} & 63.0{\color{gray}\scriptsize$\pm$24.6}
& 31.4{\color{gray}\scriptsize$\pm$1.8} & \textbf{80.8{\color{gray}\scriptsize$\pm$12.0}}
& 60.9{\color{gray}\scriptsize$\pm$5.1} \\
$\ours$-Inverse 
& 73.4{\color{gray}\scriptsize$\pm$6.3} & 76.8{\color{gray}\scriptsize$\pm$6.4} 
& 38.8{\color{gray}\scriptsize$\pm$7.1} & 67.2{\color{gray}\scriptsize$\pm$14.9}
& 31.2{\color{gray}\scriptsize$\pm$7.5} & 76.6{\color{gray}\scriptsize$\pm$14.4} 
& 60.7{\color{gray}\scriptsize$\pm$3.8} \\
$\ours$-Schedule
& 68.0{\color{gray}\scriptsize$\pm$9.9} & 78.6{\color{gray}\scriptsize$\pm$4.8} 
& 38.2{\color{gray}\scriptsize$\pm$7.3} & 54.6{\color{gray}\scriptsize$\pm$19.4}
& 24.2{\color{gray}\scriptsize$\pm$6.7} & 78.6{\color{gray}\scriptsize$\pm$10.9} 
& 57.0{\color{gray}\scriptsize$\pm$4.0} \\
\bottomrule
\end{tabular}
\end{adjustbox}
\end{table}

\begin{figure}[t]
    \centering
    \settoheight{\imgheight}{\includegraphics[width=0.66\textwidth]{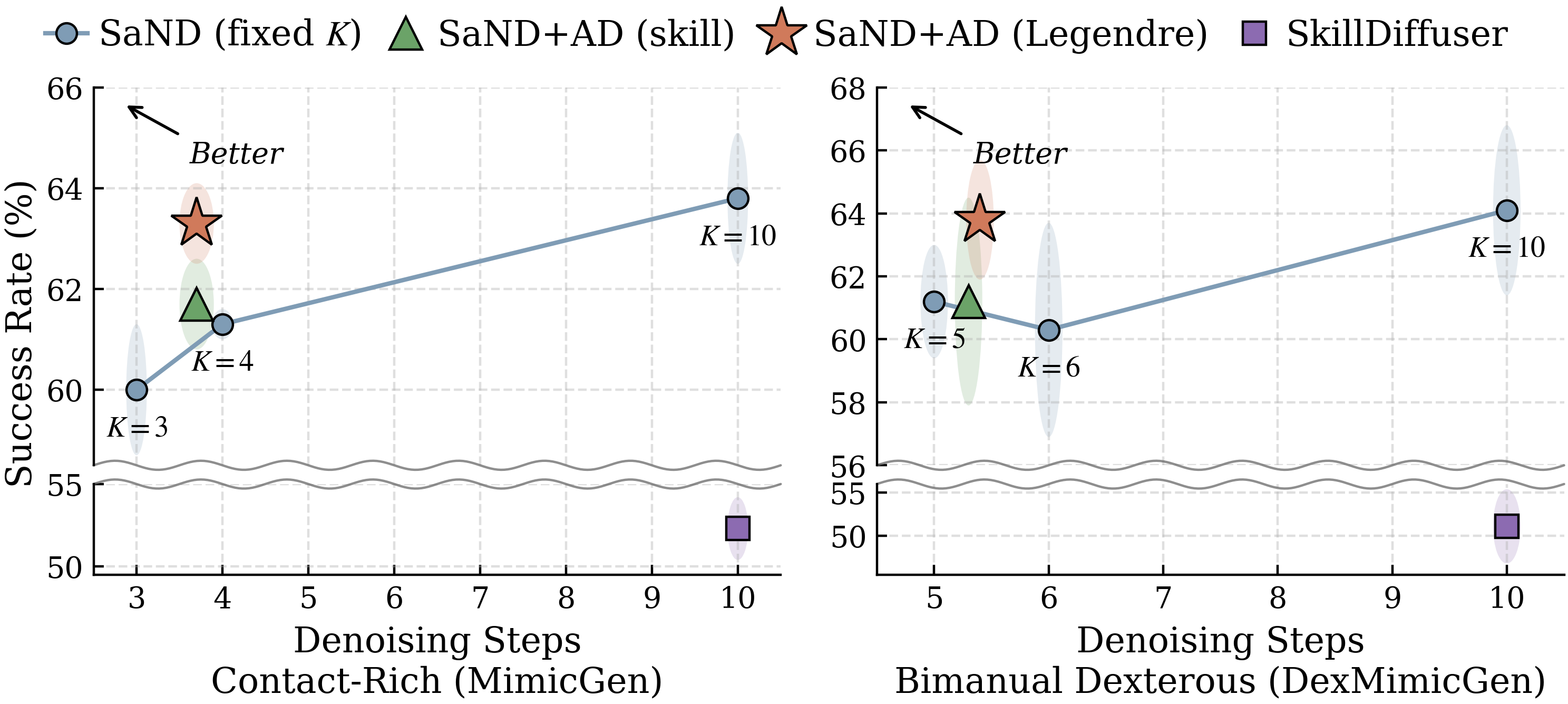}}
    \begin{minipage}{0.66\textwidth}
        \centering
        \includegraphics[width=\linewidth]{figures/ad_steps.png}
        \captionof{figure}{Ablation on schedule-driven adaptive denoising}
        \label{fig:ablation:2}
    \end{minipage}
    \hfill
    \begin{minipage}{0.32\textwidth}
        \centering
        \includegraphics[height=\imgheight]{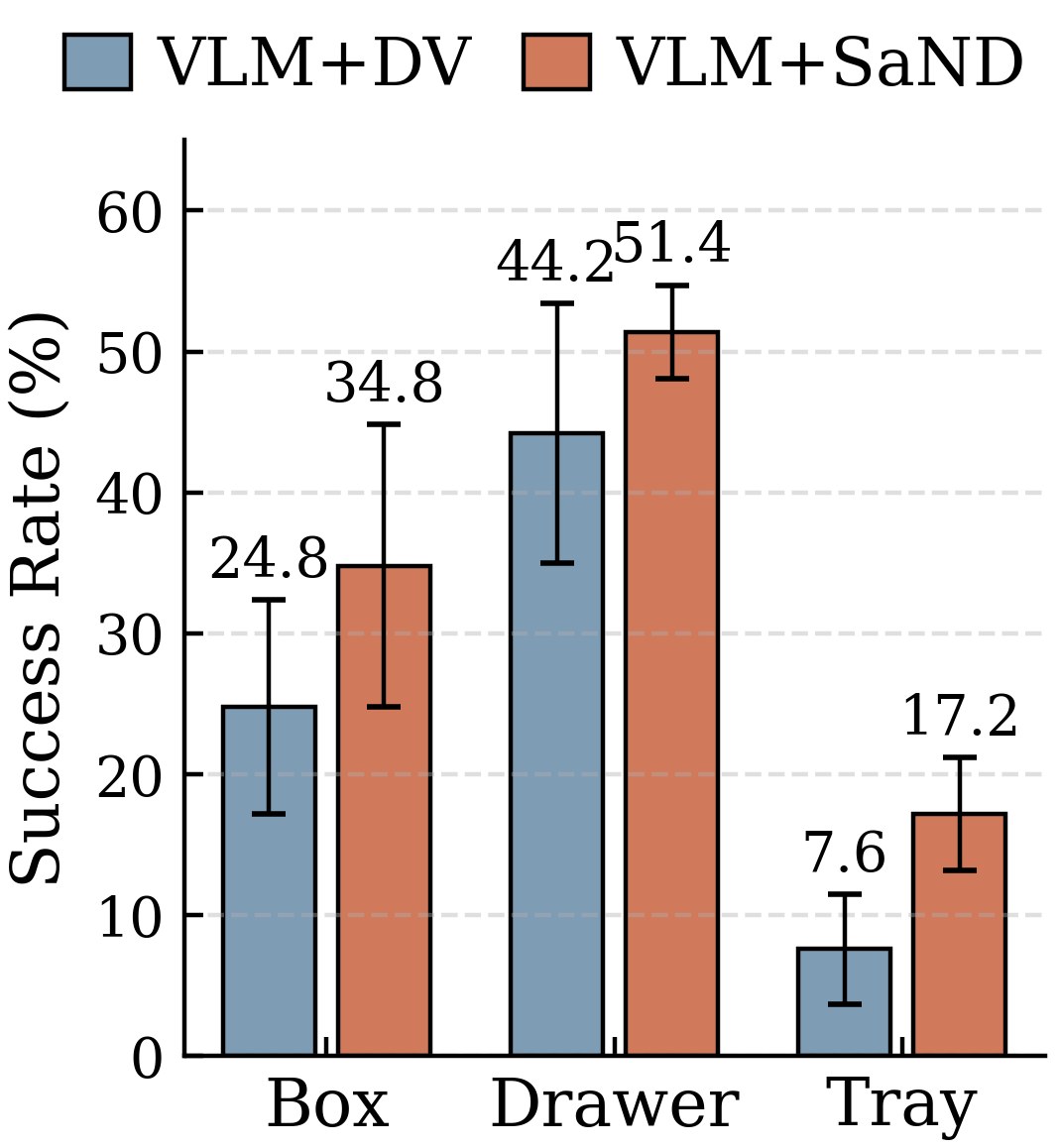}
        \captionof{figure}{Extension to VLA}
        \label{fig:case:vla}
    \end{minipage}
\end{figure}

\noindent\textbf{Ablation on schedule-driven adaptive denoising.} Figure~\ref{fig:ablation:2} depicts an ablation study on the schedule-driven adaptive denoising mechanism, aimed at addressing two questions. 
First, we examine whether the performance of $\ours$+AD could have been achieved by simply using a fixed $K$ without adaptive denoising (e.g., $K=3,4$ for MimicGen, $K=5,6$ for DexMimicGen).
Fixed $K$ fails to match $\ours$+AD for both benchmarks, while $K=10$ is required to achieve comparable performance, indicating that adaptive denoising reduces $K$ without performance loss.
Second, we assess whether schedule-based clustering yields better performance than skill embedding-based clustering for adaptive denoising, showing up to 3.7\% (2.6\% on average) improvement in success rate for MimicGen, and up to 8.2\% (4.3\% on average) improvement for DexMimicGen. Results on individual tasks are provided in Appendix~\ref{sec:appendix:c:ablation_ad}. 
This gain stems from skill embeddings capturing behavioral similarity rather than denoising complexity, whereas schedule-based clustering directly groups skills by their noise schedule geometry. Figure~\ref{fig:appendix:C:tsne_visualization} in Appendix~\ref{appendix:c:1} further supports this, showing that similar skill embeddings can be assigned to different noise schedule clusters due to the temporal continuity of skills, while remaining clearly distinguishable in the Legendre coefficient space.

\subsection{Case Study} \label{sec:exp:case}
\begin{figure*}[h]  
    \centering 
    \includegraphics[width=.98\textwidth]{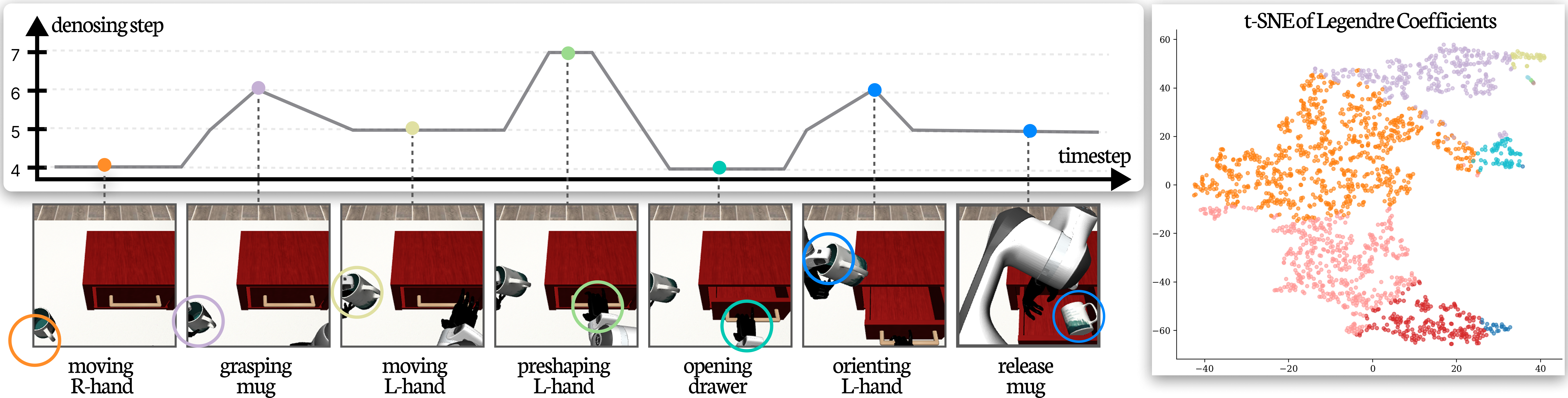}
    \caption{Visualization of schedule-driven adaptive denoising on the Drawer task}
    \label{fig:analysis:adaptive_denoising}
\end{figure*}

\noindent\textbf{Visualization of schedule-driven adaptive denoising.} Figure~\ref{fig:analysis:adaptive_denoising} illustrates how the adaptive denoising budget varies across skills on the Drawer task. The left panel plots the number of denoising steps used at each timestep, with colored dots marking representative skill snapshots shown below.
Colors correspond to clusters in the Legendre coefficient space (right panel), showing that skills with distinct denoising steps are well-separated into different clusters. Coarse motions such as arm reaching or drawer opening require as few as 4 denoising steps, whereas contact-rich skills such as grasping, preshaping, and orienting demand up to 6-7 steps. This indicates that our schedule-driven adaptive denoising effectively allocates denoising budgets according to the complexity of each skill.

\noindent\textbf{Extension to VLA.} With diffusion models increasingly serving as action experts in vision-language models (VLMs) to form vision-language-action models (VLAs)~\cite{VLA:Gr00t,VLA:pi}, we extend our framework to this setting. Following GR00T~\cite{VLA:Gr00t}, we adopt the SigLIP-2 vision-language backbone~\cite{siglip-2} as an encoder, with $\ours$ (VLM+$\ours$) or DV (VLM+DV) as the action expert.
We evaluate on three bimanual dexterous tasks from DexMimicGen (Box, Drawer, Tray), which share the same action space. 
As shown in Figure~\ref{fig:case:vla}, VLM+$\ours$ achieves a $9.0$ pp higher success rate on average than VLM+DV, demonstrating that a skill-conditioned noise schedule offers a clear advantage over a fixed one in the VLA setting. We note that only the last four layers of the VLM backbone are finetuned; consequently, the absolute success rates here are not directly comparable to those in Table~\ref{tab:main:1}, where the visual encoder is trained end-to-end on the imitation learning data.

\begin{figure*}[h] 
    \centering 
    \includegraphics[width=.98\textwidth]{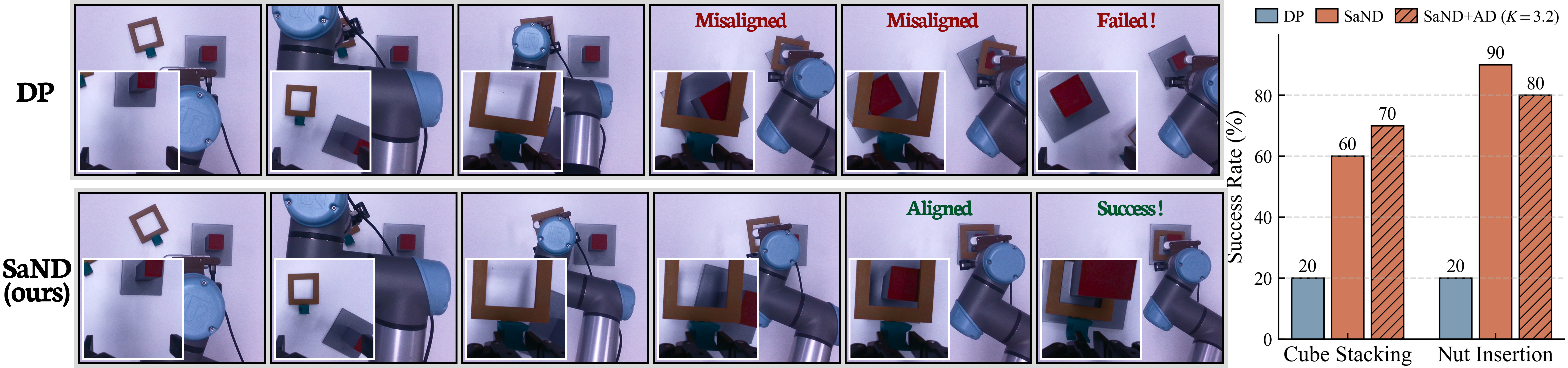}
    \caption{Real-world experiment on a UR7e arm with a Robotiq 2F-85 gripper}
    \label{fig:analysis:real-world}
\end{figure*}

\noindent\textbf{Real-world experiment.} In Figure~\ref{fig:analysis:real-world}, we validate $\ours$ on a real UR7e arm with a Robotiq 2F-85 gripper. We evaluate on two tasks, Cube Stacking and Nut Insertion, both requiring precise rotational alignment and contact-rich control under randomized object poses (see Appendix~\ref{appendix:A:real-world} for the detailed setup).
We collect 100 trajectories per task and evaluate each policy over 20 trials with randomized initial configurations, reporting success rate.
We compare against DP~\cite{DP:DP} rather than SkillDiffuser, the strongest skill-based baseline in simulation, because DP has been extensively validated on physical hardware and serves as the standard reference for real-world baseline.
As shown in the right panel, $\ours$ outperforms DP on both tasks, by $40$ pp on Cube Stacking and $70$ pp on Nut Insertion. Inspecting failures in the left panel, we find that DP most often drifts laterally or misaligns the yaw during insertion and stacking, whereas $\ours$ establishes the grasp orientation more consistently and maintains alignment throughout the approach. This is consistent with these tasks coupling dimensions of differing tolerance, which a per-dimension schedule can accommodate by assigning different denoising rates.



\section{Conclusion} \label{sec:conclusion}
In this work, we presented the $\ours$ framework to enable robust and efficient skill-based robotic manipulation, particularly tailored for contact-rich and dexterous tasks. The framework employs a diffusion-based hierarchical skill learning structure, augmented with a skill-conditioned noise schedule that adaptively adjusts the noise injection rates across action dimensions for each skill. The noise schedule is further regularized through an inverse schedule module, which reconstructs skill embeddings from the Legendre projection of the learned schedule's derivative, encouraging skill-discriminative schedule shapes. 
At inference, we exploit this geometry of the learned noise schedules to assign a skill-specific number of denoising steps via schedule-driven adaptive denoising, substantially reducing inference cost with negligible performance drop. Experiments on contact-rich and dexterous manipulation tasks demonstrate that $\ours$ consistently outperforms competitive baselines in both success rate and efficiency.

\noindent\textbf{Limitations and future work.} Our framework learns a noise schedule conditioned on skill embeddings, which is specifically tailored to hierarchical skill-based policy learning. This is because the skill embedding naturally provides meaningful conditional signals that capture temporal behaviors, making it a natural fit for context-aware schedule learning. Extending this principle to non-skill-based diffusion would require identifying equivalent context signals, which is a promising future direction.

\section*{Acknowledgements}
This work was supported
by the Institute of Information \& Communications Technology Planning \& Evaluation (IITP) grant funded by the Korea
government (MSIT)
(RS-2025-02218768, Accelerated Insight Reasoning via Continual Learning (50\%);
RS-2025-25442569, AI Star Fellowship Support Program (Sungkyunkwan Univ.)),
by the National Research Foundation of Korea (NRF) grant funded by the Korea government (MSIT) (RS-2026-25474409),
National IT Industry Promotion Agency (NIPA) grant funded by the Korea government (MSIT) (Research and Development of SDF-Specialized Physical AI Foundation Models for Heterogeneous Device Collaboration, No. RS-2026-25621569)
and by Samsung Electronics Co., Ltd. (1\%).


\bibliographystyle{plainnat}
\bibliography{reference}


\appendix
\appendix
\section{Experimental Environments} \label{appendix:A}

\subsection{Robosuite} \label{appendix:A:1}
Robosuite~\citep{sim:robosuite} is a simulation framework for robot learning built on the MuJoCo physics engine~\citep{sim:mujoco}. It provides a diverse set of robotic environments, ranging from gripper-based manipulation setups to dexterous hand systems.
In particular, we adopt two publicly available benchmark datasets from \textbf{MimicGen}
~\citep{dataset:mimicgen} and \textbf{DexMimicGen}
~\citep{dataset:dexmimicgen}, which focus on \textit{contact-rich} and \textit{bimanual dexterous} manipulation tasks, respectively.
Both datasets are automatically synthesized from a small number of human demonstrations (e.g., 10 demonstrations). 

\begin{figure}[h]
    \centering
    \subfigure[Square]{
        \centering
        \includegraphics[width=0.15\linewidth]{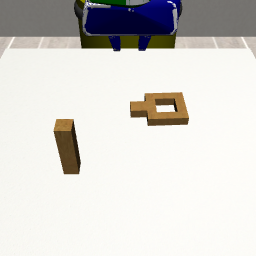}
        \includegraphics[width=0.15\linewidth]{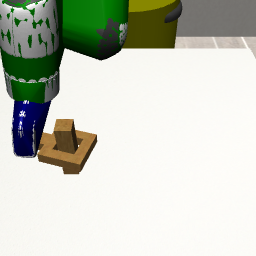}
    }
    \subfigure[Threading]{
        \centering
        \includegraphics[width=0.15\linewidth]{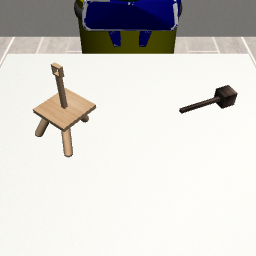}
        \includegraphics[width=0.15\linewidth]{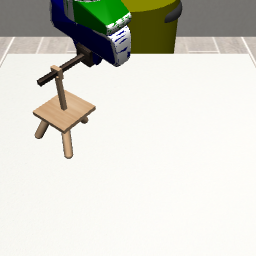}
    }
    \subfigure[Mug Cleanup]{
        \centering
        \includegraphics[width=0.15\linewidth]{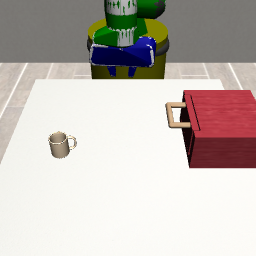}
        \includegraphics[width=0.15\linewidth]{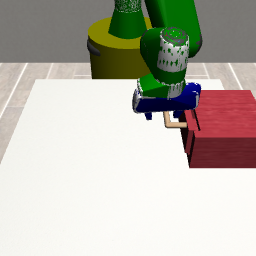}
    }
    \\
    \subfigure[Hammer Cleanup]{
        \centering
        \includegraphics[width=0.15\linewidth]{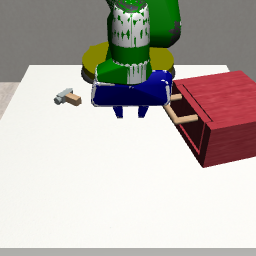}
        \includegraphics[width=0.15\linewidth]{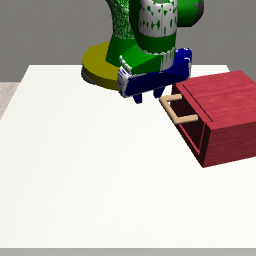}
    }
    \subfigure[Coffee Preparation]{
        \centering
        \includegraphics[width=0.15\linewidth]{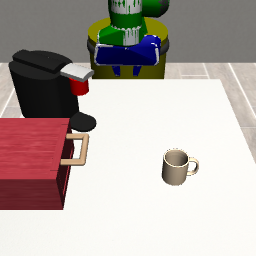}
        \includegraphics[width=0.15\linewidth]{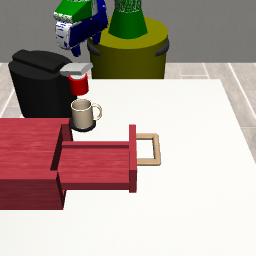}
    }
    \subfigure[Kitchen]{
        \centering
        \includegraphics[width=0.15\linewidth]{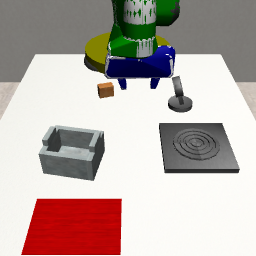}
        \includegraphics[width=0.15\linewidth]{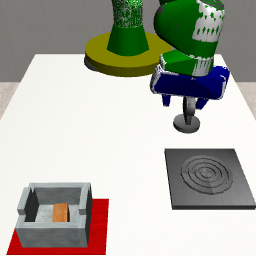}
    }
    \caption{Visualization of start and end states for each task in MimicGen}
    \label{fig:appendix:A:MimicGen}
\end{figure}

\noindent\textbf{MimicGen}~\citep{dataset:mimicgen} consists of single-arm manipulation tasks using a Panda robot equipped with a parallel-jaw gripper. We consider a \textit{low-dimensional} observation space consisting of object state and the robot's proprioceptive state.
%
The action space is 7-dimensional, consisting of Cartesian end-effector position increments ($\Delta x, \Delta y, \Delta z$), 3D rotational increments ($\Delta$\textit{yaw}, $\Delta$\textit{pitch}, $\Delta$\textit{roll}), and a gripper open/close command. 
We evaluate on the following six tasks: 
\begin{itemize}
    \item \textbf{Square} : Pick a square nut and place it on a peg. 
    \item \textbf{Threading} : Pick a needle and thread it through a hole on a tripod.
    \item \textbf{Mug Cleanup} : Open a drawer, pick a mug, and place it inside the drawer.
    \item \textbf{Hammer Cleanup} : Open a drawer, pick up a hammer, and place it inside the drawer.
    \item \textbf{Coffee Preparation} : Load a mug onto the machine, open the machine, retrieve a coffee pod from the drawer, and insert it into the machine.
    \item \textbf{Kitchen} : Turn on the stove, place a pot onto the stove, place a piece of bread into the pot, place the pot in front of the serving region and push it there, and turn off the stove.
\end{itemize}
For simplicity, each task name in the experiment tables is abbreviated using the first word (e.g., Mug Cleanup as Mug). Object positions and orientations are randomized across episodes to ensure contact-rich evaluation. For each task, we use 500 demonstrations for training. Figure~\ref{fig:appendix:A:MimicGen} visualizes the start and end states of task environments in MimicGen.

In the original MimicGen paper, performance is reported by selecting the best checkpoint based on evaluation over 50 episodes. However, due to environmental stochasticity, we observe substantial performance variance across runs. Notably, the success rate of the selected best checkpoint fluctuates significantly when re-evaluated, suggesting that checkpoint selection based on a small number of episodes is insufficient for providing a statistically reliable estimate of performance. To ensure a more statistically reliable evaluation, we instead report the success rate of the final trained model and increase the number of evaluation episodes from 50 to 500.

\begin{figure}[h]
    \centering
    \subfigure[Box Cleanup]{
        \centering
        \includegraphics[width=0.15\linewidth]{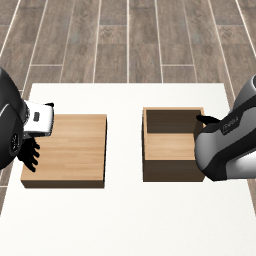}
        \includegraphics[width=0.15\linewidth]{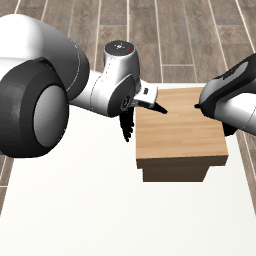}
    }
    \subfigure[Drawer Cleanup]{
        \centering
        \includegraphics[width=0.15\linewidth]{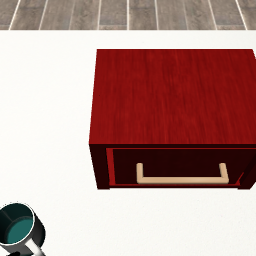}
        \includegraphics[width=0.15\linewidth]{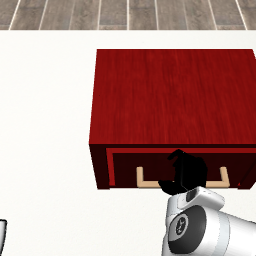}
    }
    \subfigure[Tray Lift]{
        \centering
        \includegraphics[width=0.15\linewidth]{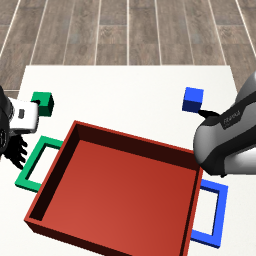}
        \includegraphics[width=0.15\linewidth]{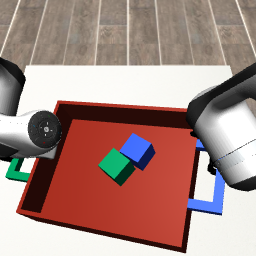}
    }
    \\
    \subfigure[Pouring]{
        \centering
        \includegraphics[width=0.15\linewidth]{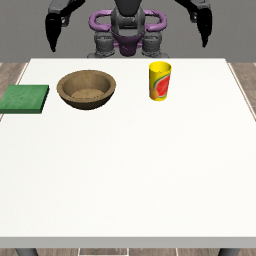}
        \includegraphics[width=0.15\linewidth]{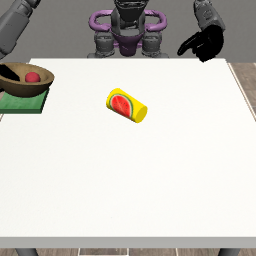}
    }
    \subfigure[Coffee]{
        \centering
        \includegraphics[width=0.15\linewidth]{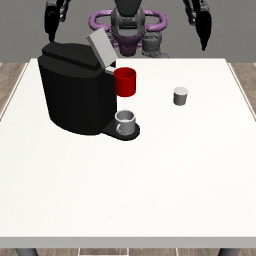}
        \includegraphics[width=0.15\linewidth]{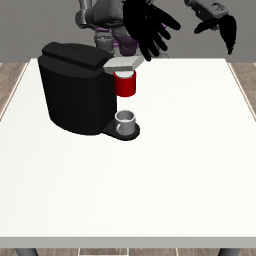}
    }
    \subfigure[Can Sort]{
        \centering
        \includegraphics[width=0.15\linewidth]{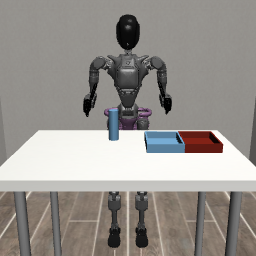}
        \includegraphics[width=0.15\linewidth]{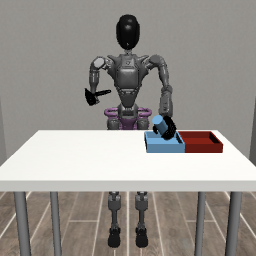}
    }
    \caption{Visualization of start and end states for each task in DexMimicGen}
    \label{fig:appendix:A:DexMimicGen}
\end{figure}

\noindent\textbf{DexMimicGen}~\citep{dataset:dexmimicgen} consists of bimanual manipulation tasks with two variants: (\romannumeral 1) dual Panda robots equipped with dexterous hands, and (\romannumeral 2) a GR-1 humanoid with dexterous hands. 
We consider a \textit{multimodal} observation space consisting of three camera views---an agent (or front) view and two robot wrist-mounted views (one for the left hand and one for the right hand)---together with the robot proprioceptive state. 
The action space is 24-dimensional for both variants, comprising 3D Cartesian end-effector position commands, 3D rotational commands, and 6D dexterous hand joint commands for each of the two arms. For the dual-Panda setup, these actions are represented as relative increments, whereas for the GR-1 humanoid, the actions are represented as absolute values.
We evaluate on the following six tasks, the first three for dual Panda robots with dexterous hands and the last three for the GR-1 humanoid, each requiring coordinated manipulation using both arms:
\begin{itemize}
    \item \textbf{Box Cleanup} : Lift the lid of a box and then close it.
    \item \textbf{Drawer Cleanup} : Open a drawer and place a mug inside it.
    \item \textbf{Tray Lift} : Pick up a cube with each arm, place them into a tray, and lift the tray together.
    \item \textbf{Pouring} : Pour a ball from a cup into a bowl, then place the bowl in the serving region.
    \item \textbf{Coffee} : Insert a coffee pod into the machine and close the lid.
    \item \textbf{Can Sort} : Pick up a can with one hand, hand it over to the other hand, and place it on a target spot.
\end{itemize}
For simplicity, each task name in the experiment tables is abbreviated using the first word (e.g., Tray Lift as Tray). For each task, we use 500 demonstrations for training, and report the success rate of the final trained model evaluated over 100 episodes. Figure~\ref{fig:appendix:A:DexMimicGen} visualizes the start and end states of task environments in DexMimicGen.

\subsection{Isaac Gym}
Isaac Gym~\citep{sim:isaacgym} is a GPU-accelerated robotics simulation framework built on the NVIDIA Omniverse platform with the PhysX physics engine. We evaluate on the \textbf{Bi-DexHands}~\citep{dataset:BiDexHands} benchmark environments, which feature \textit{bimanual dexterous} manipulation tasks. Since no publicly available demonstration dataset exists for these tasks, we train an agent using the Proximal Policy Optimization (PPO) algorithm~\citep{ppo} and collect successful trajectories from the  trained policy.

\noindent\textbf{Bi-DexHands}~\citep{dataset:BiDexHands} consists of dexterous manipulation tasks using two Shadow dexterous hands with floating bases. We consider a \textit{low-dimensional} observation space comprising object poses and robot proprioceptive state. The action space is 52-dimensional, comprising 20 actuated joint position targets per hand and 6 DoF for the floating base of each hand.
We evaluate on the following six tasks, each requiring coordinated manipulation by both hands:
\begin{itemize}
    \item \textbf{Catch Over Underarm}: Throw an object from a vertically oriented hand and catch it with a palm-up hand.
    \item \textbf{Door Open}: Push open a door that can only be opened in the outward direction.
    \item \textbf{Pot Lift}: Grasp the handles of a pot with both hands and lift it to a target position.
    \item \textbf{Scissors Open}: Open the scissors using both hands.
    \item \textbf{Pen Open}: Remove the cap from a pen using both hands.
    \item \textbf{Cup Swing}: Hold the cup handle with both hands and rotate it 90 degrees.
\end{itemize}

\begin{figure}[h]
    \centering
    \subfigure[Catch Over Underarm]{
        \centering
        \includegraphics[width=0.15\linewidth]{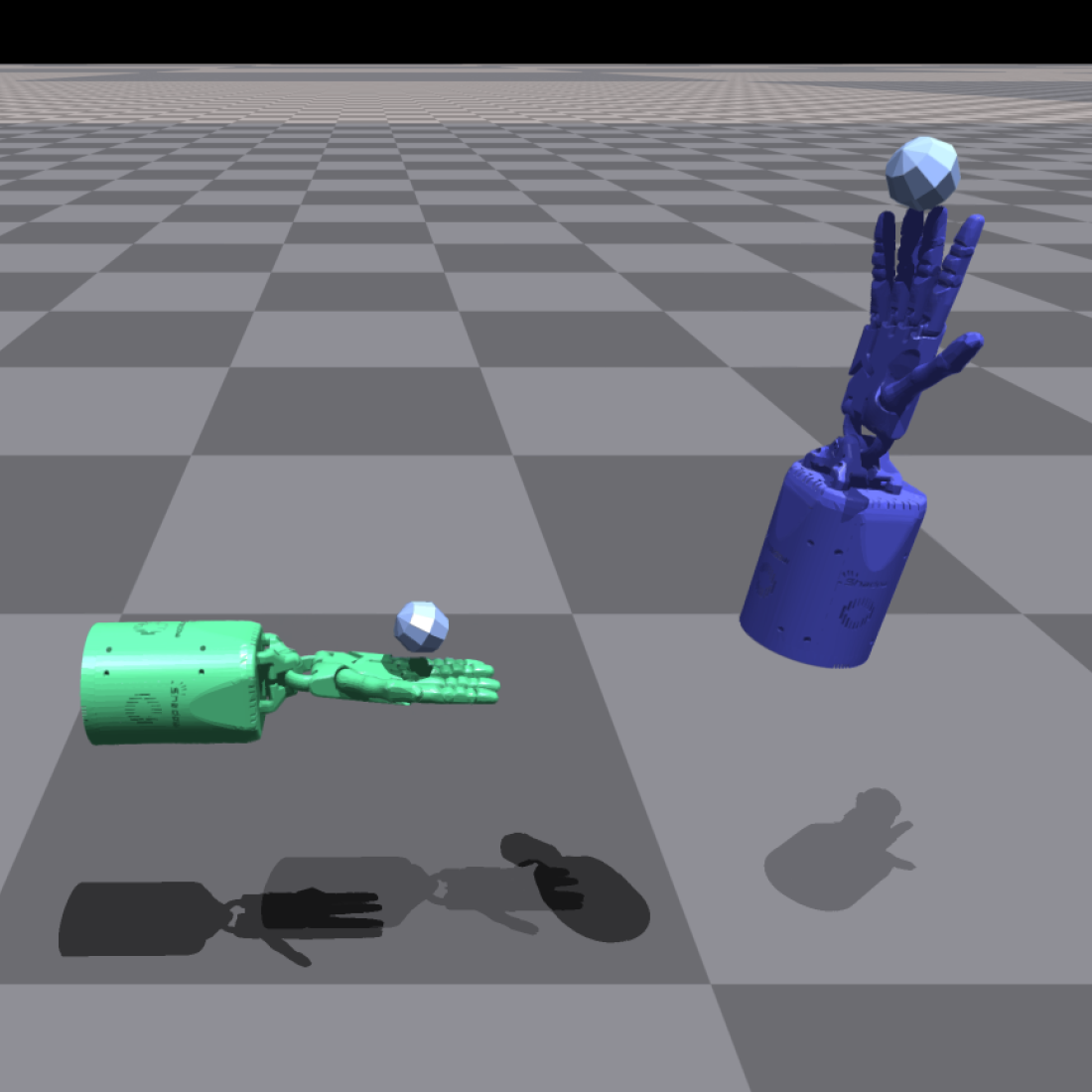}
        \includegraphics[width=0.15\linewidth]{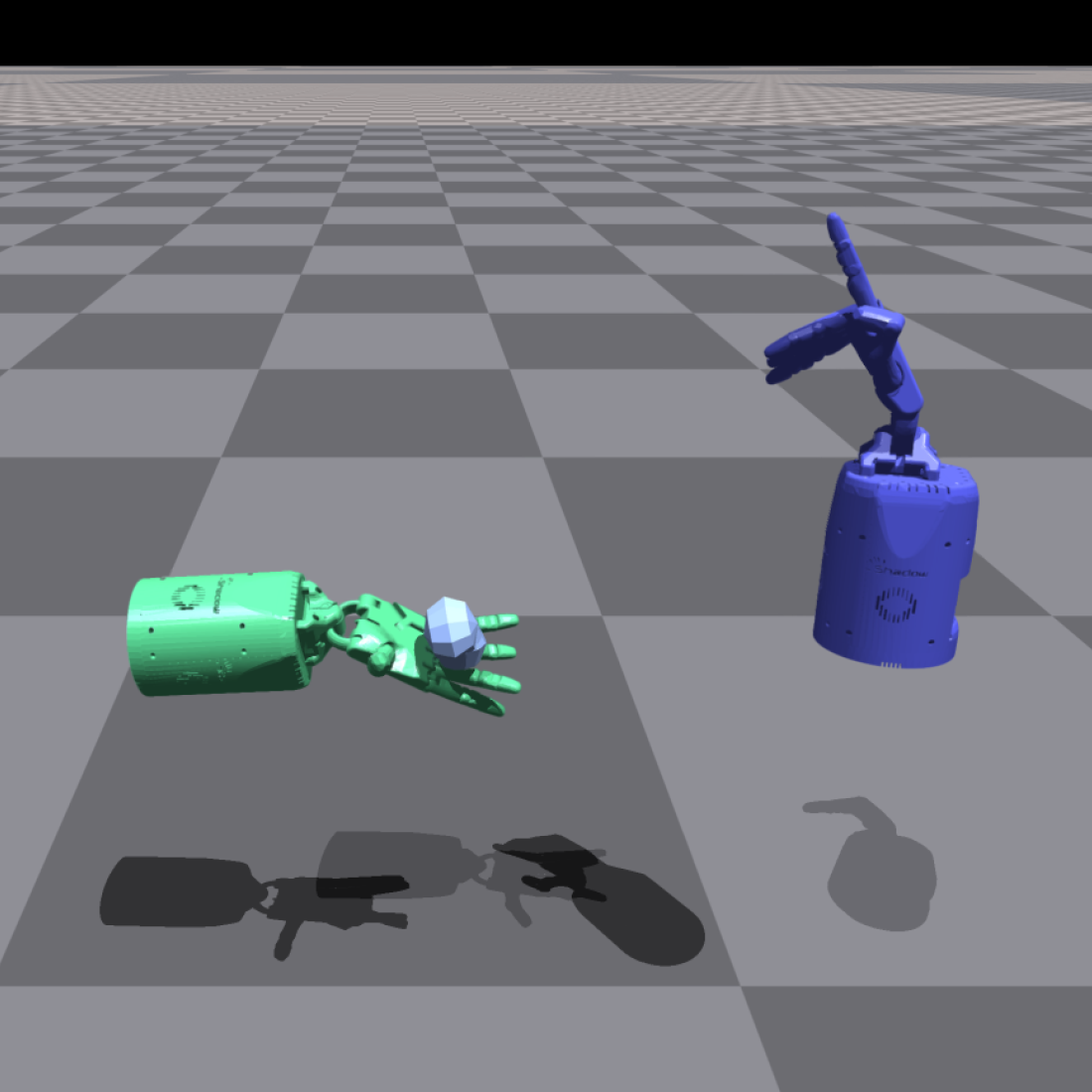}
    }
    \subfigure[Door Open]{
        \centering
        \includegraphics[width=0.15\linewidth]{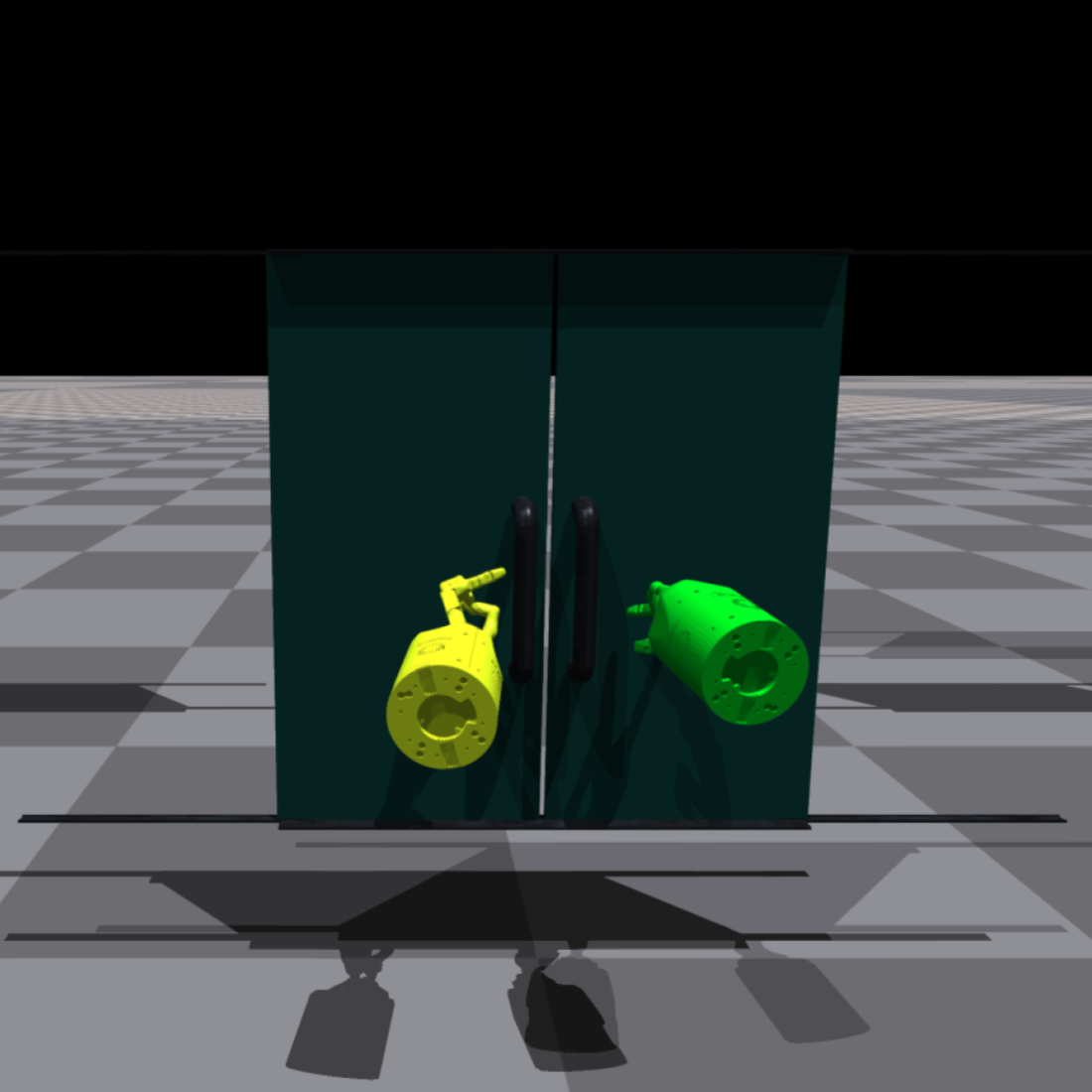}
        \includegraphics[width=0.15\linewidth]{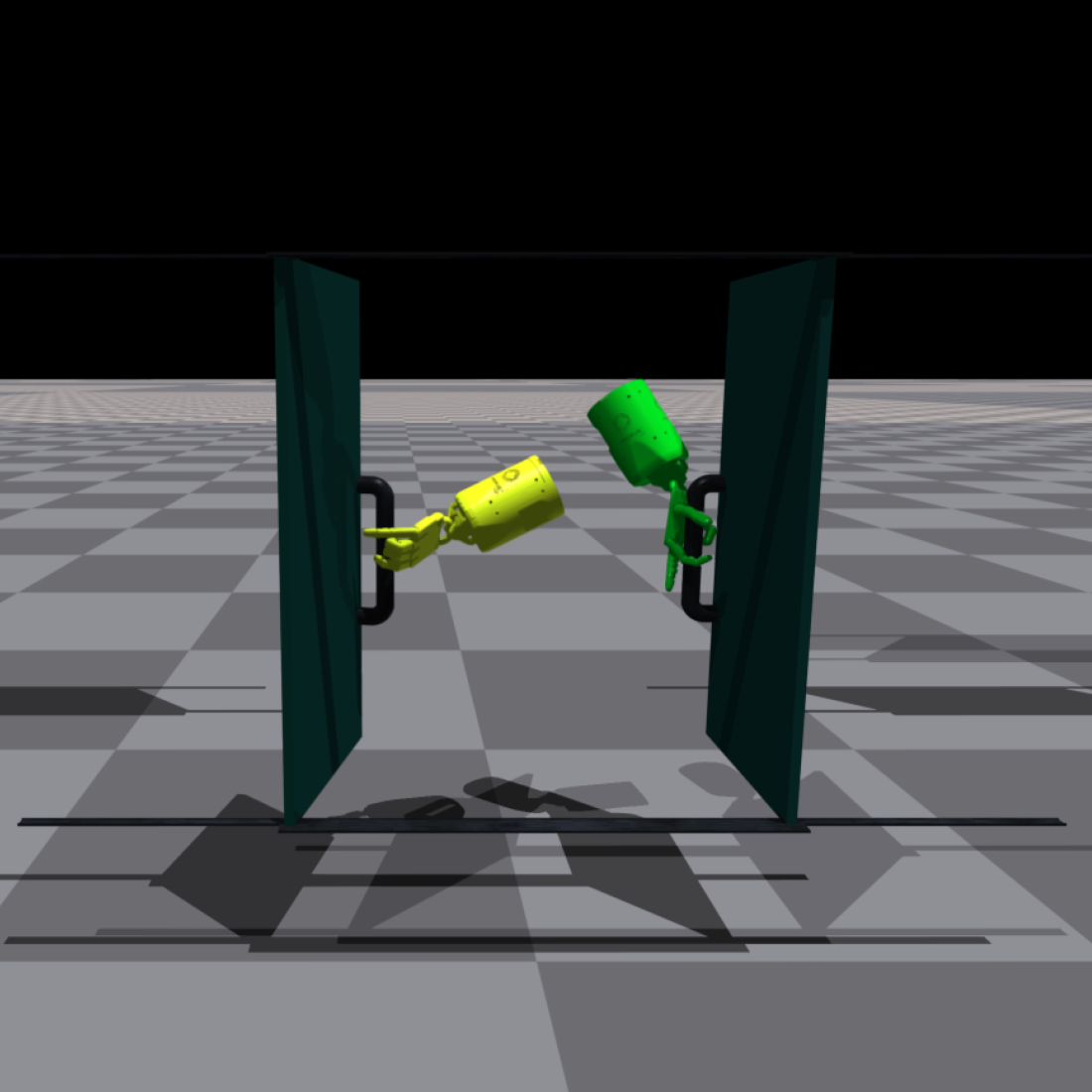}
    }
    \subfigure[Pot Lift]{
        \centering
        \includegraphics[width=0.15\linewidth]{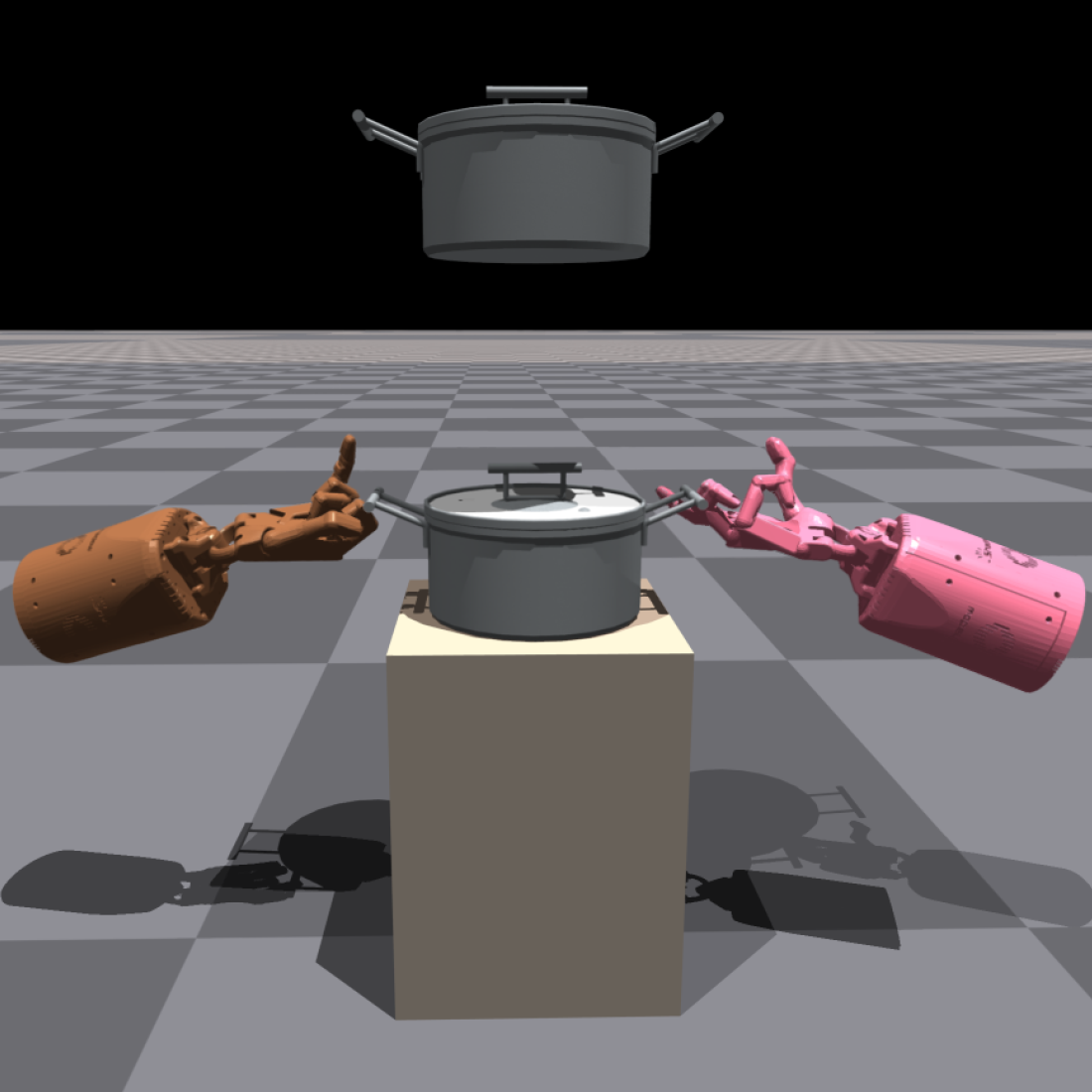}
        \includegraphics[width=0.15\linewidth]{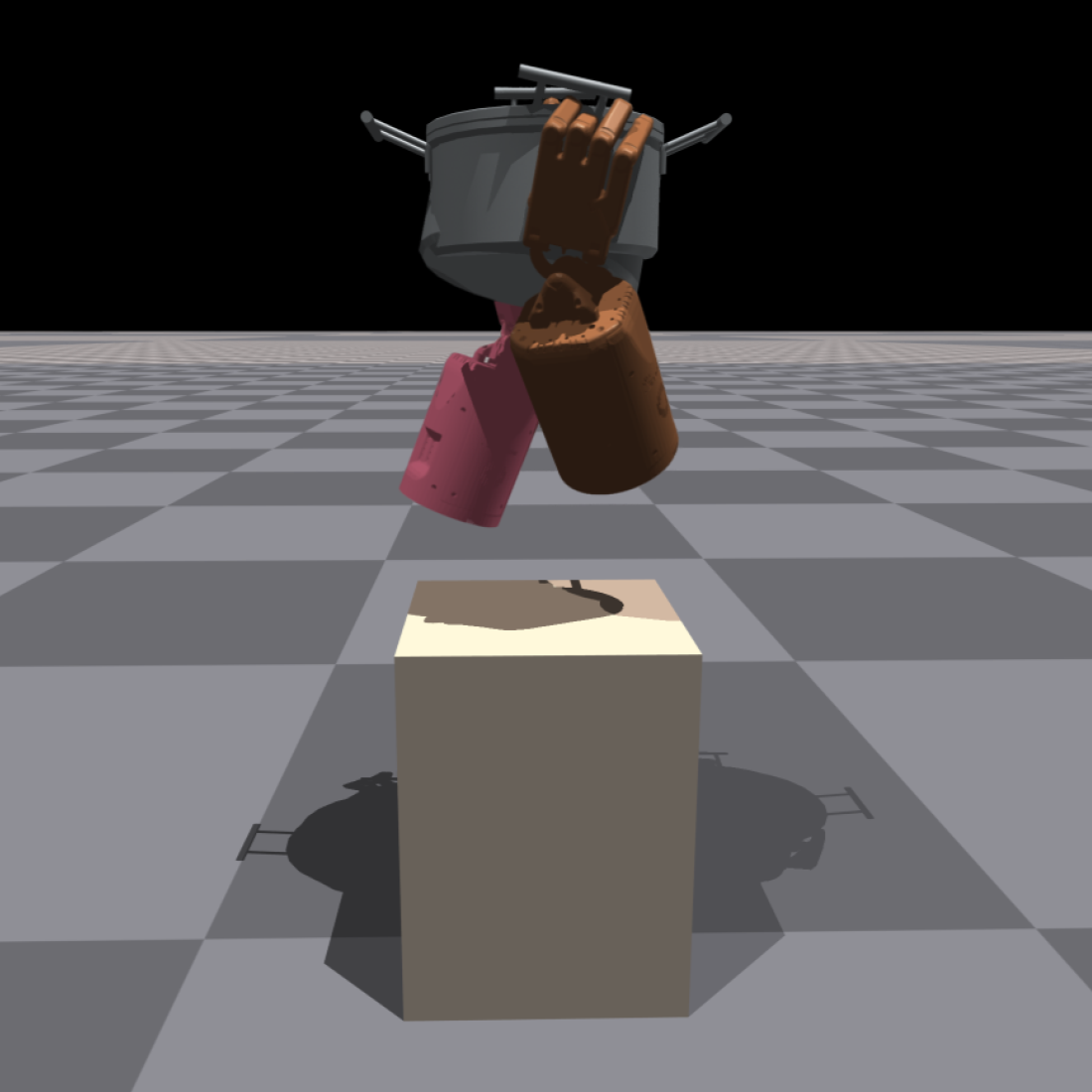}
    }
    \\
    \subfigure[Scissors Open]{
        \centering
        \includegraphics[width=0.15\linewidth]{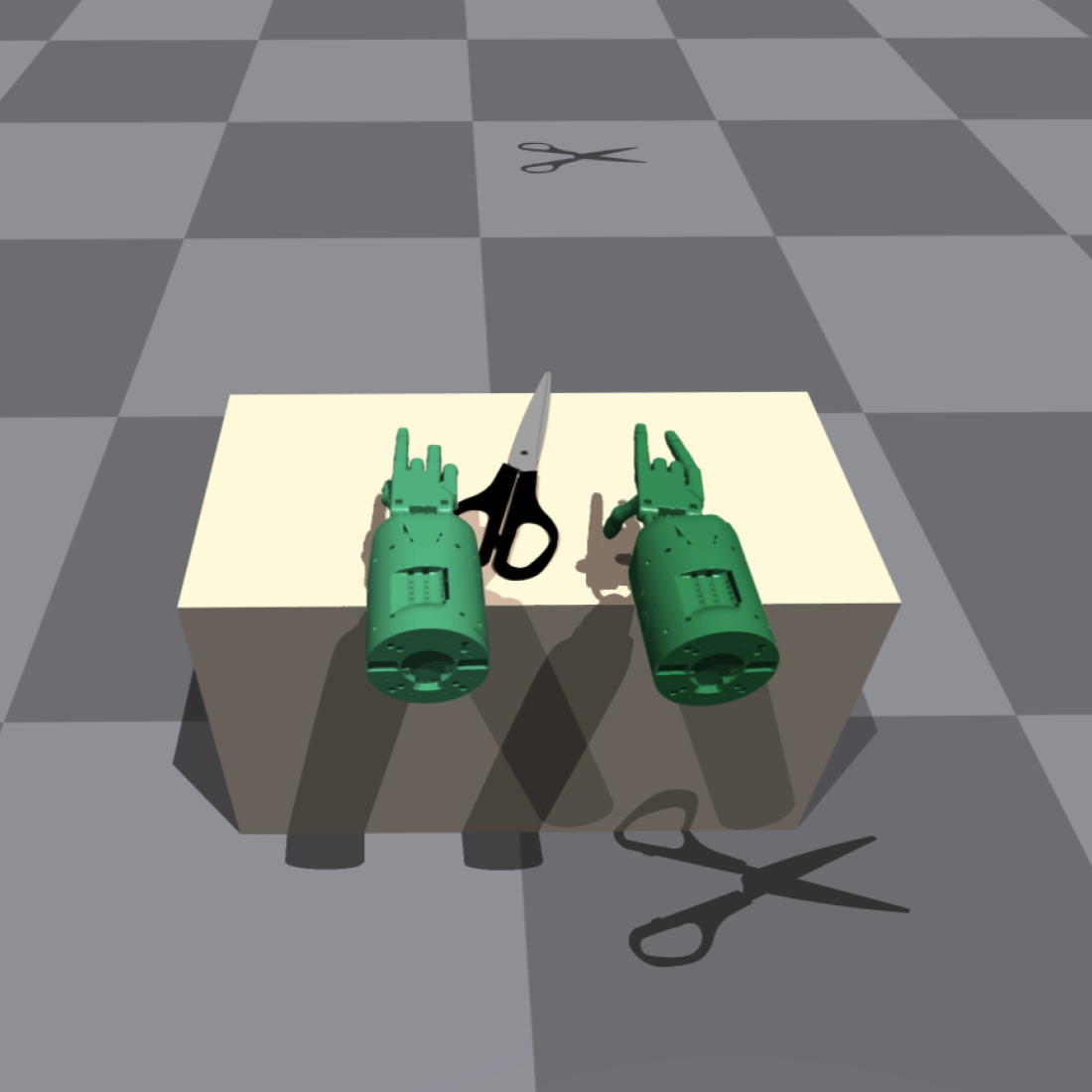}
        \includegraphics[width=0.15\linewidth]{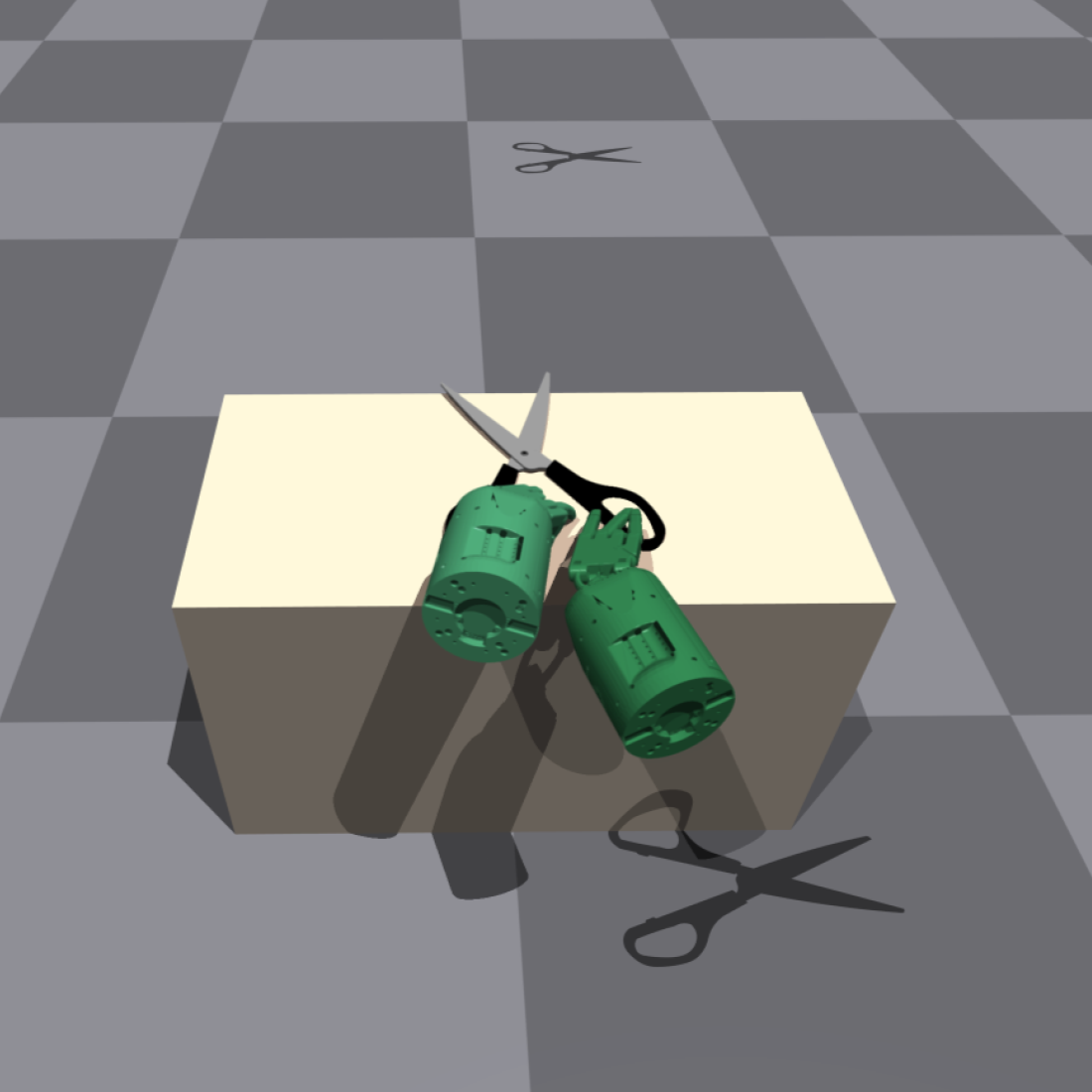}
    }
    \subfigure[Pen Open]{
        \centering
        \includegraphics[width=0.15\linewidth]{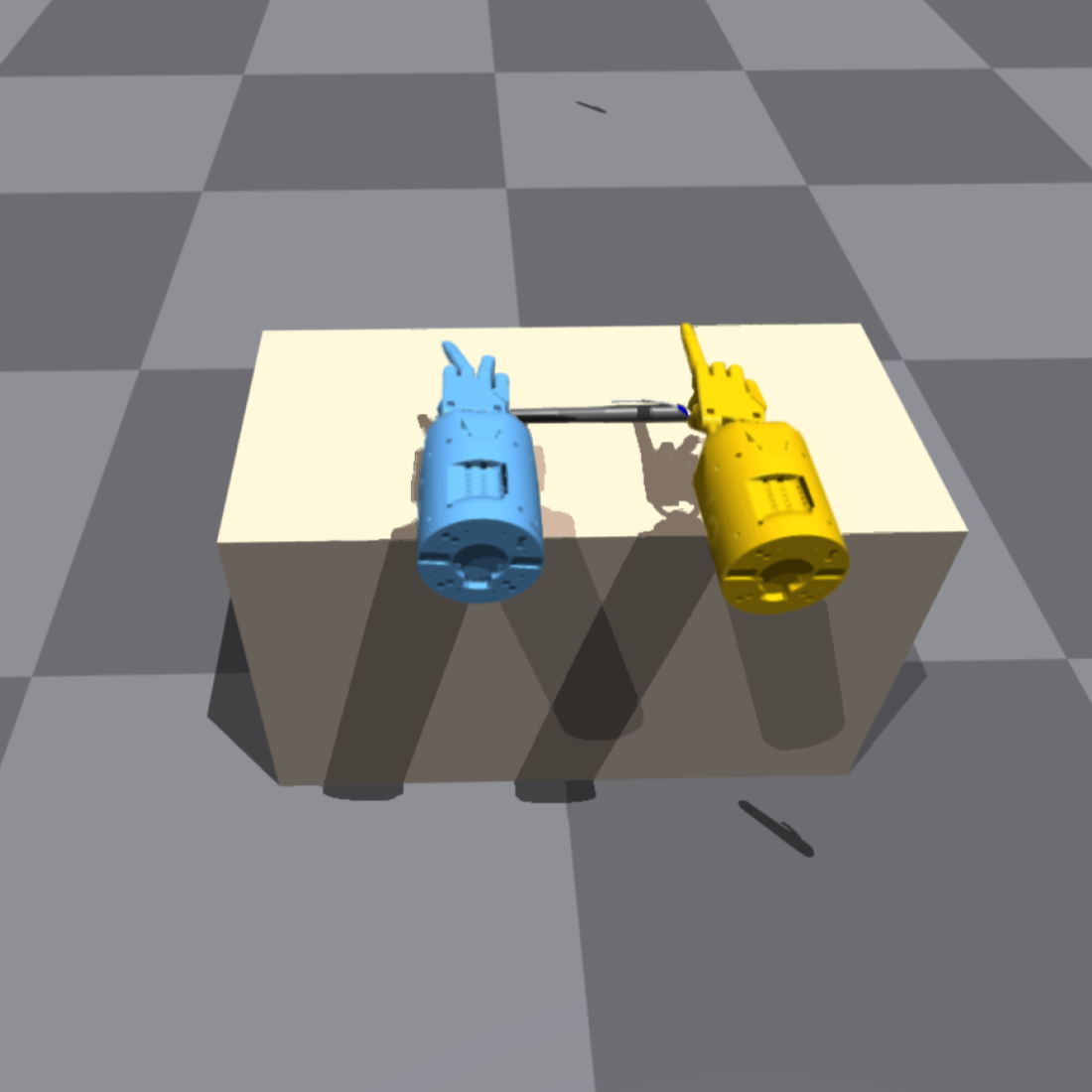}
        \includegraphics[width=0.15\linewidth]{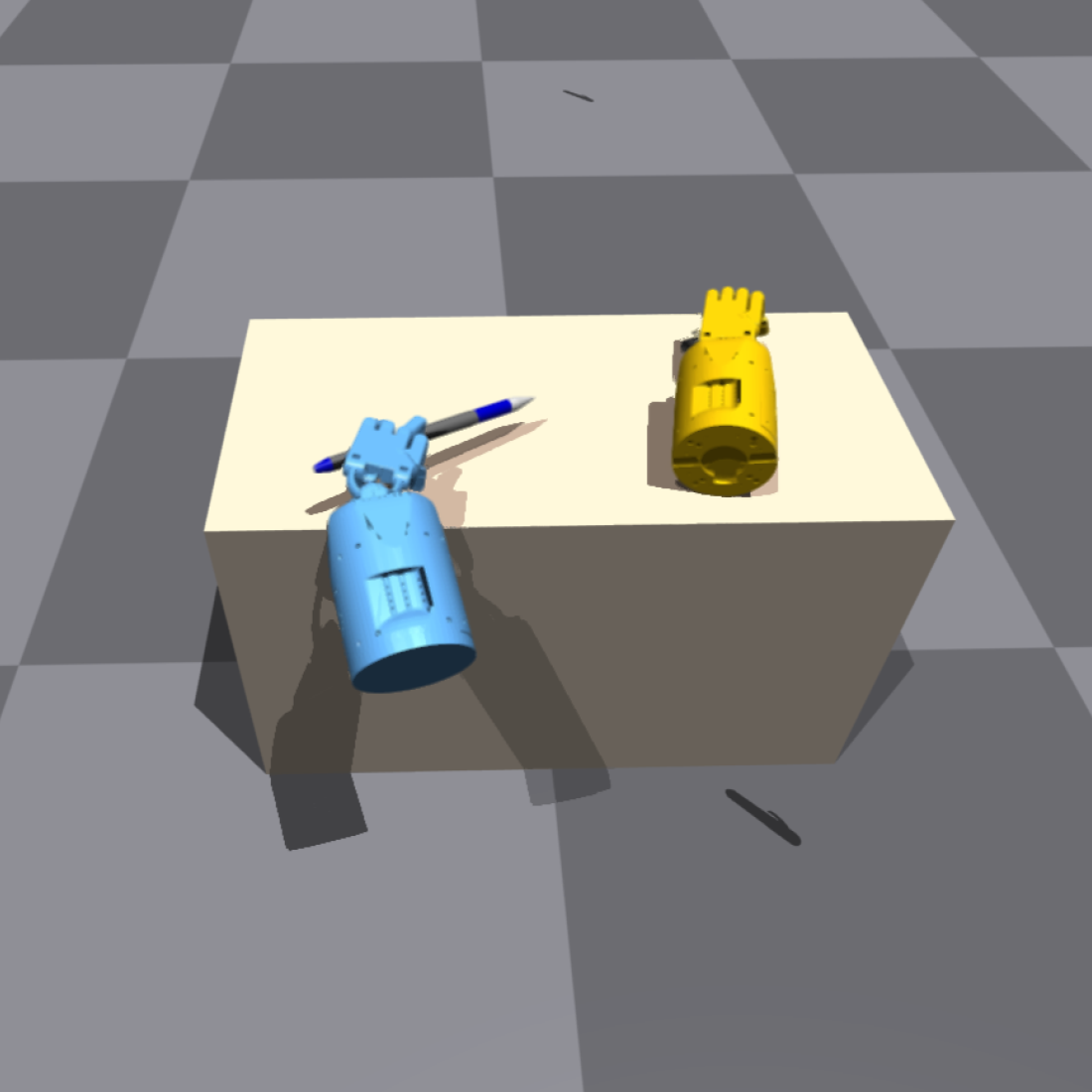}
    }
    \subfigure[Cup Swing]{
        \centering
        \includegraphics[width=0.15\linewidth]{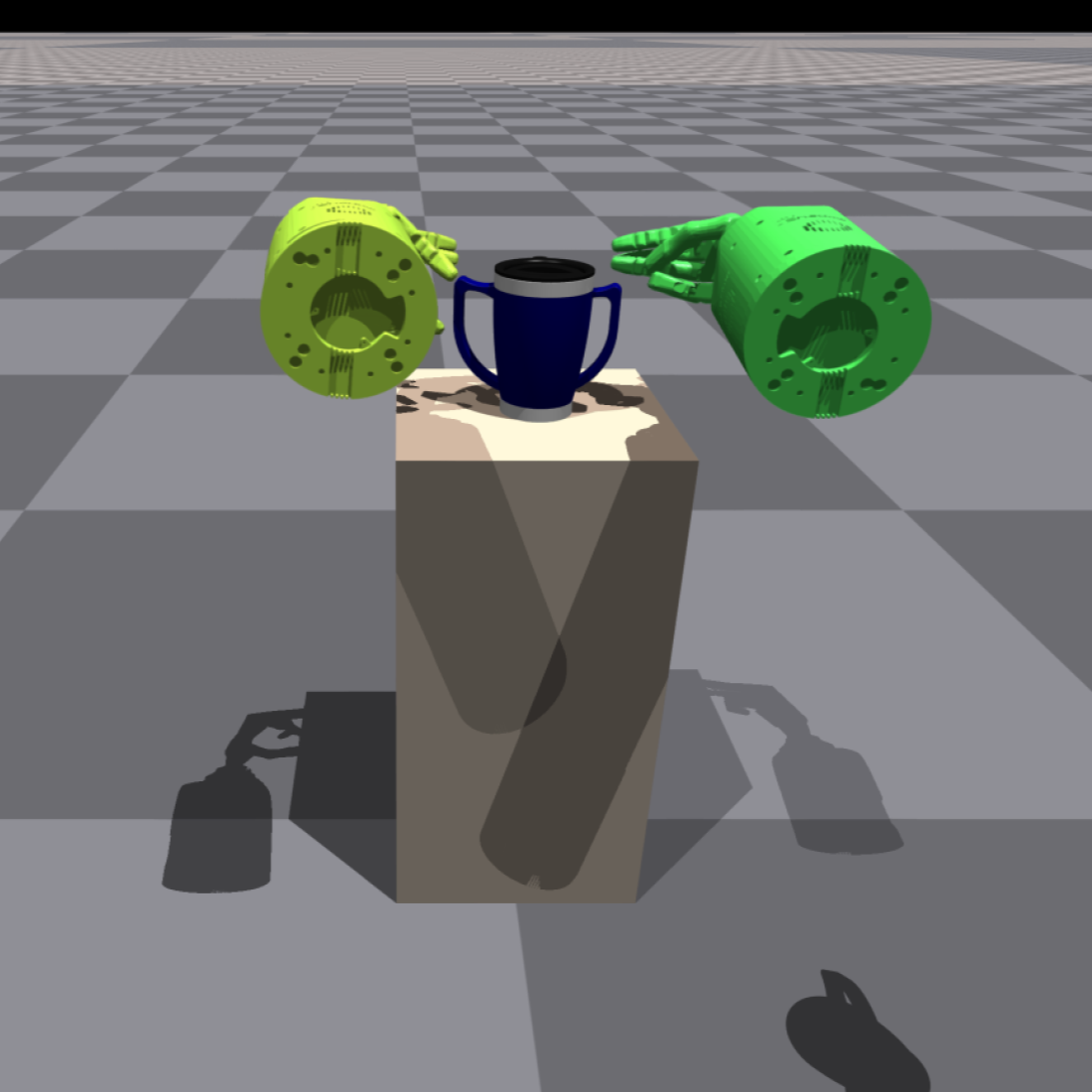}
        \includegraphics[width=0.15\linewidth]{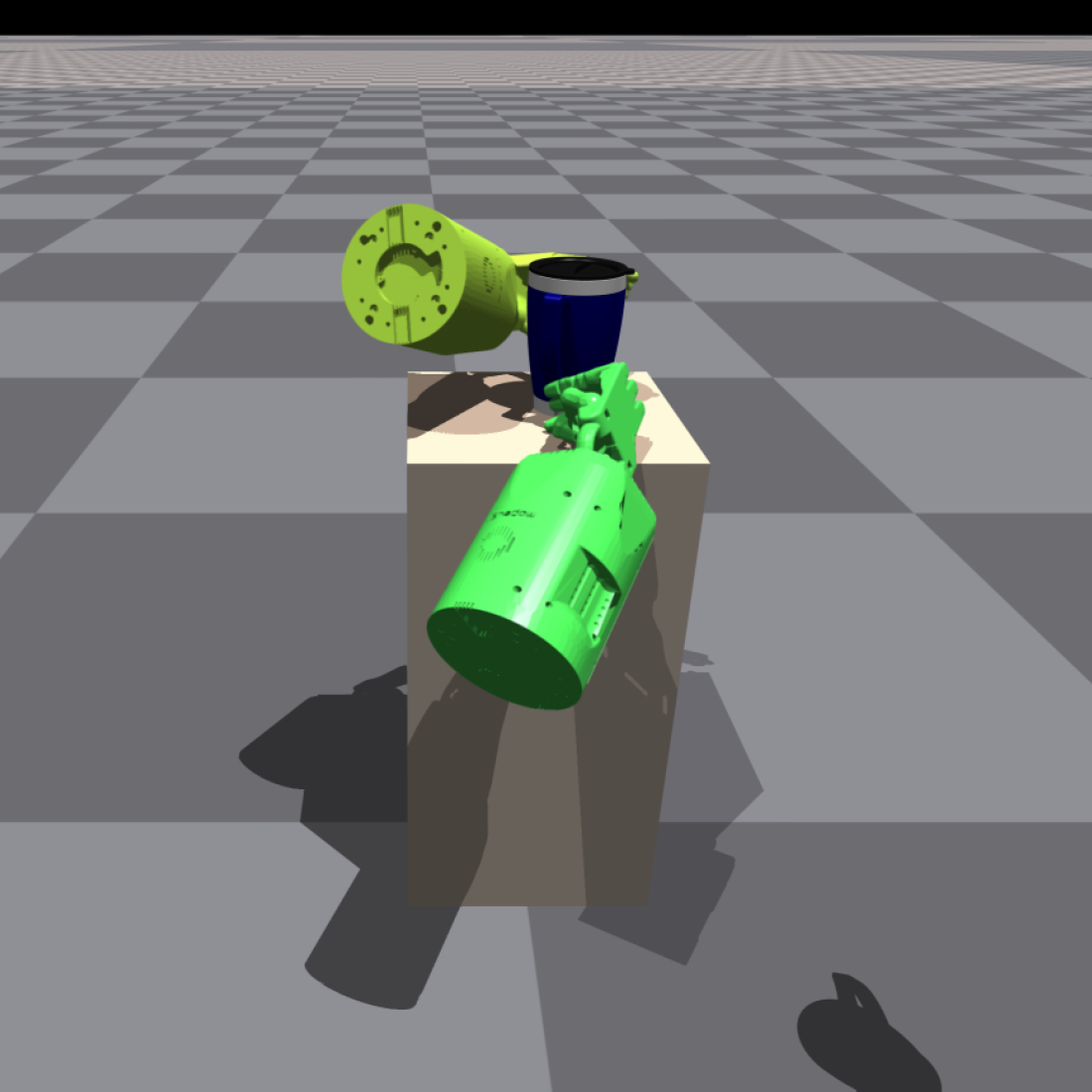}
    }
    \caption{Visualization of start and end states for each task in Bi-DexHands}
    \label{fig:appendix:A:Bi-DexHands}
\end{figure}

For each task, we use 100 demonstrations for training, and report the accumulated reward of the final trained model evaluated over 200 episodes. Figure~\ref{fig:appendix:A:Bi-DexHands} visualizes start and end states of task environments in Bi-DexHands.
%




\subsection{Real-World Experiment Setup} \label{appendix:A:real-world}
We conduct real-robot experiments using a UR7e robot arm equipped with a Robotiq 2F-85 parallel gripper. The observation space consists of RGB images captured by two Intel RealSense D435 cameras (one top-mounted and one wrist-mounted) at 256 $\times$ 256 resolution, along with the robot's joint states. The action space is 7-dimensional, comprising Cartesian end-effector position increments, rotational increments, and a binary gripper command. 

\begin{figure}[h]
    \centering
    \subfigure[Cube Stacking]{
        \centering
        \includegraphics[width=0.2\linewidth]{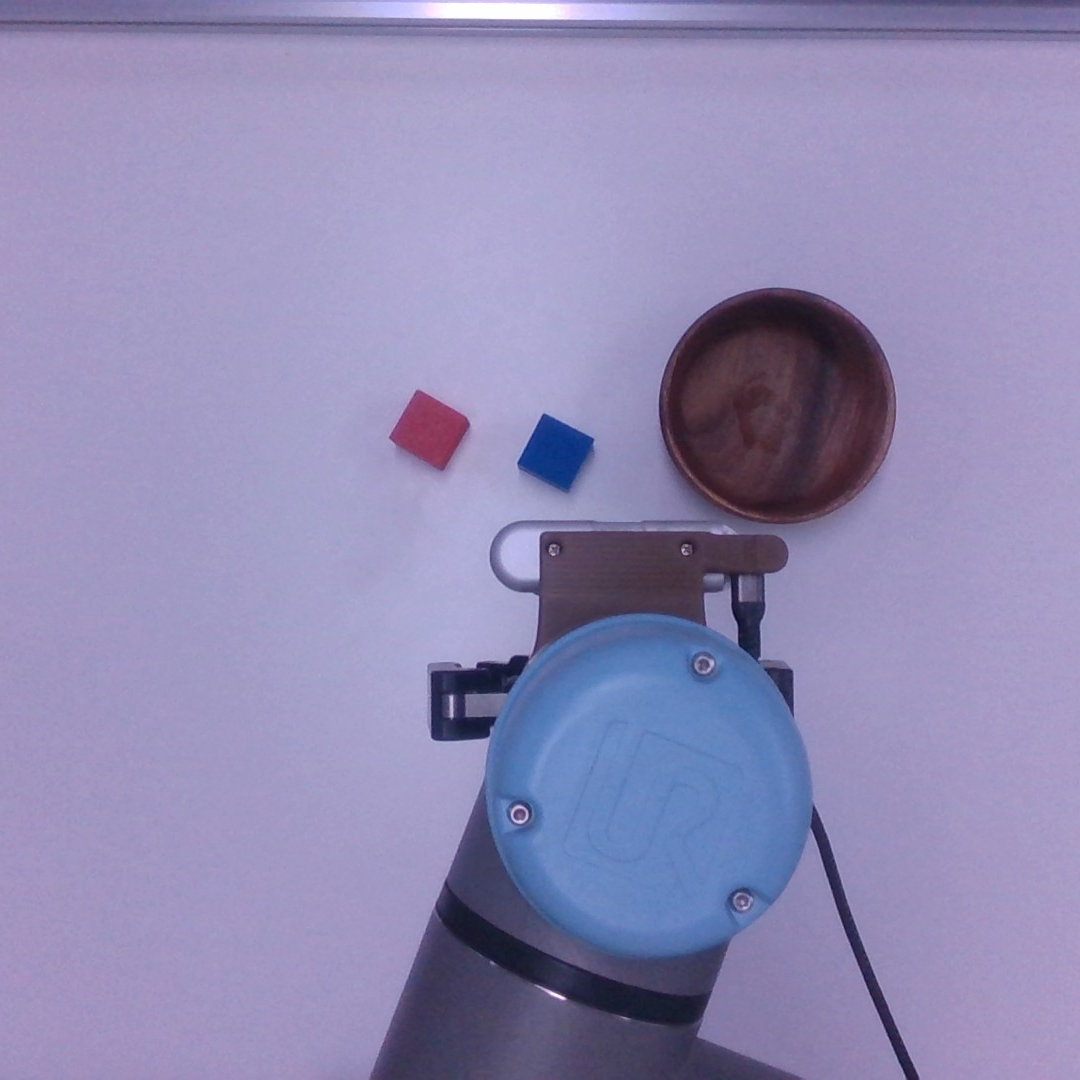}
        \includegraphics[width=0.2\linewidth]{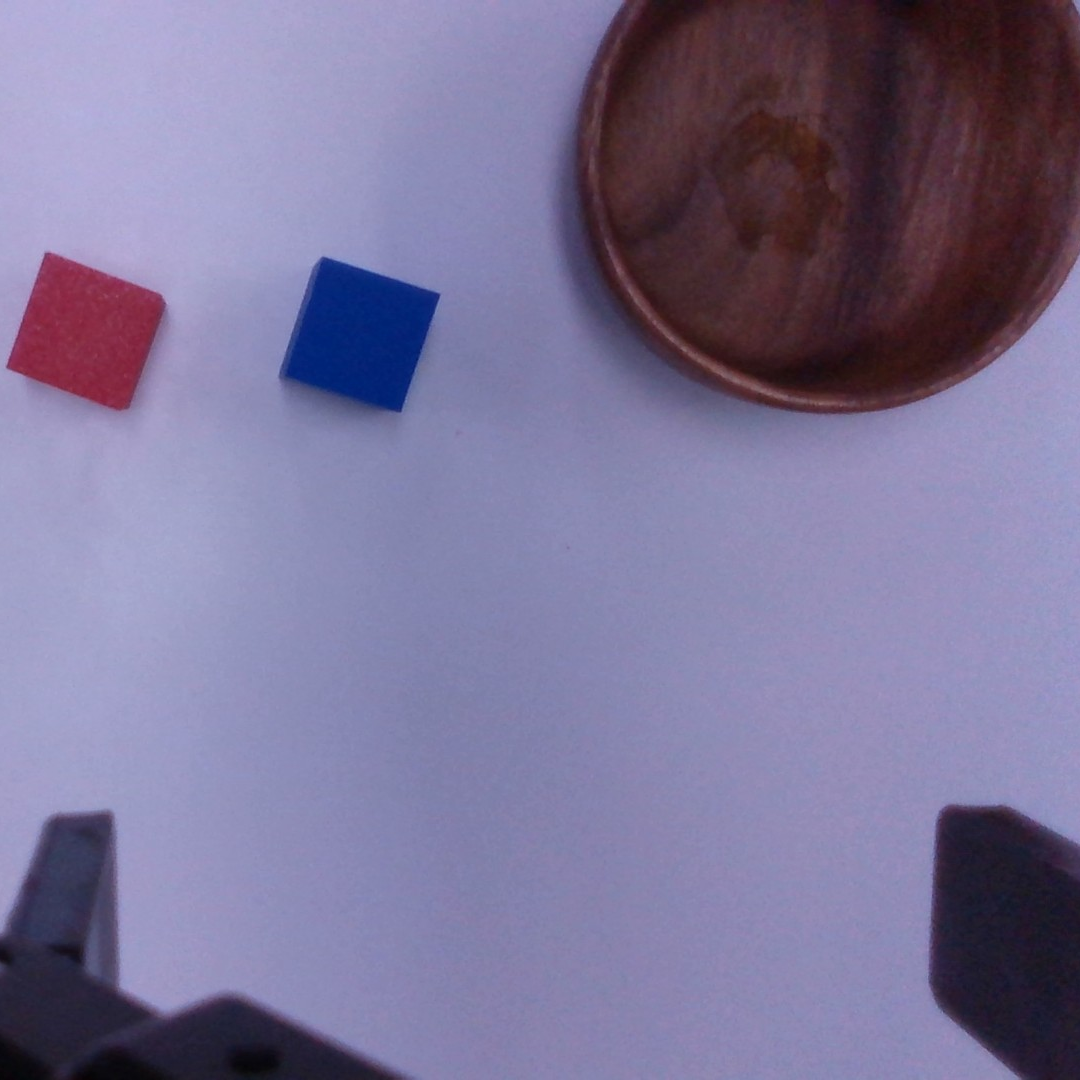}
    }
    \subfigure[Nut Insertion]{
        \centering
        \includegraphics[width=0.2\linewidth]{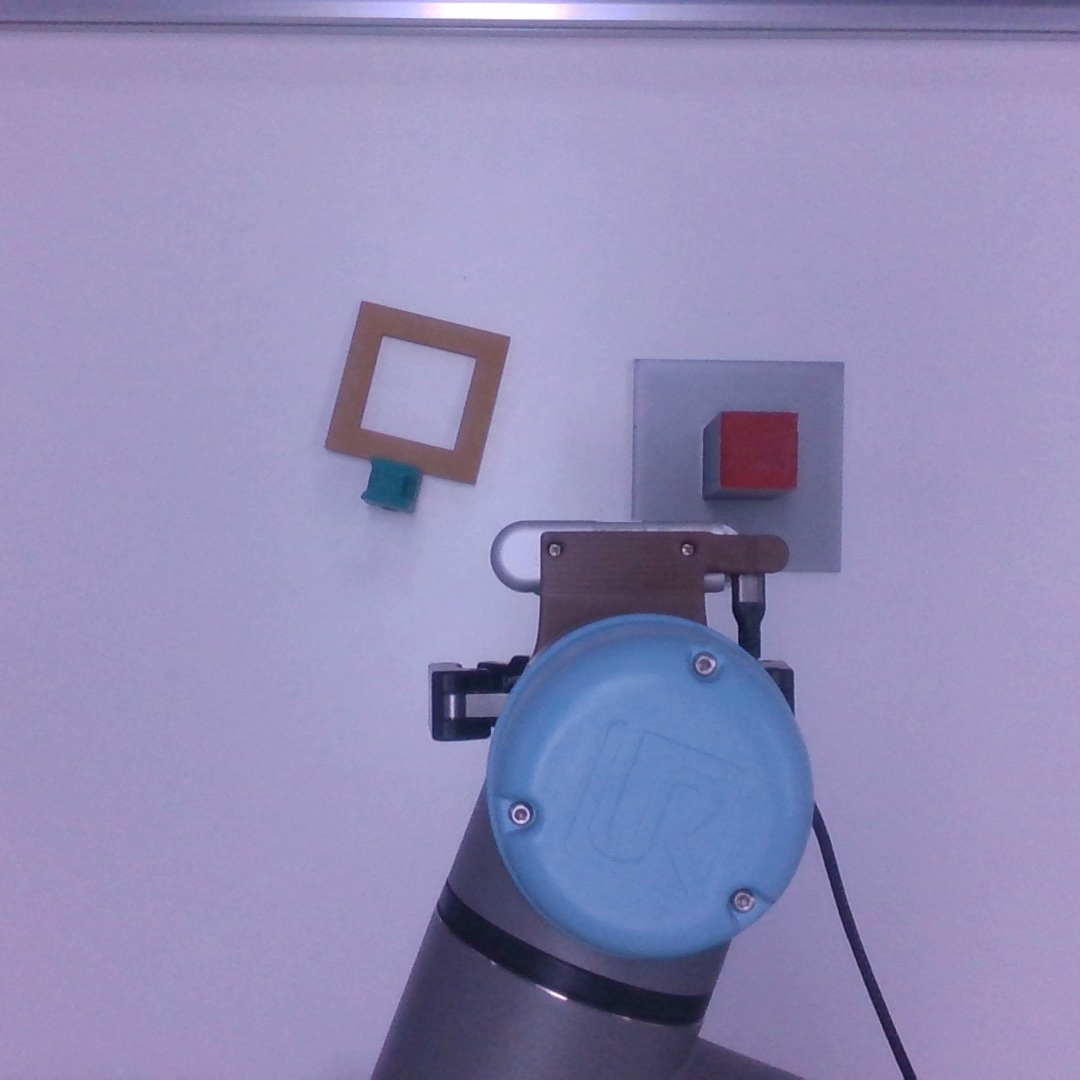}
        \includegraphics[width=0.2\linewidth]{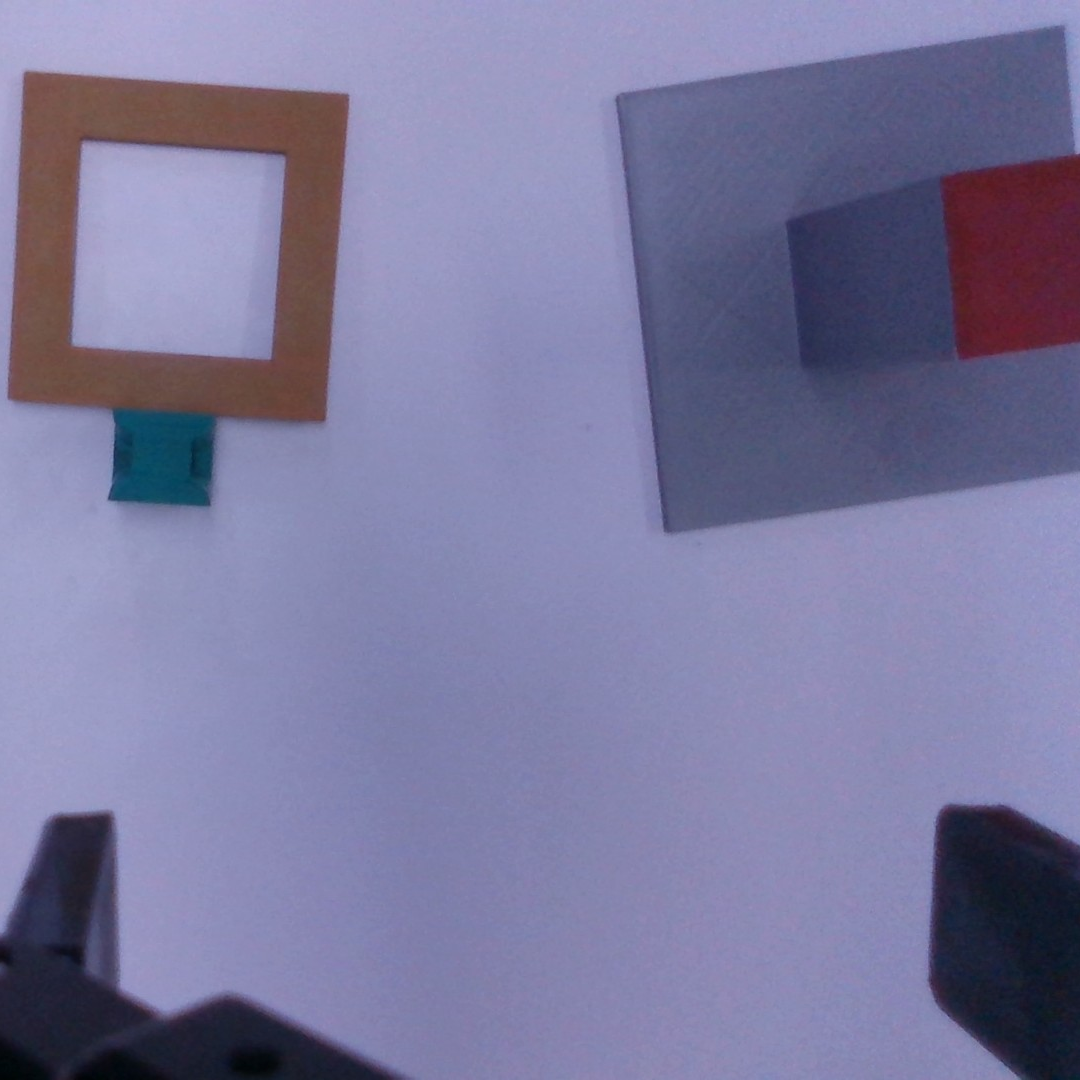}
    }
    \caption{Top and wrist camera views for each real-world task}
    \label{fig:appendix:A:realworld}
\end{figure}

For evaluation, we configure two tasks: (\romannumeral 1) In the \textbf{Cube Stacking} task, two cubes are initialized at random positions and orientations, requiring the policy to align the gripper yaw to each cube before grasping and placing it into a bowl. (\romannumeral 2) In the \textbf{Nut Insertion} task, mirroring the Square task in MimicGen, a square nut must be inserted onto a peg. The peg cross-section closely matches the nut's hole, leaving little tolerance at insertion, and the nut is initialized at a randomized orientation.
Figure~\ref{fig:appendix:A:realworld} shows the top and wrist camera views for each task.
%
%

\color{black}
%

\section{Implementation Details} \label{appendix:B}
In this section, we provide the implementation details of $\ours$ and the baselines. All experiments are conducted on a system equipped with an AMD Ryzen Threadripper PRO 5975WX 32-core CPU and an NVIDIA RTX 4090 GPU. 


\subsection{$\ours$} \label{appendix:B:ours}

\subsubsection{Model Architecture} \label{section:appendix:b:model_archi}
\noindent\textbf{Skill encoder and skill prior.} $q_{\theta_\text{e}}(z|\tau)$ and $p_{\theta_\text{p}}(z|s_{t-h:t})$ are each implemented by a 2-layer LSTM network with a hidden dimension of 512. The skill horizon $h$ is set to 8, and the skill embedding dimension is 128 for MimicGen and 256 for both DexMimicGen and Bi-DexHands.

\noindent\textbf{Skill diffusion planner.} $\epsilon_{\theta_\text{d}}(\tau^k,k,z)$ is implemented by a U-Net~\citep{unet} architecture with 6 residual blocks, and diffusion step $k$ is encoded through a 2-layer MLP with a sinusoidal activation. The skill diffusion planner predicts the $h$-horizon skill trajectory following prior works~\citep{DP:Diffuser, DP:LDuS}, and we adopt the v-parameterization~\citep{DM:MuLAN}.

\noindent\textbf{Skill-conditioned noise schedule.} $\gamma_\phi(k,z)$ is factorized as $f(k;g_\phi(z))$, where $g_\phi(z)$ is implemented by a 4-layer MLP that produces the coefficients of a monotonic fifth-degree polynomial and $f(\cdot)$ evaluates the corresponding polynomial at diffusion step $k$. The noise schedule is bounded at its endpoints, with $\gamma_\phi(0,z) = \gamma_\text{min} = -5$ and $\gamma_\phi(1,z) = \gamma_\text{max} = 5$. Further details follow the implementation of \citet{DM:MuLAN}; we refer the reader to Appendix E.2 of the original paper.

\noindent\textbf{Inverse schedule.} $\gamma_\psi(\text{Leg}(\gamma'_\phi(k,z)))$ is implemented by a Set Transformer~\citep{set_transformer} over action dimensions, each represented by its Legendre projection coefficients. Treating the action dimensions as an unordered set prevents the inverse schedule from minimizing the loss via positional shortcuts, and instead forces it to rely on the Legendre coefficients. 
Moreover, rather than directly reconstructing the skill embedding, we maintain a codebook of prototype skill embeddings and train the inverse network to predict the nearest prototype, which we found stabilizes training. This is because skills tend to recur across trajectories and form clusters in the embedding space, making prototype-based prediction more natural than continuous regression. We find that this design yields improvements of $2.8$ pp on MimicGen and $3.9$ pp on DexMimicGen over directly regressing $z$. The number of prototype skill embeddings is set to either $8$ or $16$ depending on the task.


\subsubsection{Legendre Polynomial Basis and Projection} \label{sec:appendix:b:legendre}
In this section, we provide detailed background on Legendre polynomials and describe how we use them to project the schedule's derivative $\gamma'_\phi(k,z)$ onto a finite-dimensional representation suitable for the inverse schedule module $\gamma_\psi$.

\noindent\textbf{Legendre polynomials}~\citep{legendre} $\{P_n(x)\}_{n=0}^{\infty}$ form a complete orthogonal basis for square-integrable functions on the interval $[0, 1]$\footnote{Obtained by the change of variable $u = 2x -1$ for the Legendre polynomials defined on $[-1,1]$}, i.e., the $L^2[0,1]$ space. The Legendre polynomials can be calculated from Rodrigues' formula, 
\begin{equation}
    P_n(x) = \frac{1}{n!} \frac{d^n}{dx^n}\left[(x^2 - x)^n\right],
\end{equation}
and satisfy the orthogonality relation,
\begin{equation}
    \int_{0}^{1} P_m(x) P_n(x) \, dx = \frac{1}{2n+1} \delta_{mn}, 
\end{equation}
where $\delta_{mn}$ is the Kronecker delta; i.e., $\delta_{mn} = 1 \text{~if~} m=n$, otherwise, $\delta_{mn}=0$.
Then, the first five Legendre polynomials are as follows:
\begin{equation}
    \begin{aligned}
        P_0(x) &= 1, \\
        P_1(x) &= 2x - 1, \\
        P_2(x) &= 6x^2 - 6x + 1, \\
        P_3(x) &= 20x^3 - 30x^2 + 12x - 1, \\
        P_4(x) &= 70x^4 - 140x^3 + 90x^2 - 20x + 1.
    \end{aligned}
\end{equation}

\noindent\textbf{Legendre projection of noise schedule.} For each skill embedding $z$ and action dimension $d$, the schedule derivative $\gamma'_\phi(k,z)_d: [0,1] \rightarrow \mathbb{R}$ can be expanded as 
\begin{equation}
    \gamma'_\phi(k, z)_d = \sum_{n=0}^{\infty} c_n^{(d)}(z) \, P_n(k),
\end{equation}
where $c_n^{(d)}(z) = (2n+1) \int_{0}^{1} \gamma'_\phi(k, z)_d \, P_n(k) \, dk$.
In practice, we use Legendre coefficients up to order $N_\text{Leg}$, excluding the term $c_0$ as it is constant due to the fixed boundary values of the schedule. The Legendre projection operator $\text{Leg}(\cdot)$ is then defined as 
\begin{equation}
    \text{Leg}(\gamma'_\phi(\cdot, z)) = \left[c_1^{(d)}(z), c_2^{(d)}(z), \ldots, c_{N_{\text{Leg}}}^{(d)}(z)\right]_{d=1}^{N} \in \mathbb{R}^{N_\text{Leg} \times N},
\end{equation}
where $N$ is the action dimensionality and $N_\text{Leg}$ is set to 4 throughout our experiments.
%
Since the schedule  is a monotone function whose endpoints $\gamma_\phi(0,z)$ and $\gamma_\phi(1,z)$ are fixed, we project the derivative $\gamma'_\phi$ rather than the schedule $\gamma_\phi$ itself. 
Given that our schedule function is parameterized as a polynomial, its Legendre coefficients decay exponentially~\citep{legendre_decay}, so a small number of basis functions suffices for accurate approximation.

We adopt Legendre projection, since a small number of Legendre coefficients suffices to capture the polynomial schedule's geometry, providing a suitable input representation to $\gamma_\psi$. 
An alternative would be to feed sampled values $\gamma_\phi(k_i,z)$ at multiple discrete points $\{k_i\}$ directly to $\gamma_\psi$, resulting in a larger input space and degraded performance in our ablation study (see $\ours$-Legendre variant in Table~\ref{tab:ablation:1}).
Moreover, the orthogonality of Legendre polynomials ensures that each coefficient $c_n$ captures an independent aspect of the schedule's shape, producing a discriminative representation that distinguishes schedules with different geometric structures.
Other alternatives could be B-splines and Chebyshev polynomials. B-splines, however, require $d + c$ coefficients for degree $d$ and $c$ intervals. Matching the expressiveness of our fourth-degree Legendre projection requires $d \geq 4$, so any B-spline representation uses more coefficients than ours.
Chebyshev polynomials, an orthogonal basis of the same size, therefore remain a viable alternative.
Yet, we empirically found that Legendre polynomials yield better performance, and we therefore adopt them.

\begin{algorithm}[t]
\caption{Schedule-Driven Adaptive Denoising}
\label{alg:adaptive_denoising}
\begin{algorithmic}[1]
\STATE \textbf{Inputs}: Dataset $\mathcal{D}$, skill prior $p_{\theta_\text{p}}$, noise schedule $\gamma_\phi$, number of clusters $M$, candidate denoising step set $\mathcal{K}=\{K_{\min},\ldots,K_{\max}\}$
\STATE Initialize $\mathcal{B} \gets \varnothing$
\FOR{each state sequence $s_{t-h:t}$ obtained from $\mathcal{D}$}
    \STATE Sample skill embedding $z \sim p_{\theta_\text{p}}(\cdot \mid s_{t-h:t})$
    \STATE Compute Legendre coefficients $\bm{b} \gets \text{Leg}(\gamma'_\phi(k, z))$
    \STATE $\mathcal{B} \gets \mathcal{B} \cup \{(z, \bm{b})\}$
\ENDFOR
\STATE Cluster $\mathcal{B}$ into $M$ clusters $\{\mathcal{C}_1, \ldots, \mathcal{C}_M\}$ via K-means on Legendre coefficients
\FOR{each cluster $\mathcal{C}_m$}
    \FOR{each $K \in \mathcal{K}$ in decreasing order}
        \STATE Compute reconstruction MSE on skills in $\mathcal{C}_m$ with $K$ denoising steps
    \ENDFOR
    \STATE $K_m^* \gets$ smallest $K$ before sharp MSE increase (via Kneedle Algorithm)
\ENDFOR
\end{algorithmic}
\end{algorithm}

\subsubsection{Hyperparameters} \label{appendix:b:ours_hyperparam}
For all experiments, we use the Adam optimizer with a learning rate of $1 \times 10^{-4}$, and exponential decay rates of $\beta_1=0.9, \beta_2=0.99$. The weight for the skill encoder regularization loss in Equation~\eqref{loss:diffusion} ($\beta$) is set to $1 \times 10^{-3}$, and the inverse loss $\mathcal{L}_{\text{inv}}$ in Equation~\eqref{loss:inverse} is scaled by a factor of $1 \times 10^{-2}$ ($\eta$ in Algorithm~\ref{algo:SaND}). 
%
We use a batch size of 256 and clip gradients at a maximum norm of 1.0. All models are trained for 200,000 or 400,000 steps, depending on the maximum episode length of the task, with a linear learning rate warm-up over the first 10,000 steps. At inference, we use DDIM~\citep{DM:DDIM} sampling with 10 denoising steps by default. For schedule-driven adaptive denoising, we sweep the number of denoising steps $K \in \{3,4,5,6\}$ for MimicGen and $K \in \{4,5,6,7,8,9,10\}$ for DexMimicGen. The number of clusters is set to 8 for MimicGen and 24 for DexMimicGen.

\subsection{Baselines}
The baselines in our experiments fall into three categories based on their action generation mechanism: \textbf{[a]} action-level (i.e., BC-RNN, DiffBC, DP, OneDP), \textbf{[t]} trajectory-level (i.e., Diffuser, DV), and \textbf{[s]} skill-level (i.e., SkillDiffuser, LDuS). Except for BC-RNN, all baselines employ a diffusion model for the policy. 

\textbf{[a]} \textbf{BC-RNN}~\citep{dataset:robomimic} serves as a non-diffusion baseline, implementing the policy via a recurrent neural network that maintains a history of 10 past observations. For MimicGen, we use a Gaussian mixture model to model the action distribution, while for DexMimicGen we use a single Gaussian distribution. 

\textbf{[a]} \textbf{DiffBC}~\citep{DP:DiffBC} is the most straightforward application of diffusion models to imitation learning, where a diffusion model conditions solely on the current observation and denoises a single action at each timestep. 
The denoising network is implemented as a 4-layer MLP with 512
hidden units for MimicGen, and as a multi-head attention network with 4 transformer blocks and 16 attention heads for DexMimicGen and Bi-DexHands. We adopt $x_0$-parameterization with a cosine schedule, training with 300 diffusion steps and accelerating inference via DDIM~\citep{DM:DDIM} with 10 steps.

\textbf{[a]} \textbf{DP}~\citep{DP:DP} extends diffusion-based imitation learning by predicting a chunk of future actions at each timestep, enabling temporally consistent action generation through receding horizon control. 
The denoising network is implemented as a U-Net~\citep{unet}. We set the prediction horizon $T_p=16$, executing $T_a=8$ actions per replan. All remaining training and inference settings follow DiffBC.

\textbf{[a]} \textbf{OneDP}~\citep{DP:OneDP} distills a pretrained diffusion policy into a one-step action generator via score distillation, training the student to match the teacher's score along the diffusion chain. We adopt the stochastic variant (OneDP-S), which alternately updates an auxiliary score network and the actor, both warm-started from the teacher's weights. All architecture and receding-horizon settings are inherited from the DP model described above, with the number of denoising steps set to 1.

\textbf{[t]} \textbf{Diffuser}~\citep{DP:Diffuser} is a pioneering work that leverages diffusion models as a trajectory planner. It jointly denoises an $h$-length sequence of state-action pairs while replacing the first state with the current observation. Such joint modeling of states and actions over the full horizon enables the model to enforce only local consistency within each denoising step. This local consistency accumulates into global coherence of the sampled plan after the full denoising chain. 
The denoising network architecture and all remaining training and inference settings follow DP. We set the horizon $h=8$.

\textbf{[t]} \textbf{DV}~\citep{DP:DV} is the state-of-the-art diffusion-based planning model. Unlike Diffuser, DV's diffusion component is only responsible for planning desired trajectories of future states. The actions are derived via a learned inverse dynamics model that recovers actions from consecutive future states. The denoising network is implemented as a DiT~\citep{dit} with 4 Transformer blocks and 4 attention heads, and all remaining training and inference settings follow DiffBC.

\textbf{[s]} \textbf{SkillDiffuser}~\citep{DP:SkillDiffuser} adopts a diffusion model within a hierarchical skill-based imitation learning framework~\citep{sprl:fist}. Such a hierarchical structure is especially tailored for long-horizon robotic manipulation tasks, as it decomposes the task into a sequence of skills, reducing the burden of modeling long-range dependencies directly in the action space. It uses a skill encoder and a skill decoder to learn a skill embedding, where the decoder is implemented as a diffusion model. Specifically, the diffusion-based skill decoder reconstructs a state sequence conditioned on the skill embedding, which is then passed through an inverse dynamics model to produce the corresponding action sequence. The denoising network is implemented as a U-Net.

\textbf{[s]} \textbf{LDuS}~\citep{DP:LDuS} is another hierarchical skill-based imitation learning framework that employs a diffusion model as the skill decoder. Unlike SkillDiffuser, LDuS directly denoises sequences of state-action pairs, eliminating the need for a separate inverse dynamics model. The denoising network is implemented as a U-Net structure.

For all baselines, we use the same optimizer and training budgets as $\ours$ in Appendix~\ref{appendix:b:ours_hyperparam}.



\subsection{Multimodal Encoder}
To handle the multimodal observation space for DexMimicGen in Table~\ref{tab:main:1}, we pre-train a multimodal encoder and freeze its parameters during policy learning. While end-to-end joint training of the multimodal encoder and policy is possible, it incurs prohibitive computational overhead; we therefore adopt the decoupled pre-training strategy. 

We employ a separate ResNet-18 encoder for each image modality and an MLP-based encoder for the proprioceptive sensor inputs~\citep{dataset:robomimic}. The resulting embeddings are concatenated and passed through an MLP projection layer to obtain a latent representation with a hidden dimension of 128.
To train the multimodal encoder $\Phi$, we attach an inverse dynamics head $g$ on top of the latent representation. The encoder and the inverse dynamics head are jointly optimized to predict the action given two consecutive observations, i.e.,
\begin{equation} \label{equ:idm_loss}
\begin{gathered}
    \hat{a}_t = g(\Phi(s_t),\Phi(s_{t+1})) \\
    \mathcal{L}_{\text{idm}}(g, \Phi) = \mathbb{E}_{(s_t,a_t,s_{t+1})\sim\mathcal{D}}\left[ \|\hat{a}_t - a_t\|_2^2 \right].
\end{gathered}
\end{equation}

We apply the same pre-training strategy to the Bi-DexHands benchmark, despite it using only low-dimensional observations (with no image modality). This is due to the high dimensionality of the state in Bi-DexHands, which exceeds 400 dimensions; the pre-trained encoder serves to compress this into a more manageable representation. 
As no image modalities are present, we use only an MLP-based encoder to project the input down to a 64-dimensional latent representation, and randomly mask $s_{t+1}$ with a probability of $20\%$ during training.
The same encoder is shared across $\ours$ and all baselines.

\subsection{VLA Settings}
For our VLA experiments in Figure~\ref{fig:case:vla} of Section~\ref{sec:exp:case}, we adopt SigLIP-2~\citep{siglip-2} as the vision-language foundation model and finetune only its last 4 layers to keep computational overhead manageable. 
%
We leverage the same inverse dynamics loss in Equation~\eqref{equ:idm_loss} to finetune the model.
We train and evaluate on three DexMimicGen tasks (Box Cleanup, Drawer Cleanup, and Tray Lift) that share the same 24-dimensional action space, enabling joint multi-task learning. As the images in the datasets are $84 \times 84$ in size, we resize them to $224 \times 224$ to match SigLIP-2's input resolution. Regarding the language modality, each task is paired with a natural language instruction describing the goal, as listed in Appendix~\ref{appendix:A:1}.

\newpage

\section{Additional Experiments}

\subsection{Visualization of Legendre Coefficients}\label{appendix:c:1}
\begin{figure}[h]
    \centering
    \subfigure[Threading Task (MimicGen)]{
        \centering
        \label{fig:appendix:C:tsne_visualization:1}
        \includegraphics[width=.98\linewidth]{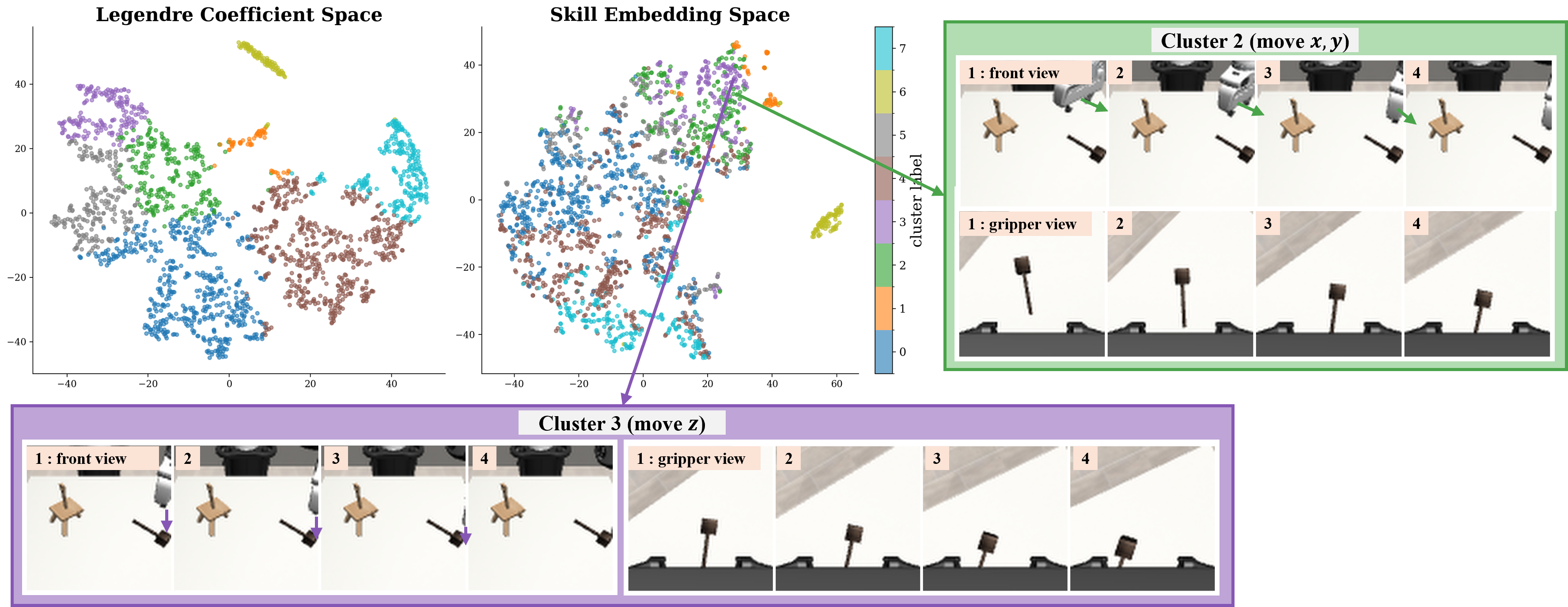}
    }
    \\
    \subfigure[Tray Lift Task (DexMimicGen)]{
        \centering
        \label{fig:appendix:C:tsne_visualization:2}
        \includegraphics[width=.98\linewidth]{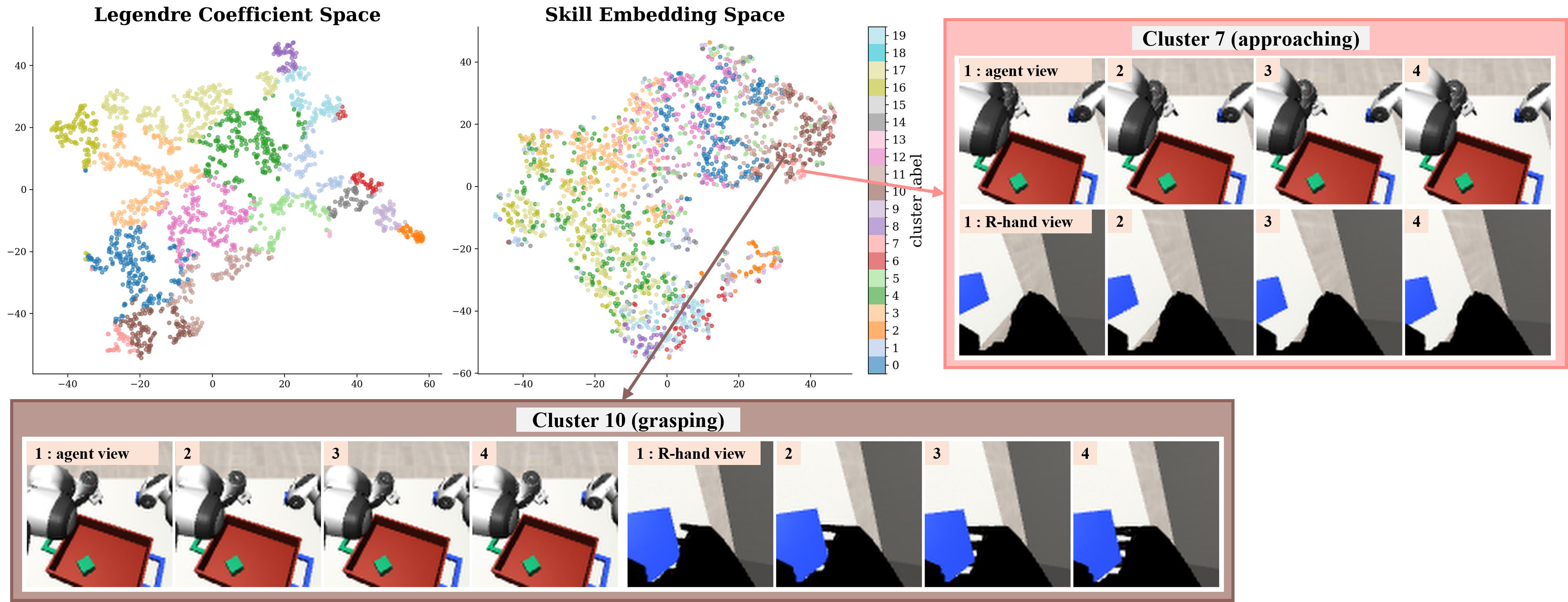}
    }
    \caption{t-SNE visualization of the Legendre coefficients (left panel) and skill embeddings (right panel): Colors indicate K-means cluster labels computed in the Legendre coefficient space and shared across both panels; identical colors indicate skills with similar noise-schedule shapes.}
    \label{fig:appendix:C:tsne_visualization}
\end{figure}

In Figure~\ref{fig:appendix:C:tsne_visualization}, we visualize the t-SNE embeddings of the Legendre coefficients $\text{Leg}(\gamma'_\phi)$ (left panel) and the skill embeddings $z$ (right panel) for the Threading (MimicGen) and Tray Lift (DexMimicGen) tasks. We perform K-means clustering in the Legendre coefficient space, and the resulting cluster labels are used to color the points in both panels -- i.e., points sharing the same color correspond to skills with similar noise schedule geometry. 

As shown in both Threading and Tray Lift tasks, the cluster labels are occasionally entangled in the skill embedding space, indicating that neighboring skill embeddings may belong to different noise schedule clusters. 
For example, in the Threading task (Figure~\ref{fig:appendix:C:tsne_visualization:1}), cluster 2 (\textcolor{ForestGreen}{green}) and cluster 3 (\textcolor{BlueViolet}{violet}) appear entangled as their corresponding skill embeddings are temporally consecutive: cluster 2 corresponds to planar movements along the $x$- and $y$-axes, while cluster 3 additionally involves downward motion along the $z$-axis. However, these two clusters are well-separated in the Legendre coefficient space. 
In the Tray Lift task (Figure~\ref{fig:appendix:C:tsne_visualization:2}), cluster 7 (\textcolor{CarnationPink}{peach}) corresponds to the skill approaching the blue cube, while cluster 10 (\textcolor{brown}{brown}) represents the subsequent skill making contact with the blue cube in preparation for lifting. Similarly, the temporal continuity between these skills leads to their entanglement in the skill embedding space, while they remain clearly disentangled in the Legendre coefficient space. 

Such observations offer a direct intuition for why Legendre-based clustering yields better performance than skill embedding-based clustering for adaptive denoising, as reported in Table~\ref{tab:ablation:2}.



%

\subsection{Performance Ceiling with Increasing Denoising Steps} \label{sec:appendix:c:step}
\begin{figure}[h]
    \centering
    \subfigure[Coffee Preparation Task (MimicGen)]{
        \centering
        \label{fig:appendix:C:step:1}
        \includegraphics[width=.4\linewidth]{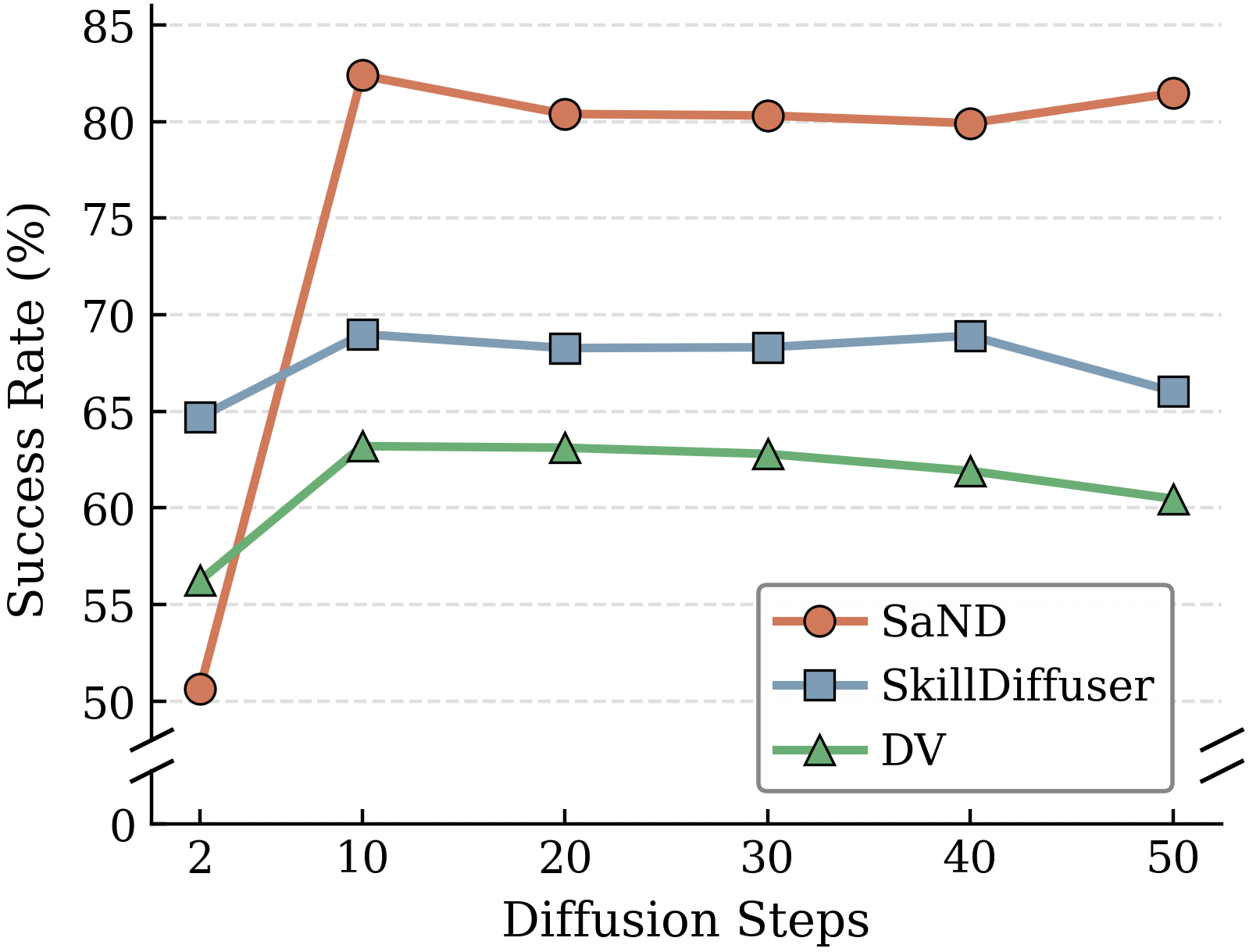}
    }
    \subfigure[Can Sort Task (DexMimicGen)]{
        \centering
        \label{fig:appendix:C:step:2}
        \includegraphics[width=.4\linewidth]{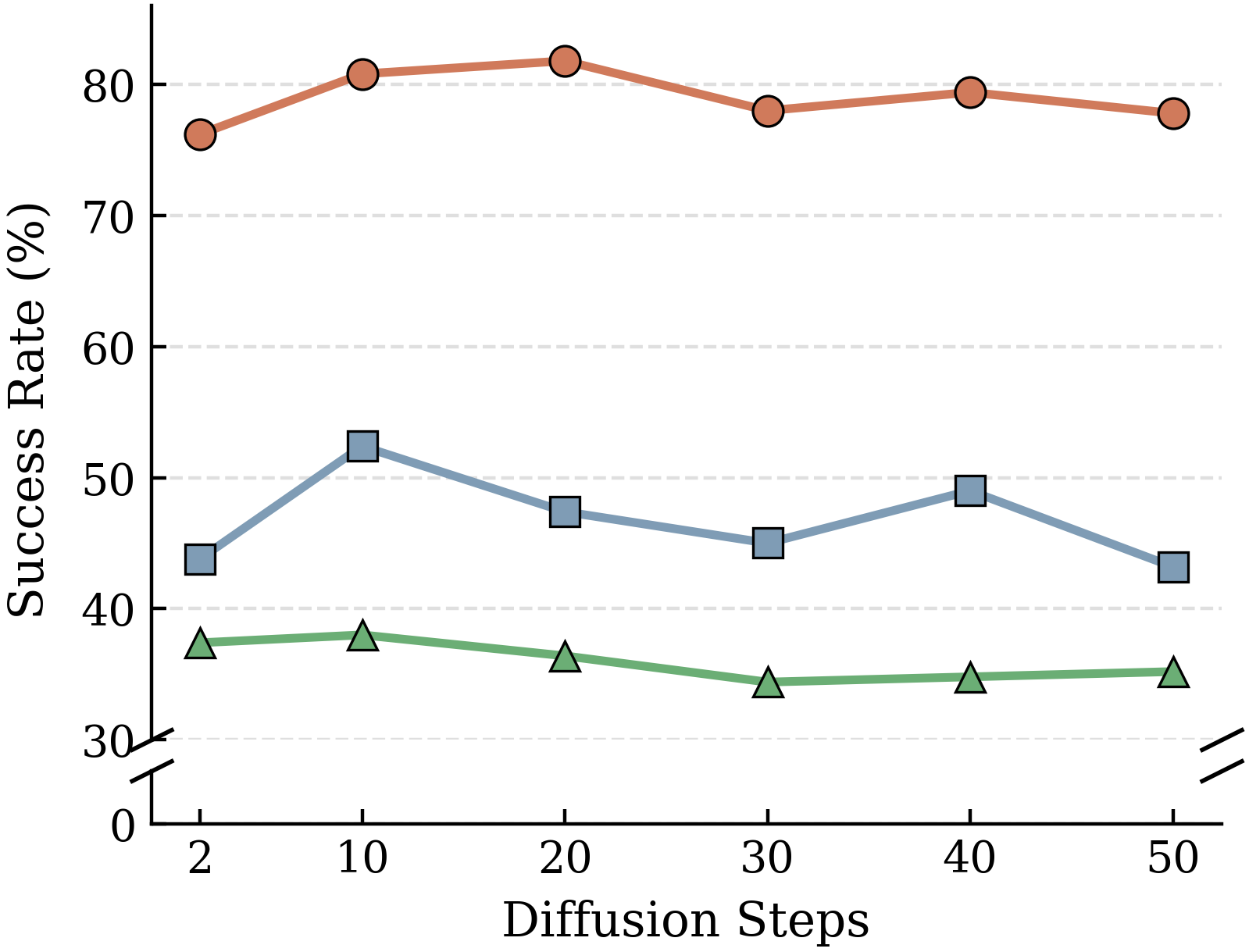}
    }
    \caption{Performance with respect to the number of denoising steps}
    \label{fig:appendix:C:step}
\end{figure}

In Figure~\ref{fig:appendix:C:step}, we visualize the performance of $\ours$, SkillDiffuser, and DV with respect to the number of denoising steps $\{2, 10, 20, 30, 40, 50\}$ on the Coffee Preparation (MimicGen) and Can Sort (DexMimicGen) tasks. 
For both tasks, the success rates of all three methods saturate as the step budget grows from $2$ to $10$, with no meaningful improvement beyond $10$ steps. Yet, $\ours$ attains a markedly higher ceiling than the baselines (SkillDiffuser, DV). These gains stem from jointly learning a skill-conditioned noise schedule with the skill diffusion planner, which tightens the ELBO at its source rather than relying on a fixed noise schedule~\citep{DM:MuLAN}. This result confirms that the attainable performance ceiling is governed by the tightness of the bound, highlighting the importance of learning the noise schedule.
%

We further analyze why simply increasing the number of denoising steps is insufficient, and motivate the need for learning a noise schedule from a theoretical perspective. 
From the continuous-time variational inference perspective~\citep{huang2021variational}, the Brownian motion driving the forward diffusion can be treated as an infinite-dimensional latent variable. Applying variational inference over this latent variable, analogous to a Variational AutoEncoder (VAE) with an infinite-dimensional code, yields a continuous-time ELBO (CT-ELBO) that lower-bounds the log-likelihood. 
%


%
The variational gap between the CT-ELBO and the true log-likelihood admits a closed form:
\begin{equation} \label{eq:elbo_infty}
\begin{gathered}
  \log p(x^1,1) = \underbrace{\int_0^1 \mathbb{E}\!\left[ \|a(\bm{x}^k,k) - \sigma^\top\nabla_{\bm{x}}\log p(\bm{x}^k,\,1-k)\|^2   \right]dk}_{\text{reducible by increasing diffusion steps}} +\underbrace{\mathcal{E}_\infty}_{\substack{\text{irreducible by} \\ \text{increasing diffusion steps}}}, \\
  \text{where} \ \mathcal{E}_\infty
  = \mathbb{E} \left[-\frac{1}{2}\int_0^1 \|a(\bm{x}^k,k)\|^2 dk + \log p_0(\bm{x}^1) - \int_0^1 \nabla_{\bm{x}} \cdot\mu(\bm{x}^k,k) dk    \right].
\end{gathered}
\end{equation}

Here $a$ is the drift of the inference SDE, the counterpart of a VAE, i.e., 
$a \approx \sigma^{\top}(k)\nabla_{\bm{x}} \log p(\bm{x})$\footnote{Here, $\sigma^{\top}$ denotes the transpose of $\sigma$.} and 
$\mu(\bm{x}^k, k) :=\frac{1}{2} (\partial_k \log \alpha^k) \left[ \bm{x}^k + 2 \nabla_{\bm{x}} \log p(\bm{x}^k) \right]$. 

That is, the gap is the time integral of the squared error between the inference drift $a$ and the optimal drift $\sigma^\top\nabla_{\bm{x}}\log p$ (the ground truth score) over the noise schedule, and vanishes if and only if $a$ recovers the true score exactly. Increasing the number of denoising steps only refines the numerical approximation of this integral; it cannot reduce the integrand itself, which is the point-wise score error at each timestep. 
Fortunately, the scheduler can affect $a$ and $\mu$ of $\mathcal{E}_{\infty}$ in Equation~\eqref{eq:elbo_infty}, tightening the ELBO even with a smaller step budget.

 
\color{blue}
\color{black}

\subsection{Additional Benchmark Experiments} \label{appendix:bidexhands}
Our method targets contact-rich and dexterous manipulation, where action dimensions are tightly coupled by contact constraints and the coupling patterns differ across skills. This is precisely the regime in which a skill-conditioned noise schedule is most advantageous, motivating our selection of MimicGen, DexMimicGen, and Bi-DexHands as primary benchmarks.
To verify that $\ours$ also maintains consistent performance on standard diffusion policy 
benchmarks, we additionally evaluate on Push-T~\citep{DP:DP}. The observation consists of a $96 \times 96$ RGB image with the agent pose, and the action is a 2-dimensional target position.
We use the official dataset of 206 expert demonstrations without modification. 
As shown in Table~\ref{tab:push-T}, $\ours$ marginally outperforms DP, though the gap is negligible. This is expected, as the 2-dimensional action space offers limited cross-dimensional structure for a per-dimension schedule to exploit, and performance on this benchmark is already near saturation.

\begin{table}[h]
\caption{Performance on the Push-T benchmark}
\label{tab:push-T}
\centering
\begin{adjustbox}{width=.30\textwidth}
\begin{tabular}{l|c|
>{\centering\arraybackslash}p{0.14\linewidth}}
\toprule
\textbf{Methods} & \textbf{$K$} & \textbf{Push-T} \\
\midrule
DP & 10
& 0.9145{\color{gray}\scriptsize$\pm$0.0075} 
\\
\textbf{$\ours$} & 10
& \textbf{0.9195{\color{gray}\scriptsize$\pm$0.0005}} \\
\bottomrule
\end{tabular}
\end{adjustbox}
\end{table}

\subsection{Analysis on Inference Cost}
In Table~\ref{tab:cost}, we measure the inference cost of $\ours$ on a single RTX 4090, reporting FLOPs and wall-clock time per skill trajectory $\tau$ generation. We compare three variants: $\ours$, $\ours$+AD, which applies adaptive denoising, and $\ours$-Schedule, which replaces the learned noise schedule with a fixed one (the ablation variant in Table~\ref{tab:ablation:1}).
The learned noise schedule adds $0.31\%$ in FLOPs and $1.7\%$ in wall-clock time over $\ours$-Schedule, supporting the negligible-overhead claim in Section~\ref{sec:method:adaptive_denoising}. 
Adaptive denoising reduces FLOPs by $56.8\%$ and wall-clock time by $48.7\%$ relative to $\ours$, confirming that fewer denoising steps translate into lower inference cost.

\begin{table}[h]
\caption{Analysis on inference cost}
\label{tab:cost}
\centering
\begin{adjustbox}{width=.46\textwidth}
\begin{tabular}{l|cc}
\toprule
\textbf{Methods} & \textbf{FLOPs} & \textbf{Wall-clock Time} \\
\midrule
\textbf{$\ours$}     & 777.1 M & 5.065 ms \\
\textbf{$\ours$+AD}  & 335.4 M & 2.599 ms \\ 
$\ours$-Schedule & 774.6 M & 4.978 ms \\
\bottomrule
\end{tabular}
\end{adjustbox}
\end{table}

\subsection{Additional Ablation Results on Skill-Adaptive Noise Diffusion}
\label{appendix:C:ablation_sand}
Table~\ref{tab:ablation:3} presents an additional ablation study to determine whether the gains of $\ours$ require both skill conditioning and per-dimension scheduling.
\textbf{$\ours$-SkillCond} removes skill conditioning and learns a single schedule shared across all skills. 
\textbf{$\ours$-Global} learns a single noise schedule shared across all action dimensions, removing dimension-wise scheduling.
Removing skill conditioning results in an average drop of $20.4$ pp, as the schedule can only fit the average coupling pattern rather than skill-specific ones.
Similarly, removing dimension-wise scheduling leads to a $23.3$ pp drop, as a single schedule must serve all action dimensions simultaneously, causing conflicting per-dimension gradients. 
These results confirm that both mechanisms are independently essential.

\begin{table}[h]
\caption{Additional ablation on skill-adaptive noise diffusion}
\label{tab:ablation:3}
\centering
\begin{adjustbox}{width=.96\textwidth}
\begin{tabular}{l|
>{\centering\arraybackslash}p{0.1\linewidth}
>{\centering\arraybackslash}p{0.1\linewidth}
>{\centering\arraybackslash}p{0.1\linewidth}
>{\centering\arraybackslash}p{0.1\linewidth}
>{\centering\arraybackslash}p{0.1\linewidth}
>{\centering\arraybackslash}p{0.1\linewidth}
|
>{\centering\arraybackslash}p{0.1\linewidth}
}
\toprule
\multicolumn{8}{l}{\textbf{Benchmark: Contact-Rich Manipulation (MimicGen~\cite{dataset:mimicgen}) -- Low Dimensional Observation}}  \\
\midrule
\textbf{Methods} & \textbf{Square} & \textbf{Threading} & \textbf{Mug} & \textbf{Hammer} & \textbf{Coffee} & \textbf{Kitchen} & \textbf{Average} \\
\midrule
\textbf{$\ours$}
& \textbf{60.9{\color{gray}\scriptsize$\pm$1.9}} 
& \textbf{66.9{\color{gray}\scriptsize$\pm$5.5}} 
& \textbf{60.0{\color{gray}\scriptsize$\pm$1.6}} & \textbf{58.4{\color{gray}\scriptsize$\pm$1.9}}
& \textbf{82.4{\color{gray}\scriptsize$\pm$2.2}} & \textbf{55.3{\color{gray}\scriptsize$\pm$4.4}} 
& \textbf{64.0{\color{gray}\scriptsize$\pm$1.0}}  \\
$\ours$-SkillCond
& 33.6{\color{gray}\scriptsize$\pm$2.9} & 33.6{\color{gray}\scriptsize$\pm$5.0} 
& 50.9{\color{gray}\scriptsize$\pm$4.5} & 40.6{\color{gray}\scriptsize$\pm$2.2}
& 64.3{\color{gray}\scriptsize$\pm$1.9} & 38.4{\color{gray}\scriptsize$\pm$3.9} 
& 43.6{\color{gray}\scriptsize$\pm$1.8}  \\
$\ours$-Global  
& 30.3{\color{gray}\scriptsize$\pm$3.8} & 36.1{\color{gray}\scriptsize$\pm$10.2} 
& 45.6{\color{gray}\scriptsize$\pm$10.9} & 42.0{\color{gray}\scriptsize$\pm$4.4}
& 54.2{\color{gray}\scriptsize$\pm$28.0} & 35.9{\color{gray}\scriptsize$\pm$1.6} 
& 40.7{\color{gray}\scriptsize$\pm$4.1}  \\
\bottomrule
\end{tabular}
\end{adjustbox}
\end{table}

\subsection{Detailed Ablation Results on Schedule-Driven Adaptive Denoising} \label{sec:appendix:c:ablation_ad}
Here, we present detailed ablation results on schedule-driven adaptive denoising across individual tasks, complementing Figure~\ref{fig:ablation:2} of Section~\ref{sec:exp:ablation}. Specifically, we compare fixed-$K$ variants of $\ours$ and $\ours$+AD with skill embedding-based clustering. Note that the highest performance is highlighted in \textbf{bold}, and the second highest is \underline{underlined}.
As shown, for all cases except the Coffee task in MimicGen, $\ours$+AD (Leg) demonstrates improved performance over fixed-$K$ variants, confirming that the performance gain stems from dynamically adapting $K$ per skill according to the corresponding Legendre coefficient cluster. Moreover, the alternative variant with skill embedding-based clustering, $\ours$+AD (skill), consistently yields degraded performance compared to Legendre-based clustering, further validating our design choice.

\begin{table}[h]
\caption{Detailed ablation on schedule-driven adaptive denoising}
\label{tab:ablation:2}
\centering
\begin{adjustbox}{width=1.\textwidth}
\begin{tabular}{l|c|
>{\centering\arraybackslash}p{0.1\linewidth}
>{\centering\arraybackslash}p{0.1\linewidth}
>{\centering\arraybackslash}p{0.1\linewidth}
>{\centering\arraybackslash}p{0.1\linewidth}
>{\centering\arraybackslash}p{0.1\linewidth}
>{\centering\arraybackslash}p{0.1\linewidth}
|
>{\centering\arraybackslash}p{0.1\linewidth}
}
\toprule
\multicolumn{8}{l}{\textbf{Benchmark: Contact-Rich Manipulation (MimicGen~\cite{dataset:mimicgen}) -- Low Dimensional Observation}}  \\
\midrule
\textbf{Methods} & \textbf{Metric} & \textbf{Square} & \textbf{Threading} & \textbf{Mug} & \textbf{Hammer} & \textbf{Coffee} & \textbf{Kitchen} & \textbf{Average} \\
\midrule
\textbf{$\ours$} ($K=10$) & SR
& 60.9{\color{gray}\scriptsize$\pm$1.9} 
& \textbf{66.9{\color{gray}\scriptsize$\pm$5.5}} 
& \textbf{60.0{\color{gray}\scriptsize$\pm$1.6}}
& {58.4{\color{gray}\scriptsize$\pm$1.9}}
& \textbf{82.4{\color{gray}\scriptsize$\pm$2.2}} 
& \textbf{55.3{\color{gray}\scriptsize$\pm$4.4}} 
& \textbf{64.0{\color{gray}\scriptsize$\pm$1.0}}  \\
\textbf{$\ours$} ($K=4$) & SR
& 60.2{\color{gray}\scriptsize$\pm$1.0} & 64.0{\color{gray}\scriptsize$\pm$4.1} 
& 51.0{\color{gray}\scriptsize$\pm$2.7} 
& \underline{59.4}{\color{gray}\scriptsize$\pm$2.2}
& \underline{81.9}{\color{gray}\scriptsize$\pm$1.6} & 53.8{\color{gray}\scriptsize$\pm$1.8} 
& 61.7{\color{gray}\scriptsize$\pm$0.4}  \\
\textbf{$\ours$} ($K=3$) & SR
& 58.0{\color{gray}\scriptsize$\pm$3.6} & 60.4{\color{gray}\scriptsize$\pm$3.9} 
& 54.4{\color{gray}\scriptsize$\pm$6.0} & 58.0{\color{gray}\scriptsize$\pm$2.4}
& 78.5{\color{gray}\scriptsize$\pm$3.8} & 50.6{\color{gray}\scriptsize$\pm$3.5} 
& 60.0{\color{gray}\scriptsize$\pm$1.3}  \\
\midrule
\multirow{2}{*}{\textbf{$\ours$+AD} (skill)} & SR
& \underline{61.4}{\color{gray}\scriptsize$\pm$3.0} 
& 64.5{\color{gray}\scriptsize$\pm$3.0} 
& 55.7{\color{gray}\scriptsize$\pm$7.2} 
& 58.9{\color{gray}\scriptsize$\pm$2.0}
& 80.8{\color{gray}\scriptsize$\pm$3.6} 
& 54.6{\color{gray}\scriptsize$\pm$3.6} 
& 62.6{\color{gray}\scriptsize$\pm$1.6}  \\
& $K$
& \cellcolor{gray!20} 3.7 {\color{gray}\scriptsize$\pm$0.2} 
& \cellcolor{gray!20} 3.5 {\color{gray}\scriptsize$\pm$0.3} 
& \cellcolor{gray!20} 3.6 {\color{gray}\scriptsize$\pm$0.4} 
& \cellcolor{gray!20} 3.9 {\color{gray}\scriptsize$\pm$0.2}
& \cellcolor{gray!20} 3.5 {\color{gray}\scriptsize$\pm$0.2} 
& \cellcolor{gray!20} 3.8 {\color{gray}\scriptsize$\pm$0.2} 
& \cellcolor{gray!20} 3.7 {\color{gray}\scriptsize$\pm$0.2} \\
\midrule
\multirow{2}{*}{\textbf{$\ours$+AD} (Leg)} & SR
& \textbf{61.7{\color{gray}\scriptsize$\pm$3.2}} 
& \textbf{66.9{\color{gray}\scriptsize$\pm$3.9}} 
& \underline{55.9}{\color{gray}\scriptsize$\pm$5.0} 
& \textbf{59.6{\color{gray}\scriptsize$\pm$2.2}}
& 81.0{\color{gray}\scriptsize$\pm$2.8} 
& \underline{54.6}{\color{gray}\scriptsize$\pm$3.6} 
& \underline{63.3}{\color{gray}\scriptsize$\pm$0.8}  \\
& $K$
& \cellcolor{gray!20} 3.7 {\color{gray}\scriptsize$\pm$0.1} 
& \cellcolor{gray!20} 3.4 {\color{gray}\scriptsize$\pm$0.2} 
& \cellcolor{gray!20} 3.5 {\color{gray}\scriptsize$\pm$0.6} 
& \cellcolor{gray!20} 4.0 {\color{gray}\scriptsize$\pm$0.6}
& \cellcolor{gray!20} 3.5 {\color{gray}\scriptsize$\pm$0.2} 
& \cellcolor{gray!20} 3.8 {\color{gray}\scriptsize$\pm$0.2} 
& \cellcolor{gray!20} 3.7 {\color{gray}\scriptsize$\pm$0.2} \\

\midrule\midrule
\multicolumn{9}{l}{\textbf{Benchmark: Bimanual Dexterous Manipulation (DexMimicGen~\cite{dataset:dexmimicgen}) -- Multimodal Observation}}  \\
\midrule
\textbf{Methods} & \textbf{Metric} & \textbf{Box} & \textbf{Drawer} & \textbf{Tray} & \textbf{Pouring} & \textbf{Coffee} & \textbf{Can} & \textbf{Average} \\
\midrule
\textbf{$\ours$} ($K=10$) & SR
& {79.4{\color{gray}\scriptsize$\pm$5.5}} 
& \textbf{81.0{\color{gray}\scriptsize$\pm$1.4}}
& \underline{40.6}{\color{gray}\scriptsize$\pm$6.0} 
& \textbf{71.0{\color{gray}\scriptsize$\pm$13.3}}
& \textbf{32.0{\color{gray}\scriptsize$\pm$4.6}} 
& \underline{80.8}{\color{gray}\scriptsize$\pm$8.7}
& \textbf{64.1{\color{gray}\scriptsize$\pm$2.7}} \\
\textbf{$\ours$} ($K=6$) & SR
& 78.0{\color{gray}\scriptsize$\pm$5.3} & 77.6{\color{gray}\scriptsize$\pm$4.5} 
& 36.4{\color{gray}\scriptsize$\pm$6.5} & 60.4{\color{gray}\scriptsize$\pm$13.3}
& 29.4{\color{gray}\scriptsize$\pm$5.9} & 79.8{\color{gray}\scriptsize$\pm$10.2} 
& 60.3{\color{gray}\scriptsize$\pm$3.4}  \\
\textbf{$\ours$} ($K=5$) & SR
& 79.0{\color{gray}\scriptsize$\pm$5.9} & 79.6{\color{gray}\scriptsize$\pm$4.1} 
& 38.2{\color{gray}\scriptsize$\pm$9.2} & 60.6{\color{gray}\scriptsize$\pm$13.8}
& 29.4{\color{gray}\scriptsize$\pm$7.3} & 80.2{\color{gray}\scriptsize$\pm$9.5} 
& 61.2{\color{gray}\scriptsize$\pm$1.8}  \\
\midrule
\multirow{2}{*}{\textbf{$\ours$+AD} (skill)} 
& SR
& \underline{80.0}{\color{gray}\scriptsize$\pm$6.6} 
& 79.0{\color{gray}\scriptsize$\pm$6.4}
& 38.8{\color{gray}\scriptsize$\pm$5.5} 
& 60.2{\color{gray}\scriptsize$\pm$11.8}
& 30.6{\color{gray}\scriptsize$\pm$7.9} 
& 78.6{\color{gray}\scriptsize$\pm$12.4} 
& 61.2{\color{gray}\scriptsize$\pm$3.3} \\
& $K$
& \cellcolor{gray!20} 5.4 {\color{gray}\scriptsize$\pm$0.2} 
& \cellcolor{gray!20} 5.4 {\color{gray}\scriptsize$\pm$0.2} 
& \cellcolor{gray!20} 5.2 {\color{gray}\scriptsize$\pm$0.1} 
& \cellcolor{gray!20} 5.7 {\color{gray}\scriptsize$\pm$0.3}
& \cellcolor{gray!20} 4.8 {\color{gray}\scriptsize$\pm$0.2} 
& \cellcolor{gray!20} 5.6 {\color{gray}\scriptsize$\pm$0.6} 
& \cellcolor{gray!20} 5.3 {\color{gray}\scriptsize$\pm$0.1} \\
\midrule
\multirow{2}{*}{\textbf{$\ours$+AD} (Leg)} & SR
& \textbf{83.6{\color{gray}\scriptsize$\pm$4.9}} 
& \underline{80.6}{\color{gray}\scriptsize$\pm$4.8}
& \textbf{42.0{\color{gray}\scriptsize$\pm$7.8}} 
& \underline{63.6}{\color{gray}\scriptsize$\pm$10.8}
& \underline{30.8}{\color{gray}\scriptsize$\pm$5.3} 
& \textbf{82.2{\color{gray}\scriptsize$\pm$8.9}} 
& \underline{63.8}{\color{gray}\scriptsize$\pm$1.9} \\
& $K$
& \cellcolor{gray!20} 5.4 {\color{gray}\scriptsize$\pm$0.2} 
& \cellcolor{gray!20} 5.3 {\color{gray}\scriptsize$\pm$0.3} 
& \cellcolor{gray!20} 5.1 {\color{gray}\scriptsize$\pm$0.1} 
& \cellcolor{gray!20} 5.9 {\color{gray}\scriptsize$\pm$0.1}
& \cellcolor{gray!20} 5.0 {\color{gray}\scriptsize$\pm$0.2} 
& \cellcolor{gray!20} 5.8 {\color{gray}\scriptsize$\pm$0.9} 
& \cellcolor{gray!20} 5.4 {\color{gray}\scriptsize$\pm$0.1} \\
\bottomrule
\end{tabular}
\end{adjustbox}
\end{table}
%






\newpage
\section*{NeurIPS Paper Checklist}

\begin{enumerate}

\item {\bf Claims}
    \item[] Question: Do the main claims made in the abstract and introduction accurately reflect the paper's contributions and scope?
    \item[] Answer: \answerYes{} 
    \item[] Justification: The abstract and introduction clearly state the paper's main contributions.

\item {\bf Limitations}
    \item[] Question: Does the paper discuss the limitations of the work performed by the authors?
    \item[] Answer: \answerYes{}{} 
    \item[] Justification: The limitations of our work are discussed in Section~\ref{sec:conclusion}.

\item {\bf Theory assumptions and proofs}
    \item[] Question: For each theoretical result, does the paper provide the full set of assumptions and a complete (and correct) proof?
    \item[] Answer: \answerNA{} 
    \item[] Justification: Our paper does not include theoretical results.

    \item {\bf Experimental result reproducibility}
    \item[] Question: Does the paper fully disclose all the information needed to reproduce the main experimental results of the paper to the extent that it affects the main claims and/or conclusions of the paper (regardless of whether the code and data are provided or not)?
    \item[] Answer: \answerYes{} 
    \item[] Justification: We include the details of benchmark environment setting in Appendix~\ref{appendix:A} and experiment settings in Appendix~\ref{appendix:B}.

\item {\bf Open access to data and code}
    \item[] Question: Does the paper provide open access to the data and code, with sufficient instructions to faithfully reproduce the main experimental results, as described in supplemental material?
    \item[] Answer: \answerYes{} 
    \item[] Justification: We provide the source code of our method in the supplemental material.

\item {\bf Experimental setting/details}
    \item[] Question: Does the paper specify all the training and test details (e.g., data splits, hyperparameters, how they were chosen, type of optimizer) necessary to understand the results?
    \item[] Answer: \answerYes{} 
    \item[] Justification: We provide the benchmark environment configuration in Appendix~\ref{appendix:A} and the experimental settings in Appendix~\ref{appendix:B}.

\item {\bf Experiment statistical significance}
    \item[] Question: Does the paper report error bars suitably and correctly defined or other appropriate information about the statistical significance of the experiments?
    \item[] Answer: \answerYes{}{} 
    \item[] Justification: We report confidence intervals for the experiment results. 

\item {\bf Experiments compute resources}
    \item[] Question: For each experiment, does the paper provide sufficient information on the computer resources (type of compute workers, memory, time of execution) needed to reproduce the experiments?
    \item[] Answer: \answerYes{} 
    \item[] Justification: We include information on the computer resources in Appendix~\ref{appendix:B}.
    
\item {\bf Code of ethics}
    \item[] Question: Does the research conducted in the paper conform, in every respect, with the NeurIPS Code of Ethics \url{https://neurips.cc/public/EthicsGuidelines}?
    \item[] Answer: \answerYes{} 
    \item[] Justification: All authors have reviewed the NeurIPS Code of Ethics. 

\item {\bf Broader impacts}
    \item[] Question: Does the paper discuss both potential positive societal impacts and negative societal impacts of the work performed?
    \item[] Answer: \answerNA{} 
    \item[] Justification: There is no societal impact of the work. 
    
\item {\bf Safeguards}
    \item[] Question: Does the paper describe safeguards that have been put in place for responsible release of data or models that have a high risk for misuse (e.g., pre-trained language models, image generators, or scraped datasets)?
    \item[] Answer: \answerNA{} 
    \item[] Justification: The paper poses no such risks.

\item {\bf Licenses for existing assets}
    \item[] Question: Are the creators or original owners of assets (e.g., code, data, models), used in the paper, properly credited and are the license and terms of use explicitly mentioned and properly respected?
    \item[] Answer: \answerYes{} 
    \item[] Justification:  We cite the original papers of the datasets used for training (e.g., MimicGen~\cite{dataset:mimicgen} and DexMimicGen~\cite{dataset:dexmimicgen}).

\item {\bf New assets}
    \item[] Question: Are new assets introduced in the paper well documented and is the documentation provided alongside the assets?
    \item[] Answer: \answerNA{} 
    \item[] Justification: The paper does not release new assets.

\item {\bf Crowdsourcing and research with human subjects}
    \item[] Question: For crowdsourcing experiments and research with human subjects, does the paper include the full text of instructions given to participants and screenshots, if applicable, as well as details about compensation (if any)? 
    \item[] Answer: \answerNA{} 
    \item[] Justification: The paper does not involve crowdsourcing nor research with human subjects.

\item {\bf Institutional review board (IRB) approvals or equivalent for research with human subjects}
    \item[] Question: Does the paper describe potential risks incurred by study participants, whether such risks were disclosed to the subjects, and whether Institutional Review Board (IRB) approvals (or an equivalent approval/review based on the requirements of your country or institution) were obtained?
    \item[] Answer: \answerNA{} 
    \item[] Justification: The paper does not involve crowdsourcing nor research with human subjects.

\item {\bf Declaration of LLM usage}
    \item[] Question: Does the paper describe the usage of LLMs if it is an important, original, or non-standard component of the core methods in this research? Note that if the LLM is used only for writing, editing, or formatting purposes and does \emph{not} impact the core methodology, scientific rigor, or originality of the research, declaration is not required.
    \item[] Answer: \answerNA{} 
    \item[] Justification: The core method development in our paper does not involve LLMs as any core components.

\end{enumerate}


\end{document}